%% file: main.tex
\documentclass[10pt]{article} 
\usepackage[accepted]{tmlr}

\usepackage{blindtext}
\usepackage{amsmath}
\usepackage{natbib}
\usepackage{amssymb}
\usepackage{array}
\usepackage{mathtools}
\usepackage{stmaryrd}
\usepackage{graphicx}
\usepackage{enumitem}
\usepackage[linesnumbered,ruled]{algorithm2e}
\usepackage[table,dvipsnames]{xcolor}
\usepackage{multicol}
\usepackage{subcaption}
\usepackage{multirow}
\usepackage{float}
\usepackage{soul} 
\usepackage{hyperref}
\usepackage[utf8]{inputenc}
\usepackage{tcolorbox}
\tcbuselibrary{skins} 
\usepackage{ulem}
\usepackage{tikz}
 \usepackage{booktabs}
\usetikzlibrary{arrows.meta,positioning,calc}

\newcommand{\BlackBox}{\rule{1.5ex}{1.5ex}}  
\ifdefined\proof
    \renewenvironment{proof}{\par\noindent{\bf Proof\ }}{\hfill\BlackBox\\[2mm]}
\else
    \newenvironment{proof}{\par\noindent{\bf Proof\ }}{\hfill\BlackBox\\[2mm]}
\fi
\newtheorem{example}{Example} 
\newtheorem{theorem}{Theorem}
\newtheorem{lemma}[theorem]{Lemma} 
\newtheorem{proposition}[theorem]{Proposition} 

\newtheorem{corollary}[theorem]{Corollary}
\newtheorem{definition}{Definition}

\SetKwFor{For}{for}{do}{}
\SetKwFor{ForEach}{for each}{do}{}
\SetKwFor{While}{while}{do}{}
\SetKwIF{If}{ElseIf}{Else}{if}{do}{elseif}{else}{}
\newcolumntype{C}[1]{>{\centering\arraybackslash}m{#1}}

\newcommand{\Pa}{\mathrm{Pa}}
\newcommand{\An}{\mathrm{An}}
\newcommand{\Dom}{\mathrm{Dom}}
\newcommand{\cons}{\mathrm{\kappa}}
\newcommand{\Mod}{\mathcal{M}}
\newcommand{\Sol}{\mathrm{Sol}}
\newcommand{\cfx}{\mathrm{cfx}}
\newcommand{\CFX}{\mathrm{CFX}^{*}}
\newcommand{\Wit}{\mathcal{W}}
\newcommand{\AACtwo}{A_{\mathrm{AC2}}}
\newcommand{\restr}[2]{#1_{\restriction #2}}
\newcommand{\dom}{\mathrm{Dom}}
\newcommand{\vars}{\mathcal{V}}
\newcommand{\exo}{\mathcal{U}}
\newcommand{\model}{\mathcal{M}}
\newcommand{\assign}{\mathrm{F}}
\newcommand{\target}{T}
\newcommand{\causes}{\mathcal{C}}

\begin{document}

\setitemize{noitemsep}
\tcbset{
  examplebox/.style={
    colback=gray!10,        
    colframe=gray!50,       
    left=2mm,
    right=2mm,
    boxrule=0pt,            
    arc=1mm,                
    outer arc=0mm,
    fontupper=\itshape,     
    before skip=4pt,
    after skip=4pt,
    title={\textbf{\footnotesize Illustration}}, 
    coltitle=gray!50,       
    fonttitle=\footnotesize\bfseries,
    colbacktitle=gray!10,   
    boxed title style={size=small, colframe=gray!50, boxrule=0pt, interior hidden}
  }
}

\def\month{09}   
\def\year{2026}  
\def\openreview{\url{https://openreview.net/forum?id=3NlYZPCH9v}}

\title{Searching for Actual Causes:\\Approximate Algorithms with Adjustable Precision}

\author{\name Samuel Reyd \email samuel.reyd@telecom-paris.fr \\
        \addr LTCI, Télécom Paris, Institut Polytechnique de Paris
        \AND
        \name Ada Diaconescu \email ada.diaconescu@telecom-paris.fr \\
        \addr LTCI, Télécom Paris, Institut Polytechnique de Paris
        \AND
        \name Jean-Louis Dessalles \email jean-louis.dessalles@telecom-paris.fr \\
        \addr LTCI, Télécom Paris, Institut Polytechnique de Paris}

\maketitle

\begingroup
\renewcommand\thefootnote{}
\footnotetext{Disclosure: A large language model was used only for language polishing and readability improvements in parts of this manuscript. All technical content, claims, and final wording were reviewed by the authors.}
\addtocounter{footnote}{-1}
\endgroup

\begin{abstract}%
    Causality has gained increasing attention in recent years, notably for improving the interpretability of machine learning models. Yet the field of explainable artificial intelligence (XAI) has been criticized for emphasizing general tendencies rather than the situation-specific facts, which users typically expect as explanations. These expectations align with the notion of actual causes, which identify what made the observed outcome happen, in the specific context at hand. Halpern and Pearl provided a formal basis for actual causation, but identifying actual causes is NP-complete. Practical identification algorithms are extremely scarce, restricted to narrow classes of models, and typically identify only the shortest cause. We address this gap between the formal theory and its applicability through two main contributions. First, we introduce a baseline approximate polynomial-time algorithm with adjustable precision, together with two complementary algorithms that improve its efficiency. Second, we provide a theoretical result showing that the actual-cause identification problem can be decomposed into smaller sub-instances that preserve the set of solutions. This result directly motivates one of the complementary algorithms. Our experiments demonstrate that the baseline method can approximate the set of actual causes, notably 
    for non-Boolean and stochastic models, and that the complementary algorithms further improve its performance in most cases.
\end{abstract}


\section{Introduction}

Causality offers several critical advantages for machine learning by enabling models to generalize beyond correlations observed in training data. Causal models distinguish between associations and true cause-effect relationships, which allows robustness to distributional shifts and interventions \citep{scholkopfCausalRepresentationLearning2021,scholkopfCausalAnticausalLearning2012}. Such models also support counterfactual reasoning, which is essential for reliable decision-making in dynamic environments \citep{pearlCausality2009}. Incorporating causal knowledge enhances sample efficiency \citep{castroCausalityMattersMedical2020}, addressing one of the most important challenges in modern deep learning, namely, data scarcity \citep{alzubaidiSurveyDeepLearning2023}. Finally, causal frameworks improve interpretability by aligning explanations with human-understandable causal mechanisms \citep{moraffahCausalInterpretabilityMachine2020,schwabCXPlainCausalExplanations2019,madumalExplainableReinforcementLearning2020}. Significant research has been devoted to leveraging these strengths and identifying causal structures from data  \citep{guoSurveyLearningCausality2020,assaadSurveyEvaluationCausal2022}.

Most of this work, however, concerns general, type-level causation (which factors bring about a kind of outcome in general) whereas explanation calls for actual, token-level causation, which identifies the facts that caused a particular outcome \citep{halpernActualCausality2016,hitchcockPreventionPreemptionPrinciple2007}. This distinction is crucial because users typically desire specific and concrete explanations of the observed situation \citep{millerExplanationArtificialIntelligence2019}. For example, while the statement “Smoking causes cancer” provides a broad understanding of a causal tendency, a person diagnosed with cancer typically seeks a situation-specific explanation, such as “Your cancer was caused by your smoking.”

Such “token level” analysis is ubiquitous in Explainable Artificial Intelligence (XAI), notably via feature-attribution methods such as LIME \citep{ribeiroWhyShouldTrust2016} and SHAP \citep{lundbergUnifiedApproachInterpreting2017}, which score how much each input influences a black-box model's prediction. As \citet{millerExplanationArtificialIntelligence2019} argues, such a list of weighted contributions is not yet an explanation: an explanation must be contrastive and selective. These considerations hint toward the usage of actual causes. Beyond XAI, actual causes can be used to improve control over systems and mitigate unwanted behaviors \citep{reydRelevantCauses2025}, as “canceling” the cause can prevent the consequence. Actual causes can also be used to assess degrees of responsibility \citep{halpernModificationHalpernPearlDefinition2015} or assign blame \citep{chocklerResponsibilityBlameStructuralModel2004}.

Intuitively, an actual cause is a fact that made the difference to an observed consequence: had the cause been absent the consequence would not have occurred. However, providing a satisfactory definition of actual causes has been a longstanding challenge in philosophy \citep{humeHumaneNature,lewisCausation1974}, logic, and computer science \citep{halpernCausesExplanationsStructuralmodel2001,halpernCausesExplanationsStructuralModel2005a,beckersCausalSufficiencyActual2021}. \cite{halpernCausesExplanationsStructuralModel2005b} proposed the most influential formalism, often referred to as the HP definition or HP cause. Although it is widely used and regarded as a solid baseline, it still has limitations, such as its treatment of responsibility \citep{chocklerResponsibilityBlameStructuralModel2004} and normality \citep{halpernGradedCausationDefaults2015}. It also falls short of producing satisfactory causal explanations, which must be contrastive and account for user expectations \citep{millerExplanationArtificialIntelligence2019,millerContrastiveExplanationStructuralmodel2021}. 
However, the primary obstacle to adopting actual causation in XAI lies in its challenging identification and in the limited attention this task has received.

Identifying actual causes is NP-complete \citep{aleksandrowiczComputationalComplexityStructureBased2017a}. Practical solutions for finding exact or approximate HP causes are scarce. Approaches that remain close to the definition are generally restricted to identifying single causes in Boolean systems and require knowledge of the system’s equations \citep{ibrahimCheckingInferenceActual2020,hopkinsStrategiesDeterminingCauses2002}.  
Other approaches approximate alternative notions inspired by the HP definition \citep{albantakisWhatCausedWhat2019,reydCIRCEScalableMethodology2024}. 
Overall, to the best of our knowledge, no practical approach exists to identify the full set of HP causes, nor to identify any HP cause when the model is non-Boolean or stochastic. Table \ref{tab:causations} summarizes this situation: while general causation is widely studied, actual causation, especially its identification, receives limited attention. This paper addresses this gap in the causality and XAI literature.

\begin{table}[t]
    \centering
    \input{tables/causality-table}
    \caption{General causation versus Actual causation. This paper addresses the red area, i.e., identifying actual causes.}
    \label{tab:causations}
\end{table}

We provide two main contributions. (1) We propose a set of practical algorithms, with limited assumptions, polynomial-time complexity, and adjustable precision, for identifying the set of actual causes. 
This includes a baseline algorithm, the first to address full HP-cause identification while supporting non-Boolean and stochastic models, and two complementary algorithms that improve its efficiency in the vast majority of the cases we evaluated.
(2) We introduce and formally prove a theoretical result stating that the HP-cause identification task can be divided into sub-instances that preserve solutions, thereby easing the search process. One of our complementary algorithms is a direct implementation of this result.

We present three algorithms as our first contribution. The Minimal Beam Search (MBS) algorithm, our baseline, adapts Beam Search \citep{beamSearch} and leverages the minimality requirement of the HP definition to explore counterfactual worlds and identify actual causes. It uses an oracle to determine whether the consequence holds in a given counterfactual situation, along with a heuristic function that estimates how close the consequence is to being “canceled”. The Iterative Sub-Instance (ISI) algorithm leverages our theoretical result and, when the system structure is available, decomposes the search space into smaller subspaces. The Lower Upper Confidence Bound (LUCB) algorithm improves the efficiency of estimating the oracle and heuristic functions in stochastic models. Our algorithms assume variables with discrete domains. They can be used through the Python module \texttt{actualcauses}\footnote{\href{https://pypi.org/project/actualcauses/}{https://pypi.org/project/actualcauses/}}.

The novelty of our approach lies in two aspects. First, to the best of our knowledge, we are the first to propose algorithms to identify actual causes in general SCMs (instead of solely deterministic Boolean ones) and to aim for the full set of causes (instead of only one instance). Second, our theoretical results and the algorithm they inspire (our second algorithm) are fully novel. The first and third algorithms are somewhat novel as we propose adaptations from the baselines of the original version. However, we do not claim that algorithms 1 and 3 are substantially novel theoretical contributions. We claim practical novelty in the approach to the problem and theoretical novelty in the second contribution.

We evaluate our methods on a Boolean system of varying size introduced by \cite{ibrahimCheckingInferenceActual2020}, and extend it to non-Boolean and stochastic variants. Our results show that our baseline algorithm approximates the set of HP causes and enables an actionable trade-off between quality and efficiency. On these models, our complementary algorithms consistently improve the baseline’s performance. We further evaluate our algorithms on a corpus of small random SCMs. This corpus is used for an ablation study that quantifies the importance of each component of our deterministic and stochastic algorithms. They also reveal that our baseline (MBS) can outperform our improved ISI for models with causes of size 1. Finally, we present additional early-stage results in the appendix, including sanity checks on classical SCMs from the literature, the impact of the graph structure on our algorithms (revealing that ISI is affected by the graph structure and density while MBS is more affected by the graph size), and preliminary theoretical and empirical time complexity results (showing that while ISI is usually much faster than MBS, it can sometimes exhibit a worse complexity). The experimental code is provided in our public repository\footnote{\href{https://github.com/SamuelReyd/SearchingForCauses}{https://github.com/SamuelReyd/SearchingForCauses}}. 

\section{Background}

An actual cause is defined as a counterfactual or but-for cause, i.e., but for the cause, the consequence would not have happened. In this section, we define the formalism of Structural Causal Models (SCM), necessary for counterfactual reasoning, the formal definition of actual causation by Halpern and Pearl, and some of its limitations. We illustrate these notions using the classic rock-throwing example.

\begin{example}[Rock throwing]\label{example:Suzy}
    Suzy and Billy are throwing rocks at a bottle. Both aim accurately and Suzy hits the bottle first. Therefore, the bottle shatters. In this scenario, we intuitively consider that the cause of the bottle shattering is Suzy's throw or Suzy's hit. 
\end{example}



\subsection{Structural Causal Models} \label{subsec:causality}

\cite{pearlCausality2009} introduced SCMs to model counterfactual reasoning. In our example, we can model what would happen if Suzy did not hit. While still assuming she aimed correctly, we consider an imaginary world where, despite her accurate throw, she missed. An SCM consists of variables that model our observations of the system and of functions that set the variable values based on the system's dynamics. SCMs also include “exogenous variables” to initialize the endogenous variables' values. Each endogenous variable value is computed using the corresponding exogenous variable and a set of other endogenous variables called its “causal parents”. 

\begin{definition}[Structural Causal Model]
    We denote an SCM $\model=(\vars,\exo,\assign,\dom)$ where $\vars$ is the set of endogenous variables, the set $\exo$ of exogenous variables, the set $\assign$ of structural assignments (i.e., the functions mentioned above), and the set $\dom$ of domains. In this formalism, $\forall X \in \vars: X \coloneq \assign_X(\Pa(X),U_X)\in \dom(X)$, where $\Pa(X)$ denotes the inputs of $F_X\in\assign$, which are called “causal parents” of $X$ and $U_X\in\exo$ is the exogenous variable associated with $X$.
\end{definition}

In this paper, we focus on “recursive” SCMs, which assume a non-cyclic relationship of $\Pa$, i.e., if $X$ is an ancestor of $Y$, then $Y$ cannot be an ancestor of $X$. It is therefore often represented via a Directed Acyclic Graph (DAG) called a causal graph. 
A context, denoted $u$, is a complete assignment of values to the exogenous variables, $u \in \dom(\exo)$. Since the SCM is recursive, a context suffices to determine the values of all endogenous variables.
To state that a certain variable $X\in \vars$ took a certain value $x$ we denote: $(\model,u)\models(X=x)$. To model counterfactual reasoning, SCMs allow for interventions, i.e., forcing an endogenous variable to take a value regardless of the output of its structural assignment. Interventions can lead to changes in the other variables. If forcing the value of $Y$ to $y'$ changed the value of $X$, we write: $(\model,u)\models[Y\leftarrow y'](X\neq x)$.

\begin{tcolorbox}[examplebox]
    In example \ref{example:Suzy}, the endogenous variables are $ST$ (Suzy Throws), $BT$ (Billy Throws), $SH$ (Suzy Hits), $BH$ (Billy Hits), and $BS$ (Bottle Shatters).  The exogenous variables are $bt$ (Billy throws) and $st$ (Suzy throws), which are set to 1 by assumption. The domains are Boolean: $\dom(X) = \{0,1\}$ for any $X\in\vars$. The following structural assignments model the system mechanisms: $ST:=st$, $BT:=bt$, $SH:=ST$, $BH:=BT\land\lnot SH$, and $BS:=BH\lor SH$. The observed values in context $u=(1,1)$ are $(ST,BT,SH,\lnot BH,BS)$.
\end{tcolorbox}

\subsection{Actual Causation and the HP-Definition} \label{subsec:actual-causation}

Actual causation aims to identify which specific events in a particular situation were responsible for a given outcome. Notably, a cause is an event that took place before its consequence, such that if it did not occur, neither would the outcome. However, if Suzy had not thrown her rock, the bottle still would have shattered because of Billy's rock. To address this problematic case, the HP-definition \citep{halpernModificationHalpernPearlDefinition2015} includes a witness set $W$, composed of variables that are allowed to keep their actual values in the counterfactual world where the cause does not happen. 

\begin{definition}[HP-cause]\label{def:hp-cause}
    Let $\model=(\vars,\exo,\assign,\dom)$ be an SCM and $u\in\Dom(\exo)$ be a context for this SCM. $C\subseteq \vars$ is an HP cause of $\target\in\{0,1\}$ in $(\model,u)$ if:
    \begin{enumerate}
        \item[AC1] $(\model,u)\models (C=c^*)$ and $(\model,u)\models \target$
        \item[AC2] $\exists W \subseteq \vars\backslash C, \exists c\in\Dom(C)$ s.t. 
        \[(\model,u)\models (W=w^*)\qquad\forall X\in C: \restr{c}{X}\neq\restr{c^*}{X} \qquad(\model,u)\models [C\leftarrow c,W\leftarrow w^*]\lnot \target\]
        \item[AC3] No subset of C satisfies the above conditions. 
    \end{enumerate}
\end{definition}

Intuitively, AC1 ensures that the candidate cause and effect actually occurred, AC2 captures the idea that changing the cause (while holding certain variables fixed) would alter the effect, and AC3 guarantees that the cause is not unnecessarily large.

\begin{tcolorbox}[examplebox]
    In our example, $C=\{ST\}$ is an HP cause of $BS$, with witness $W=\{BH\}$ since $(\model,u)\vDash[ST\leftarrow0,BH\leftarrow0]\lnot BS$. The full set of causes is $\causes=\big\{\{ST\},\{SH\}\big\}$.
\end{tcolorbox}

\subsection{Limitations and Challenges}

The HP definition omits several considerations that are important for explanatory adequacy. It does not include notions such as normality \citep{halpernGradedCausationDefaults2015} or degrees of responsibility \citep{chocklerResponsibilityBlameStructuralModel2004}, and it does not directly account for relevance \citep{reydRelevantCauses2025} or contrastiveness \citep{millerContrastiveExplanationStructuralmodel2021}, which are central to human-oriented explanations. These limitations have motivated a large body of alternative definitions of actual causation. In this work, however, we focus on the identification problem and therefore adopt the standard HP definition, which is widely used as a solid formal basis. A more detailed discussion of these conceptual aspects is provided later in Section \ref{sec:discussion}.

Identifying HP-causes poses significant intrinsic challenges. The HP definition requires reasoning over counterfactual worlds and subsets of variables, which leads to a combinatorial search space. Formally, deciding whether a candidate event is an actual cause is NP-complete \citep{halpernModificationHalpernPearlDefinition2015,aleksandrowiczComputationalComplexityStructureBased2017a}. This complexity arises directly from the structure of the definition, including the need to test alternative assignments (AC2) and to enforce minimality (AC3), and affects even relatively small models.

These intrinsic difficulties motivate the need for practical methods to identify or approximate HP-causes. We next review prior work addressing actual-cause identification.

\section{Related Work}

Existing approaches for the actual cause identification task can be broadly categorized into three families: exact methods using formal solvers, methods based on alternative causation definitions, and counterfactual explanation techniques from XAI. We review each in turn, highlighting their scope and limitations relative to the identification problem we address.

\subsection{Exact HP Cause Identification}

Early theoretical work on HP causation focused on conceptual definitions or complexity analysis rather than practical algorithms \citep{halpernActualCausality2016}. \cite{hopkinsStrategiesDeterminingCauses2002,eiterComplexityResultsStructurebased2002a,aleksandrowiczComputationalComplexityStructureBased2014,halpernModificationHalpernPearlDefinition2015,aleksandrowiczComputationalComplexityStructureBased2017a}
established the computational hardness of checking and identifying causes, but suggested only brute-force enumeration, feasible only for toy examples with few variables.

More recent efforts have leveraged constraint solvers to make brute-force search tractable. \cite{ibrahimEfficientCheckingActual2019} encoded the HP definition as SAT clauses, enabling automated verification of candidate causes using SAT solvers. Building on this, \cite{ibrahimCheckingInferenceActual2020} reformulated the problem as integer linear programming (ILP) to optimize for minimal causes, and developed tooling to automatically translate formal models of cyber systems into Boolean causal models \citep{ibrahimActualCausalityCanvas2020}. While these solver-based methods represent significant progress, they remain restricted to Boolean domains and require complete knowledge of structural equations. 

\subsection{Alternative Definitions and Domain-Specific Methods}

Researchers have proposed modified definitions, more adequate for a specific domain, alongside specialized identification algorithms. \cite{albantakisWhatCausedWhat2019} developed an information-theoretic notion of actual causation for discrete-time transition systems, while \cite{rafieioskoueiEfficientDiscoveryActual2024} adapted the HP framework to temporal logics with corresponding model-checking procedures. In reinforcement learning, \cite{chuckAutomatedDiscoveryFunctional2024} trained classifiers to identify functional causes, a relaxation of HP causes. \cite{reydCIRCEScalableMethodology2024} introduced a data-driven method that blends actual and general causation. These approaches successfully address identification in their target domains, but the definitional changes mean they do not solve the core HP identification problem. 

\subsection{Counterfactual Explanations}

The XAI literature on counterfactual explanations \citep{guidottiCounterfactualExplanationsHow2024} addresses a closely related problem: given a model prediction, find alternative inputs that would yield different outputs. These methods typically minimize the distance between the original input and counterfactual instances.
While conceptually similar to actual cause identification \citep{chouCounterfactualsCausabilityExplainable2022a}, counterfactual explanation methods differ in key ways. They focus on machine learning models rather than SCMs, seek single counterfactuals rather than exhaustive cause sets, and do not incorporate witness sets $W$ from the HP definition. Specialized extensions exist: backtracking counterfactuals \citep{kladnyDeepBacktrackingCounterfactuals2024a} reason about exogenous variables in SCMs, while \cite{karimiAlgorithmicRecourseImperfect2020} adapt algorithmic recourse (counterfactual explanations focusing on the actionability of the counterfactual) to SCMs for specific model classes. However, none of these methods is easily adaptable to full HP identification across SCM types.

Overall, the literature reveals a clear gap: no existing method practically identifies the complete set of HP causes, notably for non-Boolean and stochastic SCMs. Solver-based methods are restricted to Boolean domains with known equations. Specialized definitions are limited to specific use-cases. Counterfactual explanation techniques lack the HP definition's witness sets and cannot target larger causes. We now introduce algorithms designed to fill this gap.

\section{Our Algorithms}\label{sec:method}

We present three complementary algorithms for identifying HP-causes with polynomial-time complexity and adjustable precision. The Minimal Beam Search (MBS) algorithm serves as our baseline, requiring only an oracle to evaluate counterfactual outcomes and a heuristic to guide the search. The Iterative Sub-Instance (ISI) algorithm improves upon MBS by exploiting causal graph structure, when available, to decompose the search space into smaller subproblems; an approach justified by our theoretical result showing that this decomposition preserves the set of HP-causes. The Lower-Upper Confidence Bound (LUCB) algorithm addresses stochastic models by adaptively allocating samples to interventions where the cancellation status or heuristic value remains uncertain, substantially reducing the number of model evaluations needed. While each algorithm can be used independently, they combine naturally: ISI can call MBS to solve subproblems, and LUCB can replace deterministic evaluation in either MBS or ISI when the model is stochastic.

\subsection{Minimal Beam Search (MBS)} \label{sec:mbs}

\subsubsection{Problem Formulation and Assumptions}

MBS relaxes the restrictive assumptions of prior HP-cause identification methods. Unlike solver-based approaches \citep{ibrahimCheckingInferenceActual2020}, which require complete knowledge of structural assignments and restrict to Boolean domains, MBS requires only: (1) the set of endogenous variables $\vars$, (2) their domains $\dom$, (3) an oracle function $\phi$ which evaluates whether a given intervention $e$ cancels the target consequence, i.e., $\phi(e) = 1$ if $(\model,u)\models [e](\target)$ and $\phi(e) = 0$ if $(\model,u) \models [e](\lnot\target)$, and (4) a heuristic function $\psi$ that estimates how “close” an intervention is to canceling $\target$, guiding the search toward promising regions.

The oracle can be realized in multiple ways: by applying structural assignments when available, by querying a simulator that supports interventions, or by experimenting on a controllable physical system. The heuristic embeds domain knowledge about what makes $\target$ likely or unlikely. 

We acknowledge that the heuristic is a significant practical requirement. Unlike the oracle, $\psi$ must be crafted for each problem domain. We provide empirical comparisons of some Boolean heuristic choices in Appendix~\ref{an:heuristics} but do not offer general construction guidelines. 
Additionally, Section~\ref{sec:discussion-algo-limitations} presents an example of a domain-guided heuristic that illustrates what a “promising” intervention is, i.e., how to evaluate if an intervention is close to canceling $\target$. This remains a limitation and an avenue for future work.

Let $e$ denote an intervention, formalized as a set of variable-value pairs: $e$ = $\{(Y_i, y_i)\}_{i=0}^n$ where $Y_i \in \vars$, $y_i \in \dom(Y_i)$, and $Y_i$ are distinct. We partition the variables in $e$ into the counterfactual variables $Ctf(e)$
and the actual variables $Act(e)$.
For Boolean models, we use compact notation: $\{\mathbf{Y}\}$ for $\{(Y,1)\}$ and $\{\mathbf{\lnot Y}\}$ for $\{(Y,0)\}$; we use bold notation to differentiate the positive value from the variable. For any HP-cause $C$ with witness $W$, there exists an intervention $e$ such that $Ctf(e) = C$ and $Act(e) = W$. 

\begin{tcolorbox}[examplebox]
    In the rock-throwing scenario (Example \ref{example:Suzy}), consider $e_1 = \{\mathbf{\lnot BT}\}$ and $e_2 = \{\mathbf{\lnot SH}, \mathbf{\lnot BH}\}$. We have $Ctf(e_1) = \{BT\}$ and $\phi(e_1) = 1$ (bottle still shatters), so $e_1$ does not identify a cause. For $e_2$, $Ctf(e_2) = \{SH\}$, $Act(e_2) = \{BH\}$, and $\phi(e_2) = 0$, identifying $\{SH\}$ as a cause with witness $\{BH\}$.
\end{tcolorbox}

\subsubsection{The MBS Algorithm}

MBS explores the space of interventions to identify all pairs (C, W) satisfying AC2 (counterfactual dependence) and AC3 (minimality). The algorithm adapts classical beam search \citep{beamSearch}. At each step $k$, Beam Search maintains a beam of $b$ interventions, i.e., those with the best heuristic values $\psi(e)$. Each intervention $e$ in the beam is expanded by adding one variable-value pair (counterfactual or actual) for each variable not in $e$. The algorithm terminates when all variables have been considered or when an early-stop condition is met (at least one cause found and no promising interventions remain).

We adapt Beam Search to HP-cause identification using three “minimality improvements”. MBS initializes only counterfactual values: it prevents evaluating interventions that have no effect. MBS prunes non-canceling branches: once $\phi(e) = 0$, we do not expand $e$ further, as any extension would be a superset of an already-identified cause. MBS enforces minimality during expansion: when generating candidate interventions for step $k+1$, it discards any $e$ where $Ctf(e)$ is a superset of an already-identified cause.

Algorithm 1 provides the full pseudo-code; we summarize key steps here. The algorithm iterates for at most \texttt{max\_steps} iterations, where \texttt{max\_steps} is a settable parameter. At each iteration, the beam is expanded (line 4), producing new interventions that respect minimality constraints. Interventions are evaluated with $\phi$ (line 7), and those with $\phi(e) = 0$ are filtered for minimality and added to the output set $C$ (line 8). Minimality filtering happens once during beam expansion to avoid evaluating interventions that would not be minimal with respect to previously identified causes. Another filtering is necessary (here line 8) to ensure that all causes identified during the current step are indeed minimal with respect to each other (see step 2 of the example in Appendix \ref{an:illustrations}). The remaining interventions ($\phi(e) = 1$) are scored by $\psi$, and the top b are retained for the next iteration (line 9). 

\input{algorithms/mbs}

\begin{tcolorbox}[examplebox]
    We start with interventions $\{\mathbf{\lnot BT}\}$, $\{\mathbf{\lnot ST}\}$, $\{\mathbf{BH}\}$, and $\{\mathbf{\lnot SH}\}$ that all have $\phi(e) = 1$. Using heuristic $\psi$ = "minimize positive variables" and beam size $b = 3$, we retain $\{\mathbf{\lnot BT}\}$, $\{\mathbf{\lnot ST}\}$, and $\{\mathbf{\lnot SH}\}$. Among the 18 expanded candidates, $\{\mathbf{\lnot ST}, \mathbf{\lnot BT}\}$ and $\{\mathbf{\lnot ST}, \mathbf{\lnot BH}\}$ have $\phi(e) = 0$. Since $Ctf(\{\mathbf{\lnot ST}, \mathbf{\lnot BH}\}) = \{ST\} \subseteq Ctf(\{\mathbf{\lnot ST}, \mathbf{\lnot BT}\}) = \{ST, BT\}$, only $C=\{ST\}$ qualifies as a cause with witness $W=\{BH\}$. See Appendix~\ref{an:illustrations} for the full execution trace.
\end{tcolorbox}

\subsection{Iterative Sub-Instance Decomposition (ISI)}\label{sec:isi}

The ISI algorithm leverages knowledge of the causal graph (DAG) to divide the search space into smaller, more tractable subproblems. This introduces an additional assumption beyond MBS: access to the DAG.  

\subsubsection{Intuition and Motivating Examples}

\input{figures/proof-graphs}

The key insight is that for any cause $C$, the causal effect must propagate from $C$ to $T$ through intermediate variables (descendants of $C$). Rather than searching for causes among all variables simultaneously, we can first search for causes in $\Pa(T)$, and then, for each identified cause, perform a “causal backtrack,” i.e., search in its parents. Finally, we can recursively repeat until no parents remain.

A naive approach to “causal backtracking” (i.e., searching only within the parents of an identified cause, and forgetting everything else) fails in several ways, illustrated by the four scenarios of Figure~\ref{fig:proof-graphs}. Each scenario tells us one thing a subproblem must record, or one branch it must explore.

In a “NOT”-structure such as Figure \ref{fig:not_graph}, if $\{A\}$ is a cause with witness $\{B\}$, then “causally backtracking” to search in $\Pa(A)$ must preserve $B$ as a witness; otherwise $X$ does not cancel $T$ (in addition to searching within the parents of the previously identified cause, we must retain the current witness). 

In an “XOR”-structure such as Figure \ref{fig:xor_graph}, if $\{A\}$ is identified as a cause, “causal backtracking” requires blocking the other parents of $\target$ (by fixing $B$ to its actual value) to reveal $\{X\}$ as a cause with witness $\{B\}$ (we must fix all variables we search through either to a counterfactual value or to their actual one).

In an “OR”-structure such as Figure \ref{fig:or_graph}, “causal backtracking” from $\{A,B\}$ requires considering all its subsets independently to identify causes such as $\{X,B\}$ (we must search within all their non-empty subsets).

Finally, in a “split”-structure such as Figure \ref{fig:split_graph}, “causally backtracking” from $A$ or $B$ individually fails, and we must “causally backtrack” from the non-minimal “canceling” intervention $\{A,B\}$ to identify $\{G,H\}$ (we must search within the parents of all the non-minimal “canceling” interventions).

\subsubsection{Subproblem Construction}

These observations lead to the following notion. Given a counterfactual  “canceling intervention” $(C,c)$ with witness $W$ and a non-empty subset $S\subseteq C$, a subproblem is a tuple $\cons=(L,I,R_C,r_C,R_W)$ of pairwise disjoint sets of variables, together with values $r_C$:
\begin{itemize}
    \item $L$: the settled variables, absorbed by the backtracking steps performed so far (initially $\{\target\}$)
    \item $I$: the free search space (variables we will search over)
    \item $R_C=C\setminus S$: forced counterfactual interventions (parts of $C$ not being “backtracked”)
    \item $r_C=c_{|R_C}$: values for the forced interventions
    \item $R_W=W$: forced witness (the witness of the intervention we “backtrack” from)
\end{itemize}

A subproblem $\cons$ constrains the search: we seek interventions that assign \emph{every} variable of $I$, either counterfactually or at its actual value, and that, combined with the forced interventions $R_C \leftarrow r_C$ and $R_W \leftarrow r^*_W$, cancel $\target$. The forced witness $R_W$ is associated with its actual values $r_W^*$ that are given by the context $(\model,u)\models(R_W=r_W^*)$ and do not need to be specified in the subproblem. The solutions of a subproblem are interventions $e$ on variables of $I$ that identify the cause $C=R_C\cup Ctf(e)$, with values $c$ given by $r_C$ on $R_C$ and by $e$ elsewhere, and the witness $W=R_W\cup Act(e)$. A subproblem is a search over the assignments of $I$, and costs $|\Dom(I)|=\prod_{X\in I}|\Dom(X)|$ oracle calls when solved exhaustively. The set of settled variables $L$ allows us to avoid including redundant variables in the search. Similarly, the “backtracking subset” $S$ could be taken from the entire cause $C$, but restricting it to $C\cap I$ removes redundant subproblems.

The examples in Figure~\ref{fig:proof-graphs} give the intuition behind each part of these formulas. Instead of naively backtracking and searching within the cause's parents, we replace \emph{any non-empty subset} by its parents (as prompted by Figure~\ref{fig:or_graph}). We then set the remainder of the counterfactual-valued variables (Figure~\ref{fig:or_graph}) and all the actual-valued variables (Figure~\ref{fig:not_graph} and~\ref{fig:xor_graph}) as forced interventions. Additionally, instead of backtracking only from HP-causes, we consider all “canceling” interventions (Figure~\ref{fig:split_graph}). These examples provide intuitions and illustrations; the proofs of why the exact formulas work are presented in Appendix~\ref{an:proofs}.

\subsubsection{The ISI Algorithm}

We initialize the algorithm with the subproblem $\kappa_0=(\{\target\},\Pa(\target),\emptyset,\emptyset,\emptyset)$, i.e., nothing is settled but the target, only variables in $\Pa(\target)$ are considered as part of the search space, using their full domain, and no additional constrained interventions are added. We also initialize an empty set of causes $\mathbf{C}$. We then use a queue to implement the recursive construction of “subproblems”. As long as the queue contains elements, we dequeue a new “subproblem” described by $(L,I,R_C,r_C,R_W)$. We then identify the set of “canceling” counterfactual interventions using some provided search procedure. For each “canceling” counterfactual intervention identifying $C\leftarrow c$ with witness $W$, if $C$ is minimal in $\mathbf{C}$, we add it to $\mathbf{C}$ and remove its supersets. We then iterate over non-empty subsets $S\subseteq C\cap I$. For each $S$, we build a new “subproblem” described by $L'=L\cup S$, $I'=\Pa(S)\setminus(C\cup W\cup L')$, $R_C'=C\setminus S$, $r_C'=c_{|R_C'}$, and $R_W'=W$. We finally add $(L',I',R_C',r_C',R_W')$ to the queue. Algorithm~\ref{algo:ISI} presents the pseudo-code for ISI.

\input{algorithms/isi}

\begin{tcolorbox}[examplebox]
    On Example \ref{example:Suzy}, ISI begins with $L=\{BS\}$ and $I = \{SH, BH\}$. The only assignment of $I$ that cancels $BS$ is the “canceling” intervention $\{\mathbf{\lnot SH},\mathbf{\lnot BH}\}$, which moves $SH$ off its actual value and holds $BH$ at its own, with cause $C_1=\{SH\}$ and witness $W=\{BH\}$. “Causally backtracking” with $S = \{SH\}$ settles $SH$ and yields $I' = \{ST\}$, $R_W' = \{BH\}$. Searching in $\{ST\}$ with enforced $BH \leftarrow 0$ identifies $C_2=\{ST\}$ as a cause. Since $ST$ has no parents, the algorithm terminates, having found both $C_1=\{SH\}$ and $C_2=\{ST\}$.
\end{tcolorbox}

\subsubsection{Correctness Guarantees and Practical Use}

The following theorem justifies the ISI construction and shows that it recovers exactly the HP-causes. We present it informally here. The formalized version and full proofs can be found in Appendix~\ref{an:proofs}.

\vspace{.5em}
\textbf{Theorem (ISI correctness)} \hspace{.3em} Every pair identified by the exact ISI algorithm satisfies AC1--AC2, every HP-cause is identified, and the inclusion-minimal identified causes are exactly the HP-causes.
\vspace{.5em}

Each subproblem searches the assignments of $I$, and $I$ is contained in $\Pa(S)$, so its cost is governed by the number of parents of the variables being “backtracked” rather than by the size of the model. The assignment must moreover be total (every variable appears either in $C$ or in $W$), so an exhaustive search costs $\prod_{X\in I}|\dom(X)|$ instead of $\prod_{X\in I}(|\dom(X)|+1)$, the usual cause search having to check an additional “unused” state for each variable. This can nonetheless remain expensive on large instances. We therefore present three further ways to use ISI: a decomposition result that preserves exactness under additional assumptions, the use of MBS to search the subproblems, and approximations specific to ISI itself.

We propose an additional way to reduce the size of the subproblems. This reduction is applicable only when the parents of all variables in $C\cap I$ are outside of $L\cup C\cup W$ and are distinct. The solutions of the subproblem obtained by “backtracking” on several of them at once can then be assembled from the solutions of the subproblems obtained by “backtracking” on each of them separately: we combine one local solution per variable and keep the combinations that cancel $\target$.

Searching subproblems with MBS yields an approximate ISI that trades completeness for efficiency. MBS usually stops before every variable of $I$ has been assigned. Since a solution has to assign all of $I$, the missing variables must be assigned before “backtracking”. We present two strategies. A \emph{naive} approach holds every unassigned variable at its actual value without any further evaluation: it costs nothing, but the completed intervention is not guaranteed to cancel $\target$. A \emph{greedy} approach assigns the unassigned variables one at a time, trying the actual value first and keeping any value that preserves the cancellation. At least one value always works: forcing a variable to the value it already takes does not change the outcome of an intervention. A variable therefore costs at most $|\dom(X)|-1$ evaluations, and the completed intervention still cancels $\target$. Leaving the assignment partial would leave free variables behind, and the number of subproblems would grow quickly. In practice, one can either run MBS with the minimality condition and rely heavily on the complementary assignment strategy, or run it without that condition (i.e., as basic Beam Search).

We also propose approximate restrictions on the number of subproblems constructed. A \emph{maximum backtracking depth} stops the recursion after a fixed number of levels, which keeps the exploration close to the target and identifies the causes that lie near it. A \emph{subproblem sampling limit} keeps only a fixed number of the subproblems produced at each level, drawn at random, bounding the width of the exploration rather than its depth.

ISI reduces the search space by replacing one search over all the variables with several searches over smaller instances. However, the number of subproblems depends on the graph shape and on the size of the interventions identified during the sub-searches. Section~\ref{sec:results} compares the two algorithms empirically, and Appendix~\ref{an:sweep} analyzes how they compare on specific graph structures.

\subsection{Adaptive Sampling for Stochastic Models (LUCB)}

When the SCM is stochastic, either inherently or because interventions are modeled probabilistically (e.g., aggregated variables set via random micro-level actions), the oracle $\phi$ and heuristic $\psi$ return random values. A naive approach would consist of sampling each intervention a fixed N times and applying MBS or ISI to the estimates $\widehat\phi(e)$ and $\widehat\psi(e)$, where an intervention is considered to “cancel” $\target$ when $\phi$ is below a certain “canceling” threshold $a$. This would be wasteful: most interventions can be confidently classified with far fewer samples, while borderline cases require extensive sampling. LUCB addresses this by maintaining statistical confidence intervals (upper and lower bounds) around each estimate and adaptively resampling only interventions whose bounds remain ambiguous relative to decision thresholds, accounting for small tolerance slacks $\varepsilon$ that prevent excessive sampling on near-ties.

The LUCB algorithm adapts the Lower-Upper Confidence Bound framework \citep{kaufmannInformationComplexityBandit2013} to allocate samples adaptively, focusing computational effort on interventions whose status remains uncertain. This approach was successfully employed in the Anchors explanation method \citep{ribeiroAnchorsHighPrecisionModelAgnostic2018} for a similar beam-selection problem.

\subsubsection{Confidence Bounds and Decision Criteria}

For each intervention $e$, LUCB maintains empirical estimates $\widehat\phi(e)$ and $\widehat\psi(e)$ along with confidence intervals constructed from concentration inequalities. Because $\phi(e)$ is Bernoulli (binary outcome) and $\psi(e)$ is real-valued with unknown variance, we employ different bounds.

For $\phi$, we use the Chernoff bound \citep{boucheronConcentrationInequalitiesNonasymptotic2013} to compute the tightest intervals $[L_\phi(e), U_\phi(e)]$ such that $P\big(L_\phi(e)\le E[\phi(e)] \le U_\phi(e)\big) \ge 1 - \delta$. With $n(e)$ the number of samples, the upper bound $U_\phi(e)$ is then the largest
$q\ge \widehat{\phi}(e)$ satisfying

\[
n(e)\mathrm{KL}(\widehat{\phi}(e)\Vert q)
  =  \log\frac{2}{\delta},
\]

\noindent and $L_\phi(e)$ is the smallest $q\le \widehat{\phi}(e)$ solving the same equation \citep{garivierKLUCBAlgorithmBounded2011}. In practice, these bounds are computed via a simple monotone numerical root-finding procedure on $q\in[0,1]$. An arm is confidently “canceling” $\target$ if $U_\phi(e) < a + \varepsilon_c$ and confidently “non-canceling” if $L_\phi(e) > a - \varepsilon_{nc}$, where $a$ is the “canceling” threshold. Arms not satisfying either inequality remain ambiguous.

For $\psi$, we use the Bernstein inequality to derive the empirical Bernstein bounds \citep{maurerEmpiricalBernsteinBounds2009}. With $s^2(e)$ the empirical variance of $\psi$, $\delta$ the confidence level and $\beta=\text{log}\ 2/\delta$, we have:

\[
U_\psi(e)
  = \widehat{\psi}(e)
    + \sqrt{\frac{2s^2(e)\,\beta}{n(e)}}
    + \frac{3\,\beta}{n(e)},
\qquad
L_\psi(e)
  = \widehat{\psi}(e)
    - \sqrt{\frac{2s^2(e)\,\beta}{n(e)}}
    - \frac{3\,\beta}{n(e)}.
\]

Let $B$ be the beam composed of the $b$ (beam-size) interventions with the smallest empirical $\widehat{\psi}$ and satisfying $\widehat\phi(e)\ge a$. We identify the highest upper bound in the beam $u = \arg\max_{e\in B} U_\psi(e)$ and the lowest lower bound outside the beam $\ell = \arg\min_{e\notin B} L_\psi(e)$. The beam is certified when $U_\psi(u) - L_\psi(\ell)\le \varepsilon_b$. If this condition does not hold, we identify the set of disputed arms which are resampled in the next batch. Given a set of interventions $E$ to evaluate, the set of unsure ones can be written: $S = \{e\in E:L_\psi(e) \le U_\psi(u) + \varepsilon_b\ \text{and}\ U_\psi(e) \ge L_\psi(\ell) - \varepsilon_b\}.$

LUCB addresses two distinct decisions for each intervention $e$: whether it confidently cancels $\target$ (i.e., whether $\mathbb{E}[\phi(e)]$ falls below the threshold $a$), and whether it belongs in the beam (i.e., whether it ranks among the $b$ interventions with the smallest $\psi$). Neither is a ranking problem (both are thresholding and membership decisions) so the goal is not to find the single best intervention but to certify, with high probability, which interventions fall on which side of each boundary. The confidence-bound framework is natural here precisely because it enables \textit{adaptive} sampling: interventions whose status is already clear require no further samples, while only the ambiguous borderline cases are resampled.

\subsubsection{Execution Flow}

After an initial sampling phase in which each intervention is evaluated a fixed number of times, the algorithm proceeds iteratively by alternating between: (i) identifying unsure arms that require resampling and running the new batches, (ii) updating the bounds using the newly acquired statistics, and (iii) updating borderline indices and computing the stopping criterion. 
The loop terminates once both the canceling decisions and the beam membership are stable within their respective tolerances or when a predefined computational budget is reached. Algorithm \ref{algo:lucb} illustrates this process.

\input{algorithms/lucb}

\section{Experiments} \label{sec:experiments}

This section presents the SCMs used to evaluate the proposed algorithms and describes the experimental protocol and evaluation metrics. 

The main experiments (Section~\ref{subsec:smk}) are designed to assess three aspects of our contributions: (i) the ability of the proposed baseline to approximate HP-cause identification, (ii) the impact of complementary algorithms on efficiency and scalability, and (iii) the applicability of our approach beyond deterministic Boolean models.

Because no standard benchmarks exist for HP-cause identification beyond highly restricted settings, our main evaluation uses a single family of SCMs that can be systematically varied in size, determinism, and variable domains. All SCMs used in these experiments derive from a base scenario introduced by \cite{ibrahimCheckingInferenceActual2020}. We present the SCM modeling this scenario and describe two modified versions, a non-Boolean version and a stochastic version, introduced to illustrate our methods' wider range of applicability.

This controlled setup allows us to isolate the behavior of the algorithms under comparable conditions, but it fixes the graph structure and the size of the causes, which significantly influences the comparison between our two algorithms. We therefore complement our experiments with a corpus of random SCMs, described in Section~\ref{sec:random-corpus}, on which every component of both algorithms is ablated and the reference causes are obtained by exhaustive enumeration. In addition, we report sanity checks on small SCMs from the philosophical actual causation literature in Appendix~\ref{an:sanity}, which constitutes the only available way to validate correctness on well-established conceptual examples. Finally, Appendix~\ref{an:sweep} reports complementary results on synthetic SCMs with controlled structural properties.

\subsection{The SMK Model Family} \label{subsec:smk}

\begin{figure*}
    \centering
    \includegraphics[width=.9\linewidth]{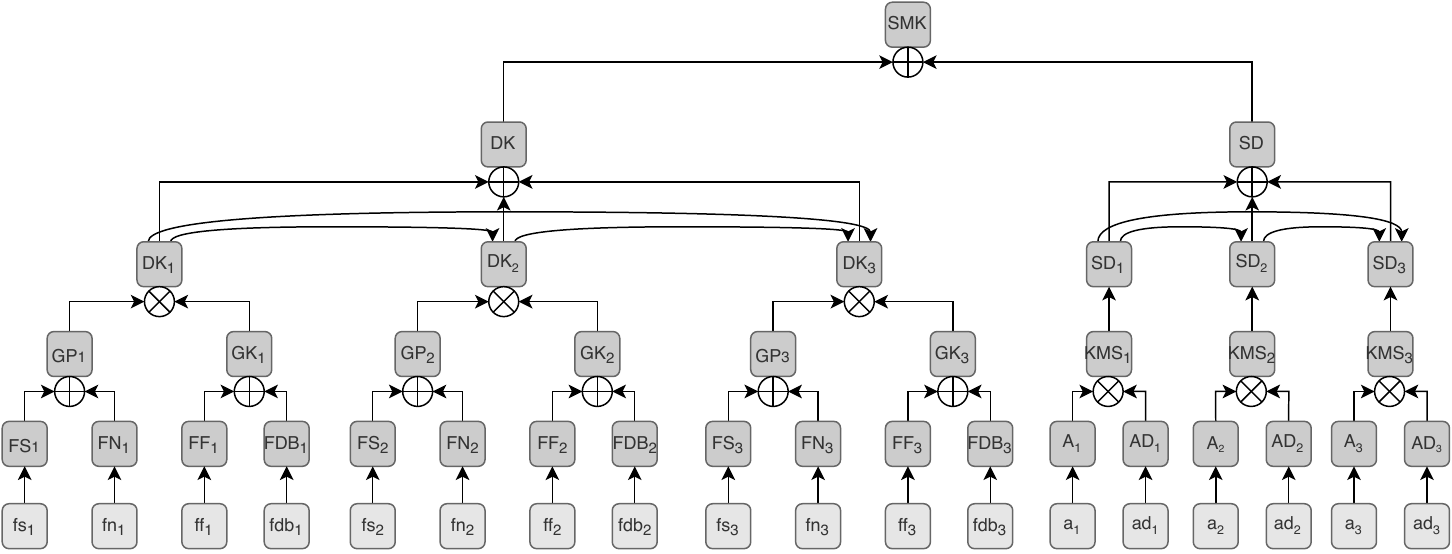}
    \caption{Diagram for the original SMK scenario with 3 attackers.}
    \label{figure:smk-diagram}
\end{figure*}

\paragraph{Base SCM.} The SMK scenario supposes that $n$ attackers attempt to obtain the master key of a computer. The target predicate $SMK$ is true if an attacker steals or decrypts the master key. These events are observed via the Boolean variables $SD$ (steal decrypted) and $DK$ (decrypt key), which are true if at least one attacker performs the corresponding action. 
For each attacker $i$, child variables model the different ways in which the key or the passphrase can be obtained. An attacker $i$ can steal or decrypt the key only if no previous attacker has already done so. For instance, if attackers 1 and 2 both meet the conditions to steal the key, then $SD_1$ is true, and $SD_2$ is false. The key is decrypted if an attacker $i$ gets the key ($GK_i$) and the passphrase ($GP_i$). The key is stolen if an attacker $i$ gets it from Key Management Service ($KMS_i$). An attacker $i$ can obtain the key from a file ($FF_i$) or a database ($FDB_i$), and can get the passphrase from the script ($FS_i$) or the network ($FN_i$). To obtain the key from $KMS$, an attacker $i$ must have access to it ($A_i$) and attach a debugger ($AD_i$). The structural equations for $n$ attackers are as follows: : $\mathbf{SMK} = DK \lor SD$, $\mathbf{DK} = DK_1 \lor ... \lor DK_n$, $\mathbf{SD} = SD_1 \lor ... \lor SD_n$, $\forall i: \mathbf{DK_i} = GP_i \land GK_i \land\bigwedge_{j\in\llbracket1,i-1\rrbracket} \lnot DK_j$, $\mathbf{SD_i} = KMS_i \land\bigwedge_{j\in\llbracket1,i-1\rrbracket} \lnot SD_j$, $\mathbf{GP_i} = FS_i \lor FN_i$, $\mathbf{GK_i} = FF_i \lor FDB_i$, and $\mathbf{KMS_i} = A_i \land AD_i$. The DAG with visual insights into the structural functions is shown in Figure \ref{figure:smk-diagram}.

\paragraph{Non-Boolean SCM.} We introduce a modified version of the SMK SCM that models the same scenario with a different representation. In this new version, variables $GP_i$ and $GK_i$ take values in larger discrete domains and are computed using summation rather than Boolean operators: $\dom(GP_i)=\dom(GK_i)=\{0,1,2\}$. Variables $DK_i$ retain their Boolean domains and are defined as $\mathbf{DK_i} = GP_i>0 \land GK_i>0 \land\bigwedge_{j\in\llbracket1,i-1\rrbracket} \lnot DK_j$.

\paragraph{Stochastic SCM.} To evaluate our algorithms in a stochastic setting, we further modify the SMK SCM by introducing noise into the structural equations. Each variable has a probability $\varepsilon_n$ (the noise level) of flipping the value prescribed by its structural function. Concretely, each variable follows a Bernoulli distribution $\mathcal{B}(\varepsilon_n)$ or $\mathcal{B}(1-\varepsilon_n)$: 

\[
X \sim  
    \begin{cases}
        \mathcal{B}(1-\varepsilon_n)  & \text{if } F_X(PA_X,U_X)=1 \\
        \mathcal{B}(\varepsilon_n) & \text{otherwise}
    \end{cases}
\]

\subsection{Experimental Setup for the SMK Family} \label{sec:exp-setup}

We test our algorithms on several sizes of the SMK model. For each experiment, we consider models with 2, 5, and 10 attackers. For each configuration, we generate 20 unique contexts by repeatedly sampling exogenous variables and retaining only contexts in which half of the exogenous variables are true and half are false, with the outcome $SMK$ being true. This ensures comparable and balanced contexts across runs.

To evaluate full HP-cause identification, we run the algorithms with beam sizes growing exponentially from 2 to 256, across all model variants (Boolean, non-Boolean, and stochastic), for each context and algorithm (MBS and ISI). 
We set $\texttt{max\_steps}$ to 7, which is another source of approximation our algorithm allows.
ISI is approximated using MBS with the reported beam size and maximum number of steps. The inner MBS is set to return all counterfactual interventions instead of only the minimal ones, i.e., the minimality improvements are turned off. We use the “greedy” assignment method for variables MBS did not assign (this happens when the search space is larger than the \texttt{max\_steps} parameter). We also use our additional decomposition step. For stochastic SCMs, each experiment is repeated with 10 different random seeds, and results are averaged across seeds. We use slacks of 10\%, a canceling threshold $a=0.65$, a confidence level $\delta=10\%$, $N=50$ samples per intervention, an initial batch size of 30, and a batch size of 10. The noise level is set to $\varepsilon_n=1.5/|\vars|$, yielding an average of 1.5 value flips per model evaluation.

For the smallest cause identification task, we run MBS with early stopping, with no maximum number of steps, using the same beam sizes and considering models with 2 to 15 attackers. Here, ISI is used with two additional approximations: its internal MBS uses early stopping and a “naive” assignment strategy. We also compare our results to the approach of \cite{ibrahimCheckingInferenceActual2020}, which we reimplemented using their ILP formulation and the Gurobi solver. 

Across all experiments, we use the same base heuristic, which minimizes the number of variables with an observed value of 1. In the non-Boolean setting, Boolean variables are treated as integers, and the heuristic sums the values of all endogenous variables. Additional results for other heuristics are reported in Appendix~\ref{an:heuristics}.

\subsection{Evaluation of the SMK Family} \label{subsec:evaluation} 

To evaluate the smallest-cause identification task, we compute the expected size of the smallest cause as the sum of $SD$ and $DK$ in the actual state. We report the accuracy score as 1 when an algorithm identifies a smallest cause of the correct size and 0 otherwise.

For the full identification task, the algorithms output a set of causes. We evaluate it by comparing this set to a reference set of expected causes. The reference set is computed based on the graph structure: the expected causes only depend on 1 or 2 attackers independently and can be reconstructed easily. The stochastic and non-Boolean versions share the same solutions as their deterministic Boolean counterpart. Hence, we used the same set of contexts for the different SCM versions and reused the same reference sets. Set similarity is measured using the DICE score: $dice(A,B)=\frac{2|A\cap B|}{|A|+|B|}$. We report DICE scores in \% for better readability.

In addition to quality metrics, we report the number of calls to the model, which reflects the computational cost of our approach while remaining independent of the internal implementation or complexity of the model. In our experiments, one model call corresponds to a single evaluation of both $\phi$ and $\psi$. For comparisons with the ILP-based approach, we report runtime instead, as that method does not rely on explicit model calls.

We summarize the 20 contexts by their median and report the 25–75 quantiles (interquartile range) as error bars, both for the quality measures (DICE) and for the number of calls. The only exception is the binary accuracy metric, which we summarize by its mean. We use the interquartile range rather than the standard deviation because the variance can be large — some contexts are more challenging than others, and small changes in the identified cause set (e.g., missing a single cause) can substantially affect the DICE score by generating many supersets — so standard-deviation bars would be hard to read and could even exceed the natural [0, 100] range of the DICE score.

\subsection{Random Graph Corpus} \label{sec:random-corpus}

The SMK family fixes both the graph's shape and the size of the causes. We use a corpus of small random SCMs to analyze our algorithms' components on more general SCMs. We keep these models small so we can perform exhaustive identification.

A model is drawn by taking $|\vars|$ Boolean variables in a topological order and making each earlier variable a parent of each later one with probability $p$. Variables the target does not depend on are pruned, and models whose target has no actual cause are discarded. Each mechanism is drawn either uniformly over the domain or, with probability $\beta$, as a \emph{coincidence gate} that takes one value on exactly one combination of its parents and another value elsewhere. The two mechanism types produce causes of different sizes. Flipping one parent of a uniform mechanism usually changes its value, so uniform models mostly have causes of size 1. A coincidence gate must be hit exactly, so its causes have the size of its in-degree. We use an uninformative heuristic equal to the oracle's output as we have no information on the models' shape.

The corpus contains $300$ models, $50$ for each $|\vars| \in \{5,\dots,10\}$, with $p$ drawn from $\{0.35, 0.5, 0.65, 0.8\}$ and $\beta$ from $\{0, 0.3, 0.6, 0.85, 0.95\}$. We have $168$, $69$, and $63$ models with domains of sizes $2$, $3$ and $4$ respectively. We obtain $128$, $88$, $37$ and $47$ models with a largest cause of size $1$, $2$, $3$ and at least $4$ respectively.

\subsection{Ablation Study} \label{sec:ablation}

We use our corpus of small random graphs to perform an ablation study of our algorithms. Each configuration modifies one component of one algorithm.

For ISI, we first disable the additional decomposition. We then bound the exploration of the subproblem tree, with a maximum “backtracking” depth of 1 to 3 levels and a sampling limit of 1 to 64 subproblems per level. We solve the subproblems with MBS rather than exhaustively. We tested an inner MBS returning all “canceling” interventions or only the minimal ones, which we complete either greedily or naively. We also run exact ISI on a DAG with one or two spurious parents added per variable, and on a DAG with up to one or two parents removed per variable, leaving the mechanisms untouched.

For MBS, we vary the beam size (4, 16, and 256) and the maximum number of steps (2, 3, and 5) one at a time. We disable the three minimality mechanisms in turn: not including actual values in the first search step, not expanding “canceling interventions”, and pruning supersets of “canceling interventions”.

Results are evaluated with the DICE score and the number of model calls similarly to the previous experiments. Additionally, since these models are simpler, we report the fraction of solved models, i.e., where the algorithm returned exactly the expected causes. We also report for each configuration the median over models of its number of calls divided by that of the reference configuration of its own family (\emph{ISI exact} or \emph{MBS exact}). This \emph{ratio} reveals how effective the approximation was.

\subsection{LUCB Study on Random Corpus} \label{sec:stochastic-setup}

The stochastic experiments of Section~\ref{sec:results} vary the model but keep the estimator's parameters fixed. We reuse the corpus of Section~\ref{sec:random-corpus}, this time varying the estimator and its parameters rather than the model.

We use the same corpus and repeat each configuration over 3 random seeds. Noise is introduced exactly as in the stochastic SMK model of Section~\ref{subsec:smk}: at every evaluation, each variable takes, with probability $\varepsilon_n$, a value other than the one its mechanism prescribes, drawn uniformly among the others, independently across variables and across evaluations. The reference causes remain those of the underlying deterministic model. The search is fixed to MBS with a beam of 8 and a limit of 5 steps in every configuration, so that the estimator is the only component that varies.

As for the stochastic SMK model evaluation, we compare our LUCB estimator with the naive, fixed-sample estimator. We vary the noise level $\varepsilon_n \in \{0.02, 0.05, 0.10\}$ (both for LUCB and the naive estimator), the budget $N \in \{50, 200\}$, the canceling threshold $a \in \{0.15, 0.30\}$ and the confidence level $\delta \in \{0.05, 0.10\}$. Averaging has no confidence level, and we only run it at $a = 0.30$, so the two estimators are compared at that threshold, and the threshold is varied within LUCB alone.

As in Section~\ref{sec:ablation}, we report the DICE score, the fraction of solved models, and the number of calls. Additionally, we report the number of calls of \emph{LUCB relative to naive averaging}. Then, we also report the frequency of incorrect canceling decisions: the fraction of evaluated interventions that the estimator places on the wrong side of the threshold with respect to the deterministic model. This metric reveals whether a mistake comes from misclassifying an intervention or from the search itself.

\section{Results} \label{sec:results}

Our main experiments on the SMK family support four main findings:
(1) the beam size parameter provides control over the trade-off between quality and computational cost;
(2) ISI substantially improves full HP-cause identification compared to MBS;
(3) ISI identifies the smallest cause more efficiently than both ILP and MBS;
(4) LUCB reduces the number of model calls in stochastic settings with limited impact on quality.

Our ablation study and stochastic study report insights about each component of our algorithms. Additionally, our small random corpus shows one important pattern for the exact version of our algorithms: \textbf{MBS outperforms ISI for small random models with only singleton causes}. Together with some specific structures reported in Appendix~\ref{an:sweep}, this is the only case where MBS outperforms ISI. 

\subsection{The Beam Size Parameter} \label{subsec:result-bs-smk}

\begin{figure}
    \centering
    \includegraphics[width=\linewidth]{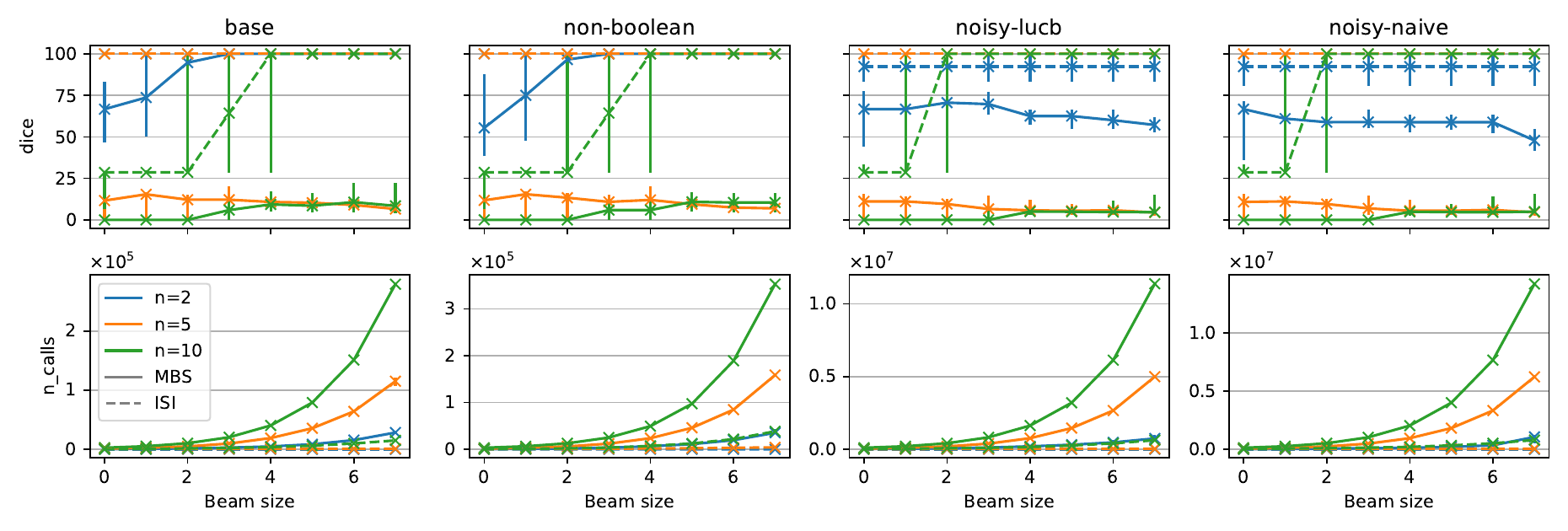}
    \caption{DICE score and number of calls for full cause identification in our three versions of SMK (column~1: base Boolean model; column~2: non-Boolean model; columns~3 and~4: stochastic model). The line color corresponds to the number of attackers. The linestyle corresponds to the identification algorithm.
    The x-axis uses a logarithmic scale. The bottom plots do not share the same scale. The lines report the median values for the 20 contexts and the error bars report the 25-75 quantiles.}
    \label{fig:main-fig}
\end{figure}

Figure \ref{fig:main-fig} shows that across all settings, increasing the beam size monotonically increases or preserves both quality and cost. The number of model calls follows a clear exponential trend, which is consistent with a polynomial complexity when accounting for the logarithmic scale of the x-axis (see Appendix~\ref{an:complexity-analysis} for details). The DICE score increases clearly in several configurations, notably for MBS with $n = 2$ and for ISI with $n = 10$ in the deterministic models. In other cases, the DICE score remains flat, particularly for MBS in the noisy model with $n = 2$ and in deterministic models with $n \geq 5$. This suggests that the baseline algorithm struggles in these settings and would require larger beam sizes to improve.

Quantitatively, doubling the beam size increases the DICE score by up to 19.1 points for MBS (from $b = 8$ to $b = 16$ in the base model with $n = 2$) and up to 14.3 points for ISI (from $b = 8$ to $b = 16$ in the base model with $n = 10$).

\begin{figure}
    \centering
    \begin{subfigure}{.32\linewidth}
        \includegraphics[width=\linewidth]{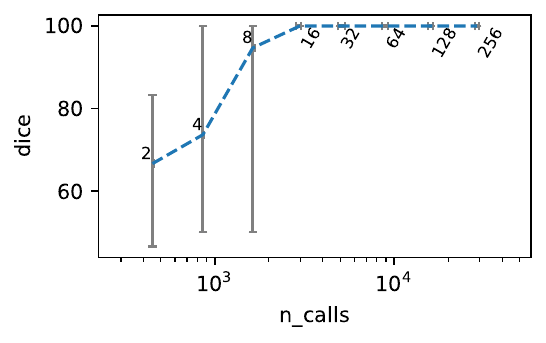}
        \caption{Full identification, MBS, \\base model, $n=2$.}
        \label{fig:tradeoff-n=2}
    \end{subfigure}
    \hfill
    \begin{subfigure}{.32\linewidth}
        \includegraphics[width=\linewidth]{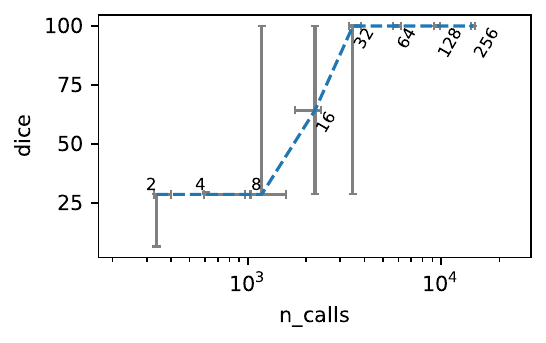}
        \caption{Full identification, ISI, \\base model, $n=10$.}
        \label{fig:tradeoff-n=10}
    \end{subfigure}
    \hfill
    \begin{subfigure}{.34\linewidth}
        \includegraphics[width=\linewidth]{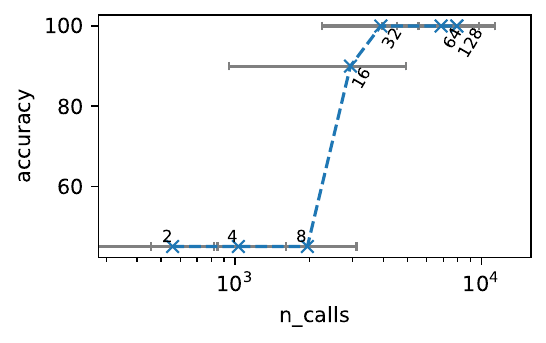}
        \caption{Smallest identification, MBS,\\ base model, $n=10$.}
        \label{fig:tradeoff-n=7}
    \end{subfigure}
    \caption{Tradeoff between quality of the identified causes and number of calls to the model. The x-axis uses a logarithmic scale. The lines report the median for 20 contexts for the left and center figures and the mean for the right figure (as accuracy is a binary metric). Error bars report the 25-75 quantiles for the first two figures and standard deviation for the third. There is no error bar on the y-axis of the third figure as accuracy is binary. In figure~\ref{fig:tradeoff-n=7}, $b=256$ and $b=128$ produce the same point.}
    \label{fig:tradeoffs}
\end{figure}

Figure \ref{fig:tradeoffs} illustrates three representative cases where increasing the beam size improves quality. We plot the quality measure (accuracy in Figure \ref{fig:tradeoff-n=7} and DICE in Figures \ref{fig:tradeoff-n=2} and \ref{fig:tradeoff-n=10}) against the number of model calls. These plots illustrate how the beam size acts as a practical lever to control the quality–efficiency trade-off.

\subsection{Impact of ISI on Full Identification in the SMK Model} 

For $n \leq 5$, Figure \ref{fig:main-fig} shows that ISI achieves near-perfect DICE scores across all beam sizes. For $n = 10$, the DICE score starts lower but increases steadily as the beam size grows.

In all evaluated models, ISI is extremely efficient and requires orders of magnitude fewer model calls than MBS. Additionally, ISI manages exact identification of the full set of HP causes for all model sizes considered in our experiments. Identifying the full set of causes took $0.02s\pm0.04$ ($\sim500\pm700$ calls), $0.15s\pm0.08$ ($\sim3e3\pm2e3$ calls), and $1.23s\pm0.32$ ($\sim6e4\pm2e4$ calls) respectively for 2, 5, and 10 attackers.

Overall, replacing MBS with ISI increases the DICE score by up to 90 points (e.g., for $n = 5$ and $b = 8$ in the base model). ISI reduces the number of model calls by up to 99\% (e.g., for $b = 256$ and $n = 2$ in the base model) and always achieves at least a 56\% reduction compared to MBS. A full comparison between MBS and ISI is reported in Appendix~\ref{an:algo-gain}.

\subsection{Smallest Cause Identification}

\begin{figure}
    \centering
    \includegraphics[width=.8\linewidth]{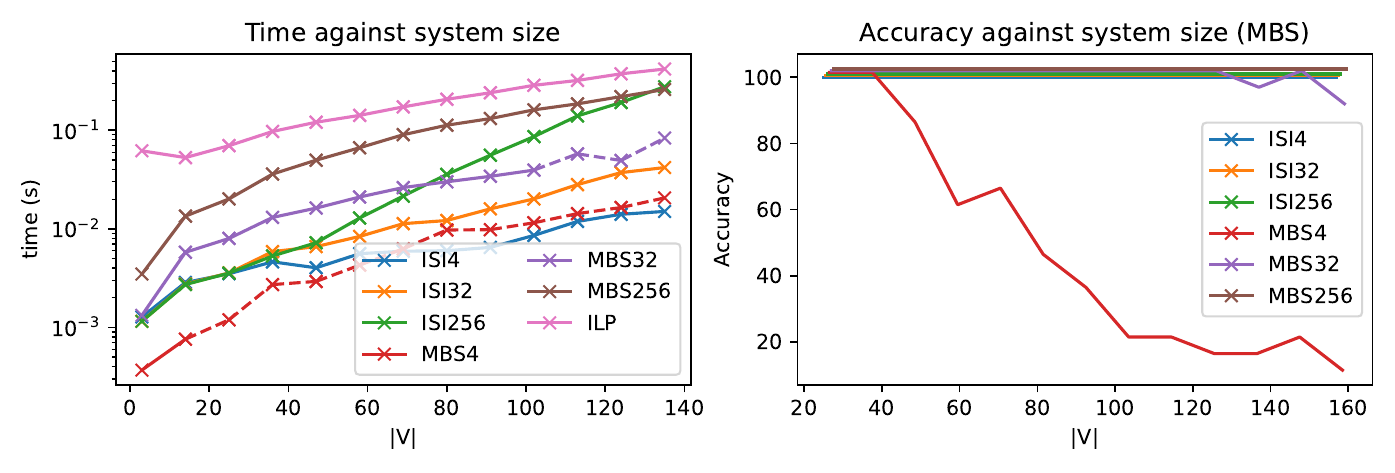}
    \caption{Time and accuracy vs. number of variables for smallest cause identification using ILP and various beam sizes for MBS and ISI. In the left figure, dotted lines are used to link points for which the accuracy is not 100\%.}
    \label{fig:smallest}
\end{figure}

Figure \ref{fig:smallest} compares the runtime of the different algorithms (ILP and our methods with varying beam sizes) on the left, and the accuracy score on the right, as a function of model size (number of endogenous variables). We omit the accuracy plots for ILP, which is an exact method.

The results show that MBS performs well on small models but exhibits a drop in accuracy as the model size increases. This drop is delayed for $b=32$ and removed for $b=256$. On the runtime side, MBS is faster than ILP and seems to have a similar scaling behavior when the number of attackers increases.

In contrast, ISI maintains perfect accuracy scores across all beam sizes and model sizes considered. For smaller beam sizes ($b=4$ and $b=32$), ISI is significantly faster than ILP and MBS with a sufficient beam size (only MBS with $b=256$ and $b=32$ maintains good accuracy scores). Conversely, ISI with a large beam size ($b=256$) scales very badly with the number of attackers and reaches the runtime of ILP and MBS with $b=256$ for a large number of attackers.

In conclusion, ISI outperforms the other approaches for small or medium beam sizes.

\subsection{Impact of LUCB in the Stochastic SMK Model}

Columns 3 and 4 of Figure \ref{fig:main-fig} report results for stochastic models. Column 3 uses our LUCB algorithm for oracle and heuristic evaluation, while Column 4 uses naive average estimators. The DICE scores remain similar across both approaches, while the number of model calls is consistently lower with LUCB.

Quantitatively, LUCB reduces the number of model calls by up to 37\%, with only minor losses in DICE score. We observe a few cases of DICE degradation (less than 3\%) and improvements up to 11\%. A detailed comparison between LUCB and naive estimators is reported in Appendix~\ref{an:algo-gain}.

\subsection{Ablation Study}
\label{sec:ablation-results}

\begin{figure}
    \centering
    \includegraphics[width=.5\linewidth]{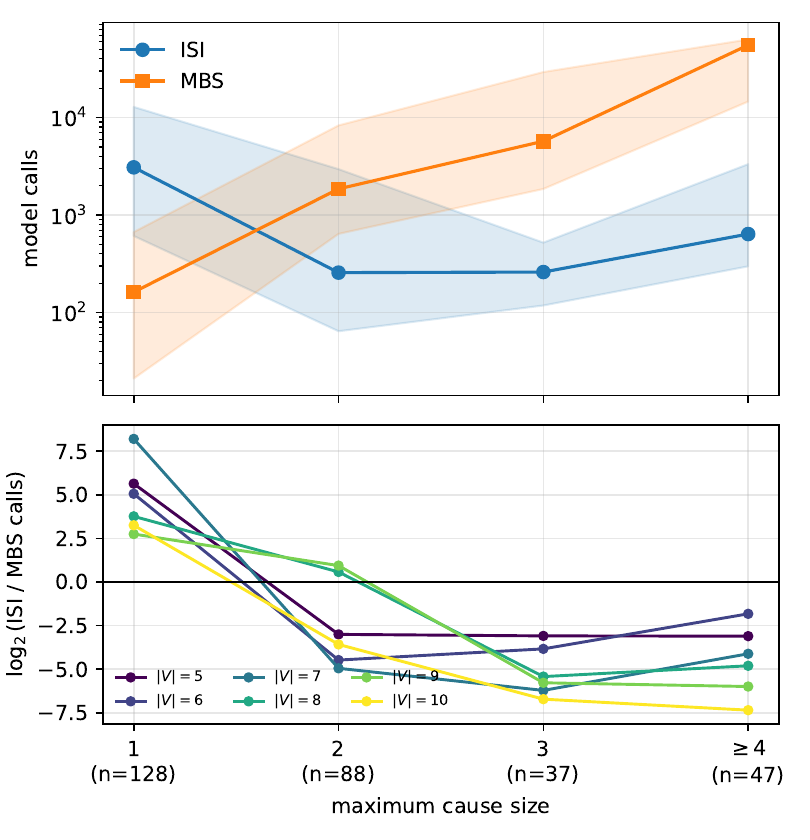}
    \caption{Cost of exact identification as a function of the largest cause of
    the model. Top: median number of calls, with the interquartile range over
    models. Bottom: the same comparison as a per-model ratio, with one line per
    number of variables, showing that the effect is not an artifact of graph size.
    Both algorithms return the exact set of causes on every model of the corpus,
    so the comparison bears on cost only.}
    \label{fig:cause-size}
\end{figure}

Table~\ref{tab:ablation} reports every component of both algorithms on the random corpus of Section~\ref{sec:random-corpus}. Each row is read against the reference configuration of its own family: the \emph{Ratio} column gives the median per-model number of calls relative to \emph{ISI exact} or \emph{MBS exact}. Figure~\ref{fig:cause-size} reveals
that ISI and MBS respond differently to the number of causes in the model. Since the corpus includes models with different cause sizes, the comparisons in Table~\ref{tab:ablation} are relevant between approximations of the same algorithm.

\input{tables/table-ablation}

\textit{The approximations trade quality for cost monotonically.} Restricting the beam or the number of steps degrades the result smoothly, which confirms, on random models, the trade-off that Section~\ref{subsec:result-bs-smk} establishes on SMK. The two approximations specific to ISI behave the same way: bounding the expansion depth and sampling the subproblems of each level smoothly degrade performance while reducing the number of model calls. Searching the subproblems with MBS costs more than the exact search when every solution is expanded, since a beam that does not enforce minimality returns many more solutions to expand.

\textit{The minimality mechanisms of MBS are important.} Starting the search with the actual values, expanding causes, and disabling superset pruning leaves the returned set exactly right  but significantly increases the number of calls.

\textit{Graph errors are not symmetric.} Adding spurious parents to the graph makes ISI search a superset of what it should, so it still recovers every cause, only using more model calls. Conversely, removing edges drops the solved rate and the number of calls. Adding edges is therefore a safe graph error for ISI, whereas removing them is not.

\textit{Block decomposition is neutral.} The block decomposition leaves both the result and the cost unchanged: both act on the free instance of a subproblem, and $|I|$ is small on random graphs, so there is nothing to decompose. This improvement is important only for graphs with large in-degrees but low edge density (one variable has many parents but two variables are unlikely to share a parent).

The \emph{naive} completion of a partial intervention scores better than the \emph{greedy} one on this corpus and costs fewer calls, since it assigns the remaining variables without evaluating anything. However, as Section~\ref{sec:isi} states, a naively completed intervention is not guaranteed to cancel the target. Notably, returned interventions under \emph{naive} assignments can be non-canceling, which can be seen as a worse type of mistake than non-minimal or missed ones.

Figure~\ref{fig:cause-size} conditions the comparison between the two exact algorithms on the size of the largest cause of each model. Additionally, since larger models are more likely to have larger causes, each model size is reported separately. MBS outperforms ISI for models with only singleton causes (and also slightly for models of size 8 and 9 with causes of size 2). This pattern occurs because MBS explores interventions by increasing size, so it must enumerate the combinations of variables up to the size of the cause it is looking for. Conversely, ISI searches the parent boundary of the variables it backtracks on, so its cost follows the shape of the graph and not the size of the causes; when every cause is a singleton, a flat beam finds all of them at its first level and the subproblem machinery of ISI is pure overhead.

\subsection{Stochastic Study}
\label{sec:results-stochastic}

\input{tables/random-graph-lucb-vs-naive}

Tables~\ref{tab:lucb-matched} and~\ref{tab:lucb-grid} report the stochastic study described in Section~\ref{sec:stochastic-setup}. The first compares the two estimators and the second varies the parameters of LUCB.

Table~\ref{tab:lucb-matched} shows that adaptive sampling pays off only when the fixed budget is generous. At $N = 200$, LUCB reaches the same cause sets as averaging for a fraction of the calls. At $N = 50$ the saving disappears entirely: allocating adaptively has a fixed overhead, and when the uniform budget is already tight there is nothing left to reallocate. The gain therefore depends on the sampling regime, not on the estimator alone: it is largest when the uniform budget is generous.

The saving shrinks as the noise level grows: the adaptive procedure stops sampling an intervention once its confidence interval falls on one side of the threshold, and noisier observations keep more interventions ambiguous for longer. When every intervention is borderline, the two estimators coincide.

Quality is mostly preserved. Almost all results in terms of proportion of solved models, DICE scores, and intervention estimation (last column) are similar. Even though there are some settings where both approaches slightly differ, there is no general tendency of one outperforming the other.

\input{tables/random-graphs-lucb-sensitivity}

Table~\ref{tab:lucb-grid} separates the two parameters of the adaptive procedure, namely the cancellation threshold and the confidence level. Results show that the cancellation threshold dominates the confidence level. On the one hand, reducing the confidence level $\delta$ systematically increases the number of calls. However, it only slightly increases the quality metrics. On the other hand, changing the cancellation threshold significantly impacts the quality metrics. The threshold has to be set relative to the noise level. The frequency of incorrect canceling decisions explains the collapse. It remains close to zero whenever the threshold is well placed, and rises in the two configurations where the solved rate drops. A misplaced threshold therefore degrades the result regardless of how accurately $\phi$ is estimated. In practice, the threshold should be chosen from the noise level of the system, and the algorithm gives no way to discover it on its own: this is a tuning requirement.

\section{Discussion} \label{sec:discussion}

\subsection{Limitations of the Proposed Algorithms}
\label{sec:discussion-algo-limitations}

A key practical limitation of our approach is the requirement of an oracle function, which must support counterfactual generalization \citep{scholkopfCausalRepresentationLearning2021}, i.e., the ability to evaluate outcomes under unseen or even unrealistic interventions. Learning such a function from data remains beyond the capabilities of current machine learning models. In physical or numerical systems, implementing interventions can also be costly or infeasible. In practice, the most natural setting for our approach is simulation-based systems, where interventions and counterfactual evaluations are readily available. Many formal models of real-world systems can also be automatically translated into the SCM formalism \citep{ibrahimActualCausalityCanvas2020}, further supporting the applicability of our method in these contexts.

We note, however, that this limitation should be understood relative to the requirements of HP causation itself. Methods grounded in the HP framework usually require a fully specified SCM, including structural equations and a causal graph. Our oracle is strictly less demanding: it requires only the ability to evaluate counterfactual outcomes, without access to the structural equations. Moreover, when using MBS alone (without ISI), no causal graph is needed either. In this sense, the oracle assumption is a relaxation of the full SCM requirement rather than an additional constraint.

Another limitation concerns the design of the heuristic function. 
Our experiments in Appendix~\ref{an:heuristics} suggest that heuristics that minimize the number of positive values in the counterfactual state (e.g., “sum-pos”) outperform neutral or counter-aligned ones, such as heuristics that minimize the number of variables that remain at their actual values (e.g., “Occam”). These considerations are valid because the system is composed of conjunctions and disjunctions with few negations. They indicate that the choice of heuristic matters and that effective proxies can often be derived from the structure of the system.
However, there are no general guidelines for constructing effective heuristics. This can limit usability in practice. 
Nevertheless, in many domains, natural heuristics arise from the semantics of the system. For example, \cite{reydRelevantCauses2025} use MBS to identify the cause of a flock of artificial agents colliding with an obstacle. In their simulation, agents that collide with the obstacle effectively merge with it; the distance to the obstacle’s center therefore serves as a heuristic, as agents farther from the center are closer to avoiding the collision and thus canceling the target outcome. This example illustrates how domain knowledge can inform heuristic design without requiring problem-specific algorithmic changes.

Our experiments in Appendix~\ref{an:heuristics} also reveal that designing a heuristic that prioritizes “good” causes can be challenging. The literature on actual causation has proposed many criteria for assessing the “quality” of actual causes \citep{halpernGradedCausationDefaults2015,chocklerResponsibilityBlameStructuralModel2004}. For instance, one might prefer causes that modify the system only slightly, as they represent a minimal departure from the actual situation. It would then seem natural to use the heuristic function to favor interventions with a limited effect on the system. However, our results show that this heuristic hurts the identification process: finding causes at all and selecting good causes can be orthogonal objectives.

Finally, our algorithms do not directly address continuous variable domains, which constitutes a major limitation. While discretization or sampling could be used in principle, this significantly reduces flexibility and may introduce additional approximation errors.

\subsection{Limitations of the Experimental Evaluation}

We acknowledge the limited scope of our experimental design. In particular, we do not compare against existing baselines for full HP-cause identification, as no general-purpose methods exist for this task. Similarly, our main evaluations rely on a single family of SCMs or on small SCMs with few structural constraints. While this choice is motivated by the absence of established benchmarks, it necessarily limits the generality of our empirical conclusions. We mitigate this limitation through controlled variations of the model (Boolean, non-Boolean, and stochastic), through sanity checks reported in Appendix~\ref{an:sanity}, and through an evaluation on synthetic SCMs reported in Appendix~\ref{an:sweep}.

The SMK benchmark also favors certain aspects of our approach, particularly ISI. 
The SMK models bound the size of non-minimal canceling interventions and restrict the number of values that cancel the outcome, limiting the combinatorial explosion that could otherwise arise. That said, our experiments on our small random corpus and on various graph structures in Appendix~\ref{an:sweep} give early results on how our algorithms and their approximations behave with various types of SCM. Notably, ISI outperforms MBS not only on the favorable SMK family but also on most random small graphs and most graph structures evaluated. The only notable exceptions are graphs with causes of size 1, graphs with large fixed in-degrees, wide \emph{layered} graphs, and \emph{shortcut-trees} with many reconvergences.

The LUCB algorithm introduces several additional parameters, which may require tuning in practice. Though this adds practical concerns, it also provides explicit control over the precision–efficiency trade-off. A systematic study of the influence of these parameters, beyond the one currently provided, is out of the scope of this paper.

Finally, we report only a preliminary complexity analysis of MBS and ISI in Appendix~\ref{an:complexity-analysis}. This complexity depends on properties of the identified causes, such as their number and size, which are difficult to characterize without carefully designed SCMs. A more comprehensive empirical complexity analysis is therefore left for future work.

\subsection{Limitations of Actual-Cause Definitions}

The HP definition of actual causation has well-known conceptual limitations, including the lack of explicit consideration for responsibility \citep{chocklerResponsibilityBlameStructuralModel2004}, normality \citep{halpernGradedCausationDefaults2015}, and contrastiveness \citep{millerContrastiveExplanationStructuralmodel2021}. In this paper, we focus on HP-causes because they constitute a widely accepted baseline and serve as the foundation for many alternative definitions.

Our algorithms assume that the first condition (AC1) is verified by the user and that the minimality condition (AC3) is approximately enforced by the search procedure. The second condition (AC2) is handled through the oracle function. While this reliance on an oracle is a limitation, it also provides flexibility: many extensions of the HP definition modify AC2, for instance to account for normality or causal sufficiency \citep{beckersCausalSufficiencyActual2021}. Treating the oracle as an input therefore allows our algorithms to adapt naturally to alternative definitions of actual causation, which is valuable given the lack of consensus on a definitive formalization.

\subsection{Future Work}

Future work will address several directions. First, we will investigate adaptations of our algorithms to continuous domains. Second, we aim to provide theoretical guarantees on approximation quality as a function of the heuristic and to develop practical guidelines for heuristic design. Third, we would like to design a comprehensive benchmark for the task of full HP-cause identification. The generalization of our results is limited because such a benchmark does not exist yet. Fourth, we would like to test our algorithms on more realistic use cases. To the best of our knowledge, no real system has been formalized as an SCM with known actual causes. Designing such a model would contribute a benchmark for the HP-cause identification task. Finally, and most importantly, we plan to integrate our algorithms into complete explanation frameworks, including causal discovery in real systems, oracle design aligned with refined notions of actual causation, and relevance-based filtering to produce human-interpretable explanations.

\section{Conclusion}

Actual causes play a central role in explainable artificial intelligence, as they align with users’ expectations for explanations grounded in specific situations rather than general tendencies. Despite decades of work on formal definitions, identifying actual causes in practice remains challenging, and existing methods are scarce, restricted to narrow model classes, or limited to the smallest causes.

In this paper, we propose a step towards full HP-cause identification that extends beyond deterministic Boolean systems to non-Boolean and stochastic models. Our baseline algorithm provides an explicit and controllable trade-off between identification quality and computational cost. Building on this foundation, we proposed a complementary algorithm, which exploits structural properties of the system to significantly improve both accuracy and efficiency for most types of SCMs, and we formally proved its correctness when used in its exact version. Finally, we introduced another complementary algorithm that improves sample efficiency in stochastic settings.

We evaluated our approach on a Boolean SCM from the literature and on non-Boolean and stochastic variants. Our results demonstrate that the beam size parameter effectively controls the quality–efficiency trade-off, that ISI usually substantially improves full cause identification compared to the baseline, that ISI outperforms existing approaches for smallest-cause identification when applicable, and that LUCB reduces the number of model calls in stochastic models. We further complemented this analysis with various types of SCMs, which revealed that these conclusions generalize in most settings but with notable exceptions, notably for SCMs with only causes of size 1.

While our algorithms are not sufficient on their own to produce user-facing explanations (or to fully solve the task of HP-cause identification without additional assumptions, such as the heuristic function) and must be combined with complementary methods for relevance filtering and model construction, they address a major practical gap in the causality and XAI literature. By making actual cause identification computationally accessible across a broader class of systems, this work provides a concrete step toward more user-centric explanations and facilitates the broader adoption of actual causation in XAI research and applications.

\section*{Acknowledgements}

The authors would like to thank Louis Jachiet for carefully reading Appendix \ref{an:proofs} and providing helpful feedback that improved the clarity of the exposition and helped identify some notational issues. We also thank Nikola Simidjievski for his guidance throughout the publication process of this work.

\bibliography{main}
\bibliographystyle{tmlr}

\newpage

\appendix
\counterwithin{figure}{section}

\input{annexes/illustrations}

\input{annexes/sanity}

\input{annexes/impact}

\input{annexes/structure_sweep}

\input{annexes/complexity}

\input{annexes/heuristics}

\input{annexes/proofs}

\end{document}

%% file: tables/causality-table.tex
\begin{tabular}{|c|c|c|}
        \hline
        \multicolumn{3}{|c|}{\textbf{Causation: Why did this happen?}} \\\hline
         & \textbf{General Causation} & \textbf{Actual Causation} \\\hline\hline
        Formulation & What \textbf{factors} contributed to this?& What \textbf{facts} caused this? \\\hline
        Example & Smoking causes cancer. & Their smoking caused their cancer. \\\hline
        \multirow{3}{*}{Definition}  & \cellcolor{LimeGreen}Bayesian networks & \cellcolor{Dandelion}HP-causes \\\cline{2-3}
         & \cellcolor{LimeGreen}Mainly accepted & \cellcolor{Dandelion}Still under debate \\\cline{2-3}
         & \cellcolor{LimeGreen}Large literature & \cellcolor{Dandelion}Some literature \\\hline
        \multirow{3}{*}{Identification} & \cellcolor{Dandelion}Causal discovery & \cellcolor{Bittersweet}No unified field \\\cline{2-3}
         & \cellcolor{Dandelion}No working solution & \cellcolor{Bittersweet}No practical solution \\\cline{2-3}
         & \cellcolor{Dandelion}Huge literature & \cellcolor{Bittersweet}Little literature \\\hline
    \end{tabular}

%% file: algorithms/mbs.tex
\begin{algorithm}[ht]
    \SetKwData{MaxSteps}{max\_steps}
    \SetKwData{EarlyStop}{early\_stop}
    \SetKwData{pos}{pos}
    \SetKwData{neg}{neg}
    
    \SetKwFunction{ExpandBeam}{expandBeam}
    \SetKwFunction{SplitRules}{splitEval}
    \SetKwFunction{FilterMinimality}{filterMinimality}
    \SetKwFunction{getBest}{getBest}
    
    \SetKwInOut{Input}{Input}
    \SetKwInOut{Output}{Output}
    
    \Input{$\vars$, $v^*$, $\dom$, $\phi$ (includes the target $\target$), $\psi$, beam size $b$, \EarlyStop, \MaxSteps}
    \Output{List of all causes}
    
    \BlankLine
    
    $\mathbf{C}\gets\emptyset$ \tcp*{Identified causes}
    $B\gets\emptyset$ \tcp*{Beam made of interventions $e$}
    
    \BlankLine
    
    \For{$t \gets 1$ \KwTo \MaxSteps}{
        $B\gets$ \ExpandBeam{$B$, $\vars$, $\dom$, $v^*$, $\mathbf{C}$}\tcp*{Interventions $e$ with $Ctf(e)$ minimal with $\mathbf{C}$}
        \If{$B=\emptyset$ or $($\EarlyStop $\&$  $\mathbf{C}\neq\emptyset$)}{break\;}
        
        \neg, \pos $\gets$ \SplitRules{$\phi(B)$}\tcp*{Distinguish $\phi(e)=1$ from $\phi(e)=0$}
        
        $\mathbf{C}\gets \mathbf{C}\ \cup$\ \FilterMinimality{\neg, $\mathbf{C}$, $v^*$}\;
        $B\gets$ \getBest{$\psi($\pos$)$, $b$, $\mathbf{C}$}\tcp*{Keep $b$ best for $\psi$ with $Ctf(e)$ minimal in $\mathbf{C}$}
    }
    \BlankLine
    \Return $\mathbf{C}$\;
    \BlankLine
    
    \caption{MBS Algorithm}
    \label{algo:beam-search}
\end{algorithm}

%% file: figures/proof-graphs.tex
\begin{figure}
    \centering
    \begin{subfigure}[b]{0.15\textwidth}
        \centering
        \begin{tikzpicture}[node distance=.4cm]
            
            \node (X) {$X$};
            \node[above=of X] (A) {$A$};
            \node[right=of A] (B) {$B$};
            \node[above=.6cm of $(A)!0.5!(B)$] (T) {$T$};
            
            \draw[->] (X) -- (A);
            \draw[->] (A) -- (B);
            \draw[->] (A) -- (T);
            \draw[->] (B) -- (T);
            
            \end{tikzpicture}
        \caption{Not-graph \\$T=A\lor B$, \\$A=X$, \\$B=\neg A$, \\$X=1$.}
        \label{fig:not_graph}
    \end{subfigure}
    \hfill
    \begin{subfigure}[b]{0.13\textwidth}
        \centering
        \begin{tikzpicture}[node distance=.4cm]
            \node (X) {$X$};
            \node[above left=of X] (A) {$A$};
            \node[above right=of X] (B) {$B$};
            \node[above=.6cm of $(A)!0.5!(B)$] (T) {$T$};
            
            \draw[->] (X) -- (A);
            \draw[->] (X) -- (B);
            \draw[->] (A) -- (T);
            \draw[->] (B) -- (T);
            
            \end{tikzpicture}
        \caption{Xor graph: \\$T=A\oplus B$, \\$A=X$, \\$B=\neg X$, \\$X=1$.}
        \label{fig:xor_graph}
    \end{subfigure}
    \hfill
    \begin{subfigure}[b]{0.12\textwidth}
        \centering
        \begin{tikzpicture}[node distance=.4cm]
            
            \node (X) {$X$};
            \node[right=of X] (Y) {$Y$};
            \node[above=of X] (A) {$A$};
            \node[right=of A] (B) {$B$};
            \node[above=.6cm of $(A)!0.5!(B)$] (T) {$T$};
            
            \draw[->] (X) -- (A);
            \draw[->] (Y) -- (B);
            \draw[->] (A) -- (T);
            \draw[->] (B) -- (T);
            
            \end{tikzpicture}
        \caption{Or-graph: \\$T=A\lor B$, \\$A=X$, \\$B=Y$, \\$X=1$,\\$Y=1$.}
        \label{fig:or_graph}
    \end{subfigure}
    \hfill
    \begin{subfigure}[b]{0.45\textwidth}
        \centering
        \begin{tikzpicture}[node distance=.3cm]
        
        \node (T) {$T$};
        \node[below left= of T] (A) {$A$};
        \node[below right= of T] (B) {$B$};
        \node[below left= of A] (X) {$X$};
        \node[below right= of A, xshift=-4mm] (G) {$G$}; 
        \node[below left= of B, xshift=4mm] (Y) {$Y$}; 
        \node[below right= of B] (H) {$H$};
        
        \draw[->] (A) -- (T);
        \draw[->] (B) -- (T);
        \draw[->] (X) -- (A);
        \draw[->] (G) -- (A);
        \draw[->] (Y) -- (B);
        \draw[->] (H) -- (B);
        
        \end{tikzpicture}
        \caption{“Split”-graph: 
        \\$T=\lnot\big((A= 0)\lor (B= 0)\lor(A=g\land B=h)\big)$, 
        \\$A=\begin{cases}
            g\ \text{if}\ G\\ 
            x\ \text{if}\ \lnot G\land X\\ 
            0\ \text{if}\ \lnot G\land\lnot X
        \end{cases}$, 
        $B=\begin{cases}
            h\ \text{if}\ H\\ 
            y\ \text{if}\ \lnot H\land Y\\ 
            0\ \text{if}\ \lnot H\land\lnot Y
        \end{cases}$, \\
        $X=1$, $G=0$, $Y=1$, $H=0$}
        \label{fig:split_graph}
    \end{subfigure}
    \caption{Four SCMs to understand how ISI “backtracks” from an identified cause.}
    \label{fig:proof-graphs}
\end{figure}

%% file: algorithms/isi.tex
\begin{algorithm}[t]
    \SetAlgoLined
    \SetKwFunction{minimalUpdate}{minimalUpdate}
    \SetKwFunction{search}{search}

    \SetKwInOut{Input}{Input}
    \SetKwInOut{Output}{Output}

    \Input{$\vars$, DAG over $\vars$, \search an algorithm that finds “canceling” counterfactual interventions assigning every variable of $I$ (with a witness)}
    \Output{List of all causes}

    \BlankLine

    $\mathbf{C}$ $\gets\emptyset$\;
    $Q\gets\emptyset$\;
    $Q$.queue($(\{\target\},\Pa(\target),\emptyset,\emptyset,\emptyset)$)\;
    \BlankLine

    \While{$Q\neq\emptyset$}{
        $(L,I,R_C,r_C,R_W)\gets Q$.dequeue()\;

        \ForEach{$(C,c, W)\in$\search{$L,I,R_C,r_C,R_W$}}{
            $\mathbf{C}\gets$\minimalUpdate{$\mathbf{C}$, $C$}\;

            \ForEach{$S\in2^{C\cap I}\setminus\emptyset$}{
                $L'\gets L\cup S$\;
                $I'\gets\Pa(S)\setminus(C\cup W\cup L')$\;
                $R_C'\gets C\setminus S$\;
                $r_C'\gets c_{|R_C'}$\;
                $R_W'\gets W$\;
                $Q$.add($(L',I',R_C',r_C',R_W')$)\;

            }
        }
    }
    \BlankLine
    \Return $\mathbf{C}$\;
    \BlankLine
    \caption{ISI algorithm}
    \label{algo:ISI}
\end{algorithm}

%% file: algorithms/lucb.tex
\begin{algorithm}
    \SetAlgoLined
    \SetKwFunction{sample}{sample}
    \SetKwFunction{search}{search}
    \SetKwFunction{findBorderline}{findBorderline}
    \SetKwFunction{findUnsure}{findUnsure}
    \SetKwFunction{computeBounds}{computeBounds}

    \SetKwInOut{Input}{Input}
    \SetKwInOut{Output}{Output}
    
    \Input{Beam size $b$, A set of interventions $E$, Cancellation threshold $a$, Slacks $\varepsilon_b$, $\varepsilon_c$, $\varepsilon_{nc}$, Oracle $\phi$, Heuristic $\psi$, \texttt{init\_batch\_size}, \texttt{batch\_size}, average budget per arm $N$}
    \Output{Estimates $\widehat{\phi}(E)$, $\widehat{\psi}(E)$}
    \BlankLine
    $\widehat{\phi}(E),\widehat{\psi}(E),s^2(E),n(E)\gets$\sample{$E$, \texttt{init\_batch\_size}, $\phi$, $\psi$}\;
    $L_\phi(E),L_\psi(E)\gets0$\;
    $U_\phi(E),U_\psi(E)\gets1$\;
    $c, nc, u, \ell\gets$\findBorderline{$L_\phi(E),U_\phi(E),L_\psi(E),U_\psi(E)$}\;
    \While{$U_\phi(c) \ge a + \varepsilon_{c}$ or $L_\phi(nc) \le a - \varepsilon_{nc}$ or $U_\psi(u) - L_\psi(\ell)> \varepsilon_b$}{
        $S\gets$\findUnsure{$L_\phi(E),U_\phi(E),L_\psi(E),U_\psi(E)$}\;
        $\widehat{\phi}(S),\widehat{\psi}(S),s^2(S),n(S)\gets$\sample{$S$, \texttt{batch\_size}, $\phi$, $\psi$}\;
        $L_\phi(S),U_\phi(S),L_\psi(S),U_\psi(S)\gets$\computeBounds{$\widehat{\phi}(S),\widehat{\psi}(S),s^2(S),n(S)$}\;
        $c, nc, u, \ell\gets$\findBorderline{$L_\phi(E),U_\phi(E),L_\psi(E),U_\psi(E)$}\;
        \If{$\sum_{e\in E}n(e)>N\times|E|$}{break\;}
    }

    \BlankLine
    \Return $\widehat{\phi}(E)$, $\widehat{\psi}(E)$\;
    \BlankLine
    \caption{LUCB algorithm}
    \label{algo:lucb}
\end{algorithm}

%% file: tables/table-ablation.tex
\begin{table}[t]
\centering
\small
\begin{tabular}{llrrrr}
\toprule
Configuration & Family & Solved & DICE & Calls & Ratio \\
\midrule
ISI exact & exact & 1.00 & 1.00 & 938 & -- \\
MBS exact & exact & 1.00 & 1.00 & 1342 & -- \\
\addlinespace
ISI exact, no decomposition & isi design & 1.00 & 1.00 & 878 & 1.00 \\
\addlinespace
ISI exact, max\_backtrack=1 & isi budget & 0.28 & 0.65 & 144 & 0.27 \\
ISI exact, max\_backtrack=2 & isi budget & 0.87 & 0.95 & 702 & 1.00 \\
ISI exact, max\_backtrack=3 & isi budget & 1.00 & 1.00 & 938 & 1.00 \\
ISI exact, sample=1 & isi budget & 0.34 & 0.69 & 149 & 0.34 \\
ISI exact, sample=4 & isi budget & 0.45 & 0.76 & 172 & 0.43 \\
ISI exact, sample=16 & isi budget & 0.61 & 0.86 & 258 & 0.69 \\
ISI exact, sample=64 & isi budget & 0.78 & 0.94 & 364 & 0.98 \\
\addlinespace
ISI + MBS(-1,-1) minimal & isi+mbs & 0.63 & 0.93 & 214 & 0.74 \\
ISI + MBS(-1,-1) all & isi+mbs & 0.96 & 0.99 & 3134 & 2.53 \\
ISI + MBS(8,5), greedy assign & isi+mbs & 0.37 & 0.65 & 118 & 0.14 \\
ISI + MBS(8,5), naive assign & isi+mbs & 0.63 & 0.70 & 106 & 0.11 \\
\addlinespace
MBS beam=4 & mbs beam & 0.59 & 0.66 & 148 & 0.11 \\
MBS beam=16 & mbs beam & 0.68 & 0.72 & 374 & 0.29 \\
MBS beam=256 & mbs beam & 0.83 & 0.84 & 1331 & 0.99 \\
MBS(8,5) & mbs beam & 0.63 & 0.70 & 200 & 0.14 \\
\addlinespace
MBS steps=2 & mbs steps & 0.56 & 0.67 & 70 & 0.06 \\
MBS steps=3 & mbs steps & 0.76 & 0.79 & 250 & 0.19 \\
MBS steps=5 & mbs steps & 0.95 & 0.96 & 935 & 0.76 \\
\addlinespace
MBS, actual values in init & mbs mechanism & 1.00 & 1.00 & 1700 & 1.15 \\
MBS, no superset pruning & mbs mechanism & 1.00 & 1.00 & 13464 & 3.09 \\
MBS, no pruning, expand causes & mbs mechanism & 1.00 & 1.00 & 16256 & 3.72 \\
\addlinespace
ISI exact, +1 spurious parent & misspecified & 1.00 & 1.00 & 1256 & 1.77 \\
ISI exact, +2 spurious parents & misspecified & 1.00 & 1.00 & 2342 & 2.38 \\
ISI exact, -1 missing edge & misspecified & 0.40 & 0.61 & 256 & 0.35 \\
ISI exact, -2 missing edges & misspecified & 0.14 & 0.44 & 58 & 0.10 \\
\bottomrule
\end{tabular}
\caption{Ablation on the mixed random corpus. Solved and DICE are means over models, costs are medians. \emph{Ratio} is the median per-model number of calls relative to the reference configuration of the same family (\emph{ISI exact} or \emph{MBS exact}).}
\label{tab:ablation}
\end{table}

%% file: tables/random-graph-lucb-vs-naive.tex
\begin{table}[t]
\centering
\small
\centering
\begin{tabular}{llrrrrrr}
\toprule
$\epsilon_n$ & $N$ & Estimator & Solved & DICE & Calls ($10^3$) & Ratio & Err. \\
\midrule
\multirow{4}{*}{0.02} & \multirow{2}{*}{50}  & AVG  & 0.72 & 0.77 & 11.7 & 1.00          & 0.000 \\
                      &                      & LUCB & 0.71 & 0.76 & 11.3 & \textbf{0.96} & 0.000 \\
\cmidrule(lr){2-8}
                      & \multirow{2}{*}{200} & AVG  & 0.76 & 0.80 & 46.0 & 1.00          & 0.000 \\
                      &                      & LUCB & 0.72 & 0.76 & 15.1 & \textbf{0.33} & 0.000 \\
\midrule
\multirow{4}{*}{0.05} & \multirow{2}{*}{50}  & AVG  & 0.74 & 0.78 & 11.7 & 1.00          & 0.002 \\
                      &                      & LUCB & 0.75 & 0.79 & 12.4 & \textbf{1.06} & 0.001 \\
\cmidrule(lr){2-8}
                      & \multirow{2}{*}{200} & AVG  & 0.77 & 0.81 & 45.5 & 1.00          & 0.000 \\
                      &                      & LUCB & 0.77 & 0.81 & 28.3 & \textbf{0.62} & 0.000 \\
\midrule
\multirow{4}{*}{0.10} & \multirow{2}{*}{50}  & AVG  & 0.68 & 0.76 & 11.7 & 1.00          & 0.045 \\
                      &                      & LUCB & 0.68 & 0.76 & 13.1 & \textbf{1.12} & 0.045 \\
\cmidrule(lr){2-8}
                      & \multirow{2}{*}{200} & AVG  & 0.71 & 0.79 & 45.6 & 1.00          & 0.050 \\
                      &                      & LUCB & 0.70 & 0.79 & 38.6 & \textbf{0.85} & 0.054 \\
\bottomrule
\end{tabular}
\caption{Adaptive sampling against fixed-size averaging, at the reference threshold $a=0.30$ and confidence level $\delta=0.10$. Each pair of rows compares the two estimators at the same noise level and sampling budget. The \emph{Ratio} column shows what LUCB gains or loses in model calls compared to the naive approach.}
\label{tab:lucb-matched}
\end{table}

%% file: tables/random-graphs-lucb-sensitivity.tex
\begin{table}
\small
\centering
\begin{tabular}{lllrrrrr}
\toprule
$\epsilon_n$ & $a$ & $\delta$ & Solved & DICE & Calls ($10^3$) & Ratio & Err. \\
\midrule
\multirow{4}{*}{0.02} & \multirow{2}{*}{0.15} & 0.05 & 0.74 & 0.78 & 18.2 & 0.40 & 0.000 \\
                      &                       & 0.10 & 0.72 & 0.76 & 15.4 & 0.34 & 0.000 \\
\cmidrule(lr){2-8}
                      & \multirow{2}{*}{0.30} & 0.05 & 0.73 & 0.77 & 17.7 & 0.39 & 0.000 \\
                      &                       & 0.10 & 0.72 & 0.76 & 15.1 & 0.33 & 0.000 \\
\midrule
\multirow{4}{*}{0.05} & \multirow{2}{*}{0.15} & 0.05 & 0.64 & 0.76 & 34.8 & 0.77 & \textbf{0.090} \\
                      &                       & 0.10 & 0.63 & 0.76 & 30.1 & 0.66 & \textbf{0.086} \\
\cmidrule(lr){2-8}
                      & \multirow{2}{*}{0.30} & 0.05 & 0.79 & 0.82 & 33.5 & 0.74 & 0.000 \\
                      &                       & 0.10 & 0.77 & 0.81 & 28.3 & 0.62 & 0.000 \\
\midrule
\multirow{4}{*}{0.10} & \multirow{2}{*}{0.15} & 0.05 & 0.38 & 0.58 & 47.9 & 1.05 & \textbf{0.241} \\
                      &                       & 0.10 & 0.37 & 0.57 & 41.6 & 0.91 & \textbf{0.240} \\
\cmidrule(lr){2-8}
                      & \multirow{2}{*}{0.30} & 0.05 & 0.71 & 0.80 & 43.7 & 0.96 & 0.054 \\
                      &                       & 0.10 & 0.70 & 0.79 & 38.6 & 0.85 & 0.054 \\
\bottomrule
\end{tabular}
\caption{Sensitivity of the adaptive estimator to the canceling threshold $a$ and the confidence level $\delta$, at a budget of $N=200$. \emph{Err.} reports the fraction of evaluated interventions for which the method estimated a wrong oracle value.}
\label{tab:lucb-grid}
\end{table}

%% file: annexes/illustrations.tex
\section{Full Illustration of our Algorithms} \label{an:illustrations}

This appendix provides a pedagogical, step-by-step illustration of the operational behavior of our algorithms. It is intended to support intuition and clarify how the procedures unfold in practice; it is not required for understanding the theoretical contributions or the experimental results.

\subsection{MBS Illustration on Example~\ref{example:Suzy}}

\begin{figure*}[ht]
    \centering
    \includegraphics[width=\linewidth]{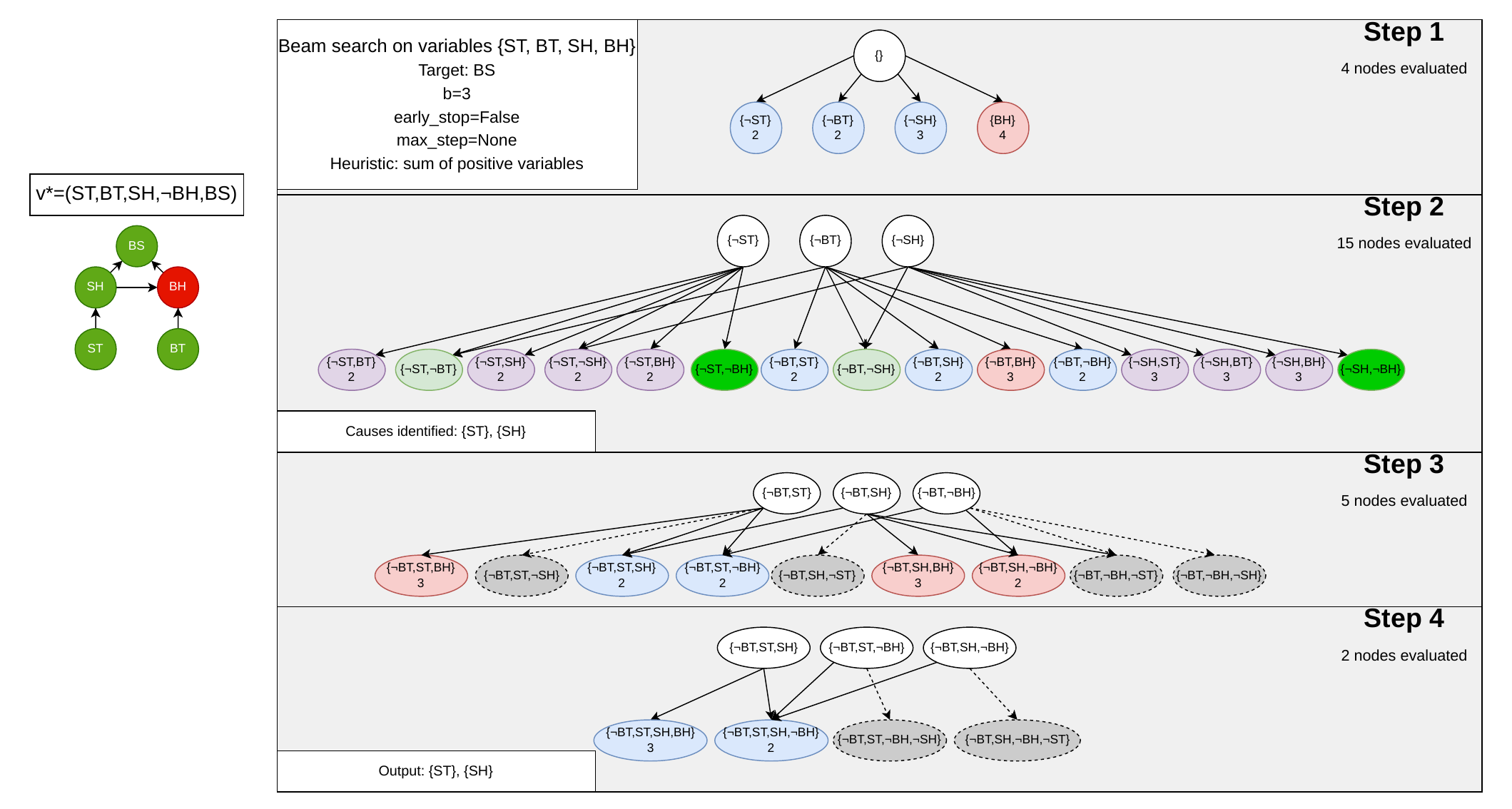}
    \caption{Illustration of our algorithm on Example \ref{example:Suzy}. Blue nodes constitute the beam, i.e., the nodes selected for expansion. Red nodes have less optimal heuristic values and are not part of the beam. Green nodes ‘cancel’ the consequence; light green ones are not minimal, and dark green ones are the identified causes. Purple nodes are evaluated, but filtered because they are supersets of a cause identified at the same depth. Nodes in gray (with dotted lines) are represented for illustration purposes, but are not evaluated.}
    \label{figure:example-bs}
\end{figure*}

We present a complete execution of the MBS algorithm on Example \ref{example:Suzy}, schematized in Figure \ref{figure:example-bs}. The causal graph is shown on the left-hand side of the figure. We run the algorithm without early stopping and without a maximum number of steps. The heuristic function $\psi$ minimizes the number of observed positive variables, and the beam size is set to $b = 3$. Each node in the search tree corresponds to an intervention, represented as a set of variable–value assignments. We use the compact notation introduced in Section \ref{sec:mbs}, where a variable appears with or without the symbol $\neg$ depending on its assigned value.

\paragraph{Step 1.} The first level of the search tree includes all singleton interventions, each setting one variable to its counterfactual value. None of these interventions cancels the consequence. All nodes are therefore scored using $\psi$, and the three best ones are retained in the beam (shown in blue in Figure \ref{figure:example-bs}). The remaining node is discarded (shown in red).

\paragraph{Step 2.} Each node in the beam is expanded. The expansion of $BH$ is shown with dotted lines to highlight that the algorithm does not build or evaluate these children. Generated nodes are evaluated; four interventions cancel the consequence and are identified as candidate causes (shown in green). Only the minimal ones are kept (dark green), while non-minimal causes (light green) are discarded. All nodes that are supersets of the identified causes are then pruned. The remaining nodes are scored, and the three best are retained.

\paragraph{Step 3.} The beam nodes are expanded again. Children that are supersets of already identified causes are not expanded further; these pruned expansions are shown in grey with dotted edges. After evaluation and scoring, the three best nodes are kept.

\paragraph{Step 4.} At this stage, two nodes are supersets of identified causes and are therefore ignored. The remaining two nodes are evaluated, but they already include all variables and cannot be expanded further. The algorithm terminates.

The final output consists of two causes of size 1, namely ${\mathit{ST}}$ and ${\mathit{SH}}$. For each cause, the algorithm also returns the associated witness set, which in both cases is ${\mathit{BH}}$.

\subsection{ISI Algorithm Illustration}

\begin{figure*}
    \centering
    \includegraphics[width=.8\linewidth]{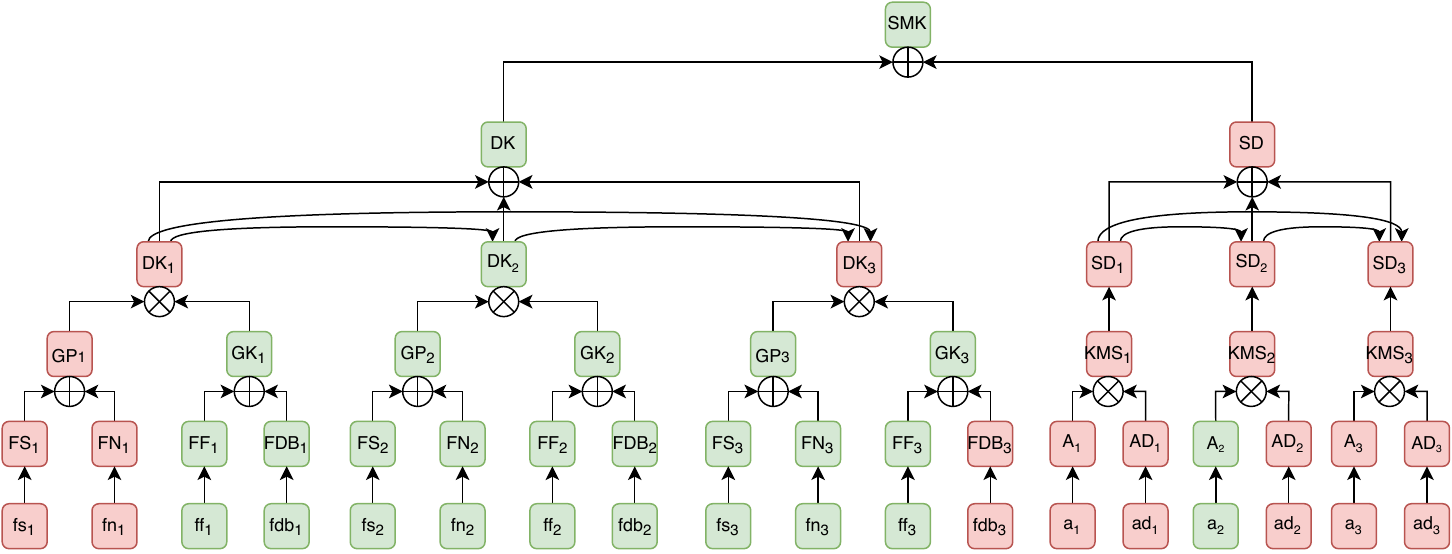}
    \caption{An instance of the SMK scenario. Green nodes have value 1 and red ones have value 0. Nodes with lowercase letters are exogenous variables.}
    \label{figure:SMK-instance}
\end{figure*}

We illustrate the behavior of ISI on the instance shown in Figure \ref{figure:SMK-instance} by first recalling how the base MBS algorithm behaves on this example, and then showing how ISI restructures the search to obtain the same causes more efficiently.

\paragraph{Behavior of MBS on this instance.}

As a baseline, we run MBS with a large beam size and a sufficiently large maximum number of steps so that no pruning is induced by these parameters. The heuristic $\psi$ again minimizes the number of positive variables.

MBS initializes the search with single-variable interventions and identifies one cause: $\{DK\}$. It expands the remaining candidates into interventions of size two, excluding $\{DK\}$. MBS then identifies $\{DK_2\}$, $\{GK_2\}$, and $\{GP_2\}$, all with witness $\{DK_3\}$ at step 2. Finally, it identifies $\{FS_2, FN_2\}$ and $\{FF_2, FDB_2\}$, again with witness $\{DK_3\}$, at step 3. MBS does not identify any cause during steps 4-6 and terminates.

This run illustrates that MBS is able to identify all HP-causes in this instance, but only after exploring a large portion of the intervention space.

\paragraph{Behavior of ISI on the same instance.}

We now show how ISI decomposes this same search into structured subproblems. For clarity, we report only the inputs and outputs of each MBS call invoked by ISI, without detailing the internal beam-search steps. We also omit the subproblems whose free instance $I$ is empty: they have nothing to search, they cost a single evaluation of their forced intervention, and on this instance none of them yields a cause that has not already been found elsewhere. Recall that a solution assigns every variable of $I$, so that the variables kept at their actual values appear explicitly in the witness.

\textit{Step 1.} ISI starts by calling MBS on the parents of $SMK$, namely $I=\{DK, SD\}$. The only assignment of $I$ that cancels $SMK$ moves $DK$ and holds $SD$ at its actual value, so this call identifies the cause $\{DK\}$ with witness $\{SD\}$ (note that MBS does not return $\{SD\}$ as a witness; it must be assigned either by the \emph{naive} or \emph{greedy} strategy). “Causally backtracking” with $S=\{DK\}$, ISI enqueues the parents of $DK$, that is, $I'=\{DK_1, DK_2, DK_3\}$, with $SD$ as forced witness.

\textit{Step 2.} ISI runs MBS on $\{DK_1, DK_2, DK_3\}$ with witness $\{SD\}$ and identifies $\{DK_2\}$ with witness $\{DK_1, DK_3, SD\}$. “Causally backtracking” with $S=\{DK_2\}$, ISI enqueues the parents of $DK_2$, namely $\{GP_2, GK_2\}$, together with the witness.

\textit{Step 3.} ISI runs MBS on $\{GP_2, GK_2\}$ with witness $\{DK_1, DK_3, SD\}$. Three assignments of the instance cancel $SMK$: moving $GP_2$ alone, which identifies $\{GP_2\}$ with witness $\{DK_1, DK_3, GK_2, SD\}$; moving $GK_2$ alone, which identifies $\{GK_2\}$ with witness $\{DK_1, DK_3, GP_2, SD\}$; and moving both, which identifies the non-minimal “canceling” intervention $\{GP_2, GK_2\}$ with witness $\{DK_1, DK_3, SD\}$. ISI backtracks from each of them, along every non-empty subset of the variables they move. Steps 4 and 5 follow the two minimal ones; Step 6 collects the branches opened by the third.

\textit{Step 4.} From $\{GP_2\}$, ISI runs MBS on $\{FS_2, FN_2\}$ with witness $\{DK_1, DK_3, GK_2, SD\}$ and identifies $\{FS_2, FN_2\}$ with the same witness. ISI still backtracks along each non-empty subset of $\{FS_2, FN_2\}$, but $FS_2$ and $FN_2$ have no parents, so all the subproblems enqueued here have an empty free instance.

\textit{Step 5.} From $\{GK_2\}$, ISI runs MBS on $\{FF_2, FDB_2\}$ with witness $\{DK_1, DK_3, GP_2, SD\}$ and identifies $\{FF_2, FDB_2\}$ with the same witness. Once again, only subproblems with an empty free instance are enqueued.

\textit{Step 6.} From the non-minimal intervention $\{GP_2, GK_2\}$, ISI backtracks along its three non-empty subsets. Taking $S=\{GP_2\}$ leaves $GK_2$ as a forced counterfactual intervention and searches $\{FS_2, FN_2\}$; taking $S=\{GK_2\}$ leaves $GP_2$ forced and searches $\{FF_2, FDB_2\}$; taking $S=\{GP_2, GK_2\}$ replaces both by their parents at once and searches $\{FF_2, FDB_2, FS_2, FN_2\}$. The first two return only supersets of $\{GK_2\}$ and of $\{GP_2\}$, and the third returns $\{FS_2, FN_2\}$ and $\{FF_2, FDB_2\}$ together with supersets of them. All of these are discarded by the minimality filter or were already identified at Steps 3 to 5.

\textit{Termination.} Once all the problems with empty instances have been tested and discarded as they reveal no new causes, the queue is now empty. ISI has generated $44$ subproblems, of which $8$ have a non-empty free instance and are the ones reported above; the remaining $36$ have an empty free instance and only restate an already identified pair. ISI identified the six causes $\{DK\}$, $\{DK_2\}$, $\{GP_2\}$, $\{GK_2\}$, $\{FS_2, FN_2\}$ and $\{FF_2, FDB_2\}$, each under the witnesses reported above.

In this instance, the branches opened at Step~6 return nothing that Steps~3 to~5 had not already produced, which is what motivates the approximation of Section~\ref{sec:isi} that backtracks only from the minimal “canceling” interventions of each subproblem. That approximation is not complete in general (as shown by Figure~\ref{fig:split_graph}), so the branches of Step~6 cannot be dispensed with without giving up our theoretical guarantees.

%% file: annexes/sanity.tex
\section{Sanity Check}\label{an:sanity}

This appendix reports sanity checks on classic examples from the actual causation literature. Specifically, we used the relevant examples used by \cite{halpernModificationHalpernPearlDefinition2015} as illustrations and means of analysis of Definition~\ref{def:hp-cause}. These examples are meant to illustrate the behavior of the actual cause definitions rather than to assess scalability or performance. Accordingly, we only aim to verify that the exact MBS and ISI generate the expected causes. Table~\ref{tab:sanity} summarizes the expected causes and the outcomes of the identification process.

While this is not the objective of this section, we can still note that with these small models with flat structure, the ISI algorithm no longer seems faster than MBS (being even slower in some cases). ISI takes advantage of the structure of the graph to reduce the computational complexity; when this structure is not advantageous, or when there are so few variables that the computational complexity does not matter, ISI no longer outperforms MBS. Notably, for the “Ranch extended scenario”, ISI subproblem creation creates a runtime overhead despite running fewer model calls than MBS. However, all runtimes remain comfortably below 1s.

\paragraph{Forest fire \citep{halpernCausesExplanationsStructuralmodel2001}.} 

This example contrasts conjunctive and disjunctive causation. A forest fire $FF$ may be caused by lightning $L$ or a dropped match $MD$. In the disjunctive case, $FF := L \lor MD$, both $\{L\}$ and $\{MD\}$ are expected causes when both occur. In the conjunctive case, $FF := L \land MD$, the only expected cause is $\{L, MD\}$. \cite{halpernModificationHalpernPearlDefinition2015} proposes an extended version introducing intermediary variables $A := L \land \neg MD$, $B := \neg L \land MD$, and $C := L \land MD$, with $FF := A \lor B \lor C$. In this model, $L$, $MD$, and $C$ are expected causes, each under
appropriate witnesses.

\begin{table}[ht!]
    \centering
    \input{tables/sanity}
    \caption{Sanity check of MBS and ISI with an unlimited beam size against some classic small SCMs from the literature.}
    \label{tab:sanity}
\end{table}

\paragraph{Rock Throwing \citep{paulCausationUsersGuide2013}.} 
This is Example~\ref{example:Suzy} of the main paper. We recall only that the expected causes are Suzy throwing the rock $\{ST\}$ and Suzy hitting the bottle $\{SH\}$.

\paragraph{Prisoners \citep{hopkinsClarifyingUsageStructural}.} 
This scenario illustrates how the HP definition excludes irrelevant variables. A prisoner dies if either $A$ loads $B$'s gun and $B$ shoots, or if $C$ loads and shoots his own gun: $D := (A \land B) \lor C$. In the context $a = c = 1$ and $b = 0$, the only expected cause is $\{C\}$, while $A$ is correctly excluded.

\paragraph{Assassin \citep{hitchcockPreventionPreemptionPrinciple2007}.}
This scenario illustrates a case where the HP-definition conflicts with common intuitions and motivates the introduction of normality constraints. An assassin may poison a victim’s drink ($AP$), while a bodyguard may add an antidote ($BA$), which neutralizes the poison and ensures the victim survives ($VS$). The model is given by $VS := \neg AP \lor BA$. We consider the context $ap = 0$ and $ba = 1$, where the assassin does nothing and the bodyguard adds the antidote. Intuitively, the victim’s survival is expected and does not seem to have a meaningful cause. However, under the HP-definition, the set $\{AP, BA\}$ is identified as an actual cause of $VS$, highlighting the need for additional criteria such as normality to filter out such unintuitive causes.

\paragraph{Lamp \citep{weslakeAPartialTheoryofActualCausation2015}.} 
This example is another illustration of how the HP definition excludes irrelevant variables. In this scenario, a lamp $L$ is controlled by three switches $A$, $B$, and $C$ that each have position $\{-1,0,1\}$. The lamp is on only if two switches share the same position. We consider the case where $A$ has position $1$, while $B$ and $C$ have position $-1$. Here, $\vars=\{A,B,C,L\}$, $\exo=\{a,b,c\}$, $A:=a$, $B:=b$, $C:=c$, $L:=(A=B)\lor(A=C)\lor(B=C)$. The expected causes are $\{B\}$ and $\{C\}$.

\paragraph{Ranch \citep{glymourActualCausationStone2010}.} 
This example models a group decision at a ranch, where five agents $A_1,\ldots,A_5$ vote on whether to leave a campfire for a roundup. The outcome $O$ follows a conditional protocol: if $A_1=A_2$, their vote determines the outcome; if $A_2,\ldots,A_5$ agree while $A_1$ disagrees, $A_1$ decides; otherwise, the majority vote applies. In the context $a_1=a_2=1$ and $a_3=a_4=a_5=0$, the first rule applies. The HP-causes are $\{A_1\}$ and all combinations involving $A_2$ and $A_3,\dots,A_5$. Indeed, canceling $A_2$ alone would trigger the second rule, while canceling $A_2$ and another (besides $A_1$) would trigger rule 3 and cancel the consequence. \cite{halpernModificationHalpernPearlDefinition2015} introduces intermediary variables $M_1$, $M_2$, and $M_3$ to represent which decision rule is active, yielding the more intuitive causes $A_1$, $A_2$, and $M_1$.

\paragraph{Vote \citep{livengoodActualCausationSimple2013}.} 
In a majority vote between two candidates with a $4$-$2$ outcome, any pair of voters who voted for the winner constitutes a cause. Extending the model to three candidates with a $4$-$2$-$0$ outcome preserves the same set of expected causes. We implement both cases with contexts $a_1=a_2=a_3=a_4=1$ and $a_5=a_6=0$.

%% file: tables/sanity.tex
\begin{tabular}{C{4cm}|cc|cc|cc}
\toprule
 & \multicolumn{2}{c}{n\_calls} & \multicolumn{2}{c}{t (ms)} & \multicolumn{2}{c}{correct?} \\
 & ISI & MBS & ISI & MBS & ISI & MBS \\
Model &   &  &  &  &  &  \\
\midrule
Forest Fire Disjunctive & 5 & 5 & 9.31 & 0.12 & {\color{green}\checkmark} & {\color{green}\checkmark} \\
\cline{1-7}
Forest Fire Conjunctive & 5 & 2 & 0.13 & 0.04 & {\color{green}\checkmark} & {\color{green}\checkmark} \\
\cline{1-7}
Rock Throwing &  4 & 33 & 0.11 & 0.21 & {\color{green}\checkmark} & {\color{green}\checkmark} \\
\cline{1-7}
Prisoners  & 12 & 11 & 0.18 & 0.10 & {\color{green}\checkmark} & {\color{green}\checkmark} \\
\cline{1-7}
Assassin Variant 1  & 5 & 5 & 0.09 & 0.06 & {\color{green}\checkmark} & {\color{green}\checkmark} \\
\cline{1-7}
Assassin Variant 2  & 3 & 5 & 0.03 & 0.06 & {\color{green}\checkmark} & {\color{green}\checkmark} \\
\cline{1-7}
Lamp  & 36 & 12 & 0.52 & 0.39 & {\color{green}\checkmark} & {\color{green}\checkmark} \\
\cline{1-7}
Forest Fire Extended & 12 & 56 & 0.23 & 0.34 & {\color{green}\checkmark} & {\color{green}\checkmark} \\
\cline{1-7}
Ranch & 175 & 96 & 8.81 & 0.72 & {\color{green}\checkmark} & {\color{green}\checkmark} \\
\cline{1-7}
Ranch Extended & 3184 & 3218 & 189.09 & 19.74 & {\color{green}\checkmark} & {\color{green}\checkmark} \\
\cline{1-7}
Vote &  127 & 118 & 2.32 & 0.59 & {\color{green}\checkmark} & {\color{green}\checkmark} \\
\cline{1-7}
Vote 3 Ways & 502 & 312 & 6.68 & 1.33 & {\color{green}\checkmark} & {\color{green}\checkmark} \\
\cline{1-7}
Rock Throwing Extended &  8 & 56 & 0.16 & 0.30 & {\color{green}\checkmark} & {\color{green}\checkmark} \\
\cline{1-7}
\bottomrule
\end{tabular}

%% file: annexes/impact.tex
\section{Measure of the Impact of our Algorithms}\label{an:algo-gain}

\begin{figure}
    \centering
    \begin{subfigure}{.49\linewidth}
        \input{tables/call_gain_ISI}
        \caption{\% of calls gained using ISI}
    \end{subfigure}
    \hfill
    \begin{subfigure}{.49\linewidth}
        \input{tables/dice_gain_ISI}
        \caption{DICE points gained using ISI}
    \end{subfigure}
    \caption{Performance improvement due to ISI}
    \label{fig:isi-perf-gain}
\end{figure}

We report the decrease in the number of model calls (in \%) and the increase in the quality of the identified cause set (in DICE points) when using the improved algorithms over their naive counterparts, i.e., ISI over Minimal Beam Search in Figure \ref{fig:isi-perf-gain} and LUCB over Naive Average in Figure \ref{fig:lucb-perf-gain}.

ISI always significantly reduces the number of model calls. The gain in DICE points with ISI is also systematic. For $n = 5$, ISI always increases the DICE score by more than 80 points. For $n = 10$, the improvement is nuanced for small beam sizes ($b \leq 16$), but represents more than 60 DICE points otherwise. For $n = 2$, the gains are less pronounced only because MBS performs well and ISI saturates at 100\% DICE.

Similarly, LUCB reduces the number of model calls by 11\% to 37\% in nearly all settings. The only exception occurs for $n = 2$ when using MBS. The algorithm’s execution explains this behavior: naive average estimates are sufficiently noisy that the algorithm sometimes adds incorrect causes and terminates one step too early, which reduces the total number of model calls, even though LUCB reduces the number of calls at each step.

\begin{figure}
    \centering
    \begin{subfigure}{.49\linewidth}
        \input{tables/call_gain_LUCB}
        \caption{\% of calls gained using LUCB}
    \end{subfigure}
    \hfill
    \begin{subfigure}{.49\linewidth}
        \input{tables/dice_gain_LUCB}
        \caption{DICE points gained using LUCB}
    \end{subfigure}
    \caption{Performance improvement due to LUCB}
    \label{fig:lucb-perf-gain}
\end{figure}

In most settings, LUCB either improves the DICE score or limits its degradation to less than three points. This loss notably occurs for $n = 10$ using ISI for lower beam size ($b\leq 32$). However, such low beam sizes are not recommended in practice, and we consider these cases to be non-representative edge cases rather than indicative of the algorithm’s typical behavior.

%% file: tables/call_gain_ISI.tex
\centering
\begin{tabular}{cc|cc}
\toprule
 &  & Boolean & Non-Boolean \\
n & bs &  &  \\
\midrule
\multirow{8}{*}{2} & 2 & 57 \textcolor{gray}{\scriptsize$\pm$ 45} & 62 \textcolor{gray}{\scriptsize$\pm$ 42} \\
 & 4 & 80 \textcolor{gray}{\scriptsize$\pm$ 23} & 79 \textcolor{gray}{\scriptsize$\pm$ 22} \\
 & 8 & 90 \textcolor{gray}{\scriptsize$\pm$ 12} & 89 \textcolor{gray}{\scriptsize$\pm$ 10} \\
 & 16 & 94 \textcolor{gray}{\scriptsize$\pm$ 7} & 94 \textcolor{gray}{\scriptsize$\pm$ 5} \\
 & 32 & 97 \textcolor{gray}{\scriptsize$\pm$ 4} & 97 \textcolor{gray}{\scriptsize$\pm$ 3} \\
 & 64 & 98 \textcolor{gray}{\scriptsize$\pm$ 2} & 98 \textcolor{gray}{\scriptsize$\pm$ 2} \\
 & 128 & 99 \textcolor{gray}{\scriptsize$\pm$ 1} & 99 \textcolor{gray}{\scriptsize$\pm$ 1} \\
 & 256 & 99 \textcolor{gray}{\scriptsize$\pm$ 1} & 99 \textcolor{gray}{\scriptsize$\pm$ 1} \\
\cline{1-4}
\multirow{8}{*}{5} & 2 & 56 \textcolor{gray}{\scriptsize$\pm$ 27} & 63 \textcolor{gray}{\scriptsize$\pm$ 19} \\
 & 4 & 77 \textcolor{gray}{\scriptsize$\pm$ 11} & 76 \textcolor{gray}{\scriptsize$\pm$ 8} \\
 & 8 & 87 \textcolor{gray}{\scriptsize$\pm$ 6} & 85 \textcolor{gray}{\scriptsize$\pm$ 5} \\
 & 16 & 93 \textcolor{gray}{\scriptsize$\pm$ 3} & 90 \textcolor{gray}{\scriptsize$\pm$ 3} \\
 & 32 & 96 \textcolor{gray}{\scriptsize$\pm$ 2} & 93 \textcolor{gray}{\scriptsize$\pm$ 1} \\
 & 64 & 98 \textcolor{gray}{\scriptsize$\pm$ 1} & 95 \textcolor{gray}{\scriptsize$\pm$ 1} \\
 & 128 & 99 \textcolor{gray}{\scriptsize$\pm$ 0} & 96 \textcolor{gray}{\scriptsize$\pm$ 0} \\
 & 256 & 99 \textcolor{gray}{\scriptsize$\pm$ 0} & 97 \textcolor{gray}{\scriptsize$\pm$ 0} \\
\cline{1-4}
\multirow{8}{*}{10} & 2 & 83 \textcolor{gray}{\scriptsize$\pm$ 7} & 81 \textcolor{gray}{\scriptsize$\pm$ 4} \\
 & 4 & 84 \textcolor{gray}{\scriptsize$\pm$ 5} & 82 \textcolor{gray}{\scriptsize$\pm$ 3} \\
 & 8 & 86 \textcolor{gray}{\scriptsize$\pm$ 4} & 83 \textcolor{gray}{\scriptsize$\pm$ 2} \\
 & 16 & 88 \textcolor{gray}{\scriptsize$\pm$ 3} & 84 \textcolor{gray}{\scriptsize$\pm$ 2} \\
 & 32 & 90 \textcolor{gray}{\scriptsize$\pm$ 2} & 86 \textcolor{gray}{\scriptsize$\pm$ 1} \\
 & 64 & 92 \textcolor{gray}{\scriptsize$\pm$ 1} & 87 \textcolor{gray}{\scriptsize$\pm$ 1} \\
 & 128 & 93 \textcolor{gray}{\scriptsize$\pm$ 1} & 89 \textcolor{gray}{\scriptsize$\pm$ 1} \\
 & 256 & 95 \textcolor{gray}{\scriptsize$\pm$ 1} & 89 \textcolor{gray}{\scriptsize$\pm$ 1} \\
\cline{1-4}
\bottomrule
\end{tabular}

%% file: tables/dice_gain_ISI.tex
\centering
\begin{tabular}{cc|cc}
\toprule
 &  & Boolean & Non-Boolean \\
n & bs &  &  \\
\midrule
\multirow{8}{*}{2} & 2 & 37 \textcolor{gray}{\scriptsize$\pm$ 21} & 35 \textcolor{gray}{\scriptsize$\pm$ 27} \\
 & 4 & 26 \textcolor{gray}{\scriptsize$\pm$ 25} & 26 \textcolor{gray}{\scriptsize$\pm$ 27} \\
 & 8 & 19 \textcolor{gray}{\scriptsize$\pm$ 24} & 19 \textcolor{gray}{\scriptsize$\pm$ 25} \\
 & 16 & 0 \textcolor{gray}{\scriptsize$\pm$ 0} & 0 \textcolor{gray}{\scriptsize$\pm$ 0} \\
 & 32 & 0 \textcolor{gray}{\scriptsize$\pm$ 0} & 0 \textcolor{gray}{\scriptsize$\pm$ 0} \\
 & 64 & 0 \textcolor{gray}{\scriptsize$\pm$ 0} & 0 \textcolor{gray}{\scriptsize$\pm$ 0} \\
 & 128 & 0 \textcolor{gray}{\scriptsize$\pm$ 0} & 0 \textcolor{gray}{\scriptsize$\pm$ 0} \\
 & 256 & 0 \textcolor{gray}{\scriptsize$\pm$ 0} & 0 \textcolor{gray}{\scriptsize$\pm$ 0} \\
\cline{1-4}
\multirow{8}{*}{5} & 2 & 83 \textcolor{gray}{\scriptsize$\pm$ 24} & 84 \textcolor{gray}{\scriptsize$\pm$ 24} \\
 & 4 & 89 \textcolor{gray}{\scriptsize$\pm$ 9} & 90 \textcolor{gray}{\scriptsize$\pm$ 9} \\
 & 8 & 90 \textcolor{gray}{\scriptsize$\pm$ 9} & 90 \textcolor{gray}{\scriptsize$\pm$ 8} \\
 & 16 & 86 \textcolor{gray}{\scriptsize$\pm$ 8} & 89 \textcolor{gray}{\scriptsize$\pm$ 7} \\
 & 32 & 87 \textcolor{gray}{\scriptsize$\pm$ 9} & 85 \textcolor{gray}{\scriptsize$\pm$ 8} \\
 & 64 & 87 \textcolor{gray}{\scriptsize$\pm$ 10} & 87 \textcolor{gray}{\scriptsize$\pm$ 10} \\
 & 128 & 87 \textcolor{gray}{\scriptsize$\pm$ 14} & 88 \textcolor{gray}{\scriptsize$\pm$ 12} \\
 & 256 & 84 \textcolor{gray}{\scriptsize$\pm$ 25} & 88 \textcolor{gray}{\scriptsize$\pm$ 17} \\
\cline{1-4}
\multirow{8}{*}{10} & 2 & 22 \textcolor{gray}{\scriptsize$\pm$ 22} & 22 \textcolor{gray}{\scriptsize$\pm$ 22} \\
 & 4 & 32 \textcolor{gray}{\scriptsize$\pm$ 27} & 32 \textcolor{gray}{\scriptsize$\pm$ 27} \\
 & 8 & 46 \textcolor{gray}{\scriptsize$\pm$ 34} & 46 \textcolor{gray}{\scriptsize$\pm$ 34} \\
 & 16 & 56 \textcolor{gray}{\scriptsize$\pm$ 39} & 57 \textcolor{gray}{\scriptsize$\pm$ 38} \\
 & 32 & 63 \textcolor{gray}{\scriptsize$\pm$ 38} & 67 \textcolor{gray}{\scriptsize$\pm$ 38} \\
 & 64 & 74 \textcolor{gray}{\scriptsize$\pm$ 32} & 72 \textcolor{gray}{\scriptsize$\pm$ 33} \\
 & 128 & 80 \textcolor{gray}{\scriptsize$\pm$ 25} & 80 \textcolor{gray}{\scriptsize$\pm$ 25} \\
 & 256 & 88 \textcolor{gray}{\scriptsize$\pm$ 9} & 87 \textcolor{gray}{\scriptsize$\pm$ 10} \\
\cline{1-4}
\bottomrule
\end{tabular}

%% file: tables/call_gain_LUCB.tex
\centering
\begin{tabular}{cc|cc}
\toprule
 &  & MBS & ISI \\
n & bs &  &  \\
\midrule
\multirow{8}{*}{2} & 2 & -20 \textcolor{gray}{\scriptsize$\pm$ 19} & 30 \textcolor{gray}{\scriptsize$\pm$ 2} \\
 & 4 & -20 \textcolor{gray}{\scriptsize$\pm$ 15} & 30 \textcolor{gray}{\scriptsize$\pm$ 2} \\
 & 8 & -19 \textcolor{gray}{\scriptsize$\pm$ 14} & 30 \textcolor{gray}{\scriptsize$\pm$ 2} \\
 & 16 & -26 \textcolor{gray}{\scriptsize$\pm$ 14} & 30 \textcolor{gray}{\scriptsize$\pm$ 2} \\
 & 32 & -38 \textcolor{gray}{\scriptsize$\pm$ 22} & 30 \textcolor{gray}{\scriptsize$\pm$ 2} \\
 & 64 & -51 \textcolor{gray}{\scriptsize$\pm$ 29} & 30 \textcolor{gray}{\scriptsize$\pm$ 2} \\
 & 128 & -30 \textcolor{gray}{\scriptsize$\pm$ 37} & 30 \textcolor{gray}{\scriptsize$\pm$ 2} \\
 & 256 & 11 \textcolor{gray}{\scriptsize$\pm$ 27} & 30 \textcolor{gray}{\scriptsize$\pm$ 2} \\
\cline{1-4}
\multirow{8}{*}{5} & 2 & 19 \textcolor{gray}{\scriptsize$\pm$ 5} & 37 \textcolor{gray}{\scriptsize$\pm$ 16} \\
 & 4 & 19 \textcolor{gray}{\scriptsize$\pm$ 4} & 31 \textcolor{gray}{\scriptsize$\pm$ 2} \\
 & 8 & 20 \textcolor{gray}{\scriptsize$\pm$ 0} & 29 \textcolor{gray}{\scriptsize$\pm$ 2} \\
 & 16 & 20 \textcolor{gray}{\scriptsize$\pm$ 0} & 28 \textcolor{gray}{\scriptsize$\pm$ 2} \\
 & 32 & 20 \textcolor{gray}{\scriptsize$\pm$ 0} & 27 \textcolor{gray}{\scriptsize$\pm$ 2} \\
 & 64 & 20 \textcolor{gray}{\scriptsize$\pm$ 0} & 27 \textcolor{gray}{\scriptsize$\pm$ 2} \\
 & 128 & 20 \textcolor{gray}{\scriptsize$\pm$ 0} & 27 \textcolor{gray}{\scriptsize$\pm$ 2} \\
 & 256 & 20 \textcolor{gray}{\scriptsize$\pm$ 0} & 27 \textcolor{gray}{\scriptsize$\pm$ 2} \\
\cline{1-4}
\multirow{8}{*}{10} & 2 & 20 \textcolor{gray}{\scriptsize$\pm$ 1} & 27 \textcolor{gray}{\scriptsize$\pm$ 16} \\
 & 4 & 20 \textcolor{gray}{\scriptsize$\pm$ 0} & 30 \textcolor{gray}{\scriptsize$\pm$ 7} \\
 & 8 & 20 \textcolor{gray}{\scriptsize$\pm$ 0} & 28 \textcolor{gray}{\scriptsize$\pm$ 5} \\
 & 16 & 20 \textcolor{gray}{\scriptsize$\pm$ 0} & 27 \textcolor{gray}{\scriptsize$\pm$ 3} \\
 & 32 & 20 \textcolor{gray}{\scriptsize$\pm$ 0} & 25 \textcolor{gray}{\scriptsize$\pm$ 3} \\
 & 64 & 20 \textcolor{gray}{\scriptsize$\pm$ 0} & 23 \textcolor{gray}{\scriptsize$\pm$ 2} \\
 & 128 & 20 \textcolor{gray}{\scriptsize$\pm$ 0} & 22 \textcolor{gray}{\scriptsize$\pm$ 1} \\
 & 256 & 20 \textcolor{gray}{\scriptsize$\pm$ 0} & 21 \textcolor{gray}{\scriptsize$\pm$ 1} \\
\cline{1-4}
\bottomrule
\end{tabular}

%% file: tables/dice_gain_LUCB.tex
\centering
\begin{tabular}{cc|cc}
\toprule
 &  & MBS & ISI \\
n & bs &  &  \\
\midrule
\multirow{8}{*}{2} & 2 & 4 \textcolor{gray}{\scriptsize$\pm$ 6} & 0 \textcolor{gray}{\scriptsize$\pm$ 4} \\
 & 4 & 6 \textcolor{gray}{\scriptsize$\pm$ 5} & 0 \textcolor{gray}{\scriptsize$\pm$ 4} \\
 & 8 & 9 \textcolor{gray}{\scriptsize$\pm$ 9} & 0 \textcolor{gray}{\scriptsize$\pm$ 4} \\
 & 16 & 11 \textcolor{gray}{\scriptsize$\pm$ 12} & 0 \textcolor{gray}{\scriptsize$\pm$ 4} \\
 & 32 & 6 \textcolor{gray}{\scriptsize$\pm$ 12} & 0 \textcolor{gray}{\scriptsize$\pm$ 4} \\
 & 64 & 4 \textcolor{gray}{\scriptsize$\pm$ 12} & 0 \textcolor{gray}{\scriptsize$\pm$ 4} \\
 & 128 & 3 \textcolor{gray}{\scriptsize$\pm$ 9} & 0 \textcolor{gray}{\scriptsize$\pm$ 4} \\
 & 256 & 9 \textcolor{gray}{\scriptsize$\pm$ 7} & 0 \textcolor{gray}{\scriptsize$\pm$ 4} \\
\cline{1-4}
\multirow{8}{*}{5} & 2 & 0 \textcolor{gray}{\scriptsize$\pm$ 1} & -3 \textcolor{gray}{\scriptsize$\pm$ 10} \\
 & 4 & -0 \textcolor{gray}{\scriptsize$\pm$ 1} & -1 \textcolor{gray}{\scriptsize$\pm$ 3} \\
 & 8 & 0 \textcolor{gray}{\scriptsize$\pm$ 1} & 0 \textcolor{gray}{\scriptsize$\pm$ 0} \\
 & 16 & -0 \textcolor{gray}{\scriptsize$\pm$ 1} & 0 \textcolor{gray}{\scriptsize$\pm$ 0} \\
 & 32 & 0 \textcolor{gray}{\scriptsize$\pm$ 1} & 0 \textcolor{gray}{\scriptsize$\pm$ 0} \\
 & 64 & -0 \textcolor{gray}{\scriptsize$\pm$ 1} & 0 \textcolor{gray}{\scriptsize$\pm$ 0} \\
 & 128 & -0 \textcolor{gray}{\scriptsize$\pm$ 1} & 0 \textcolor{gray}{\scriptsize$\pm$ 0} \\
 & 256 & 0 \textcolor{gray}{\scriptsize$\pm$ 2} & 0 \textcolor{gray}{\scriptsize$\pm$ 0} \\
\cline{1-4}
\multirow{8}{*}{10} & 2 & 0 \textcolor{gray}{\scriptsize$\pm$ 0} & -0 \textcolor{gray}{\scriptsize$\pm$ 17} \\
 & 4 & -0 \textcolor{gray}{\scriptsize$\pm$ 0} & -3 \textcolor{gray}{\scriptsize$\pm$ 10} \\
 & 8 & 0 \textcolor{gray}{\scriptsize$\pm$ 0} & 1 \textcolor{gray}{\scriptsize$\pm$ 13} \\
 & 16 & 0 \textcolor{gray}{\scriptsize$\pm$ 0} & -1 \textcolor{gray}{\scriptsize$\pm$ 10} \\
 & 32 & 0 \textcolor{gray}{\scriptsize$\pm$ 1} & -2 \textcolor{gray}{\scriptsize$\pm$ 6} \\
 & 64 & 0 \textcolor{gray}{\scriptsize$\pm$ 1} & -0 \textcolor{gray}{\scriptsize$\pm$ 3} \\
 & 128 & -0 \textcolor{gray}{\scriptsize$\pm$ 1} & 0 \textcolor{gray}{\scriptsize$\pm$ 0} \\
 & 256 & -0 \textcolor{gray}{\scriptsize$\pm$ 1} & 0 \textcolor{gray}{\scriptsize$\pm$ 0} \\
\cline{1-4}
\bottomrule
\end{tabular}

%% file: annexes/structure_sweep.tex
\section{Structure Sweep}\label{an:sweep}

The evaluation in Section~\ref{sec:experiments} centers on a single family of SCMs. To assess how our conclusions generalize to other causal structures, we run our algorithms on synthetic DAGs whose structural properties we control explicitly. All graphs share a single target sink $T$.

\subsection{Experimental Setup}

We use four random-graph generators, each isolating different structural properties. The \emph{layered} generator arranges nodes in successive layers with edges only between consecutive layers. A parameter controls the probability $p$ of each node from a given layer to connect with each node of the next layer. The \emph{tree-with-shortcuts} generator starts from a pure in-tree, and adds a controlled number $r$ of shortcut edges, each introducing one diamond (two independent paths to $T$); $r$ is thus a direct knob on reconvergence. An attachment bias parameter controls how close to $T$ nodes attach, and hence whether the tree is deep or bushy. The \emph{fixed-in-degree} generator gives every non-source node exactly $k$ parents, isolating the effect of a high in-degree. The \emph{Erd\H{o}s--R\'enyi} generator adds forward edges independently at a target average degree. Additionally, we complete the \emph{Erd\H{o}s--R\'enyi} SCMs by giving one random parent to all nodes that do not have one. This process yields heterogeneous in-degrees as a realistic average-case baseline.

We also vary the structural assignment used to obtain SCMs from these DAGs. By default, we consider a threshold function: a variable takes a positive value when more than a fraction of its parents are positive, which is controlled with a threshold parameter $\theta$. We also use negative thresholds, i.e., a variable takes a positive value when less than a fraction of its parents are positive; and parity, i.e., a variable takes a positive value when it has an even number of positive parents.

For every configuration, we compute the exact set of HP-causes with MBS at beam size $b=-1$ and no step limit, and use it as the reference against which the approximate variants are scored. We restrict ourselves to graphs small enough to keep this exact identification tractable.

\subsection{Beam Size Tradeoff}

\input{tables/table1}

Table~\ref{tab:dag_types} evaluates full identification across the four families, each in a base, a denser, and a larger (or longer) configuration, sweeping the beam size over $b\in\{2,\dots,256\}$.
We report the minimum and maximum DICE scores and the number of calls across the tested beam sizes. Additionally, we report the number of beam size increasing steps that lead to an increasing, flat, or decreasing ratio of DICE gain over call gain. This evaluates how often the beam size parameter offers the expected control over the precision-efficiency tradeoff. We report these quantities for 13 graph parameters spanning 4 different graph structures. Reported results are the median values across 20 random contexts and graphs (all generated using the reported parameters).

For the \emph{layered graph}, we use base, long, large, and dense settings, using widths in $(3, 3, 5, 3)$; lengths in $(5, 6, 3, 5)$; and edge probabilities in $(0.5, 0.5, 0.5, 0.7)$. For the \emph{tree-with-shortcuts} graphs, we use base, dense, and large settings, with node counts in $(15, 15, 20)$; and numbers of reconvergences in $(5, 7, 5)$. For the \emph{fixed-in-degree} graphs, we also use base, dense, and large settings, with node counts in $(20, 20, 30)$; and in-degrees in $(3, 5, 4)$. Finally, for the \emph{Erd\H{o}s--R\'enyi} graphs, the base, dense, and large settings have node counts in $(15, 15, 20)$; and average in-degrees in $(2, 3, 2)$.

For the approximate algorithms, we used a maximum number of steps of 10. We used the strongest set of approximations for the ISI algorithm: MBS reports only minimal interventions instead of the full interventions needed by exact ISI; the remaining assignments are done using the naive approach presented in Section~\ref{sec:isi}.

For every graph we test, the tradeoff induced by the beam size is immediately visible. The DICE-per-log-call ratio is never decreasing. The strong approximations used with ISI allow it to keep a very small number of model calls, but sometimes limit its DICE score, notably for the large “fixed in-degree” graph.

\subsection{ISI - MBS Comparison}

\begin{figure}[t]
    \centering
    \includegraphics[width=\linewidth]{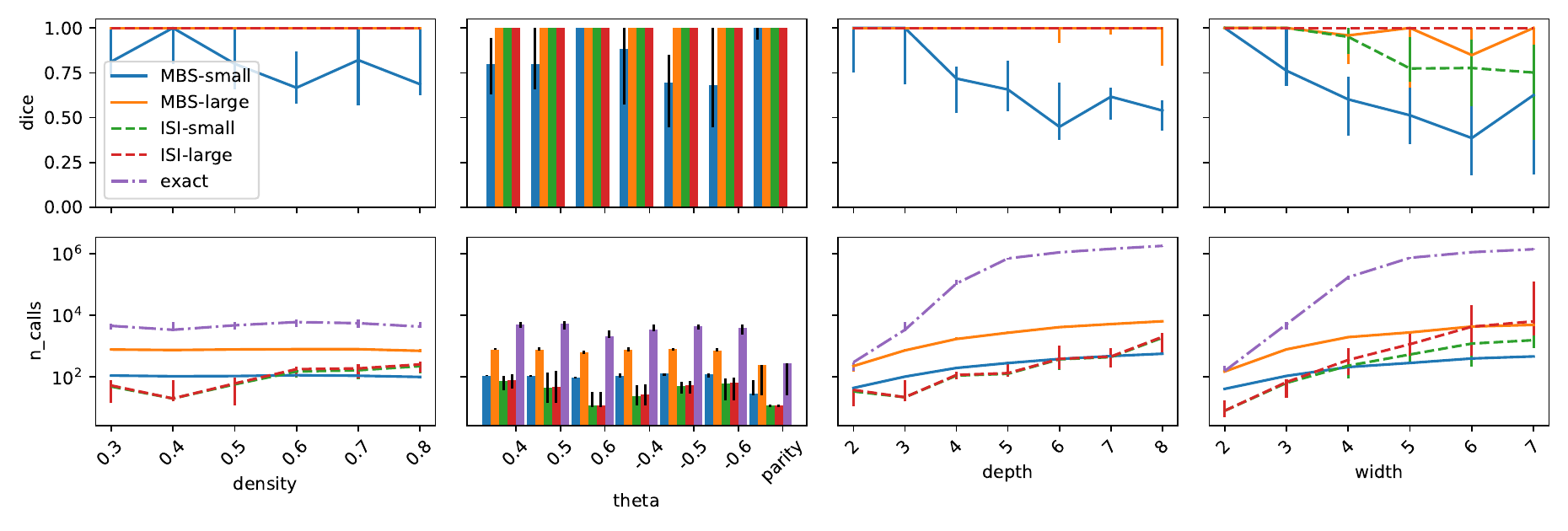}
    \caption{DICE score and number of calls for full cause identification when sweeping through four properties of the “Layers” SCM (column 1: density; column 2: type of structural assignment; column 3: depth; column 4: width). The line color and style correspond to the type of identification. The lines report the median values for 20 contexts, and the error bars report the 25-75 quantiles.}
    \label{fig:layer-sweep}
\end{figure}

We now report how ISI and MBS scale in terms of DICE score and model calls when we sweep through the structural properties of the \emph{layered graph} and the \emph{shortcut tree}. We compare both ISI and MBS with beam sizes $b=2$ and $b=32$. Once again, we use the most approximate version of ISI with minimal interventions returned by the inner MBS and naive complementary assignments.

Figure~\ref{fig:layer-sweep} reports the layered graphs as we sweep four properties one at a time: edge density, the type of structural assignment (threshold sign and parity), depth, and width. The base configuration for parameters that are not swept is: density $0.4$, fractional positive threshold, depth $3$, width $3$. The DICE score (top row) is almost always 1. Only MBS with a small beam size consistently drops in DICE score. ISI with a small beam size also exhibits a drop in DICE score for large widths.

The number of model calls (bottom row) is more sensitive. First, the beam size notably affects MBS but not ISI (except slightly when varying the width). Second, the structural properties have little effect on MBS. Notably, the density has almost no effect at all. Conversely, ISI exhibits notably worse scalability when varying the density, depth, and width of the graph. Finally, contrary to the graph structure, the type of structural assignment has little effect on ISI but affects MBS. The difference is notably important with a high positive threshold and a low negative threshold, which correspond to regimes with more positive values and potentially longer causes.

Overall, these results show that ISI is sensitive to the graph structure whereas MBS is sensitive to the size of the causes and to the beam size. Notably, for dense, wide, or long layered graphs, ISI can end up less performant than MBS.

\begin{figure}[t]
    \centering
    \includegraphics[width=\linewidth]{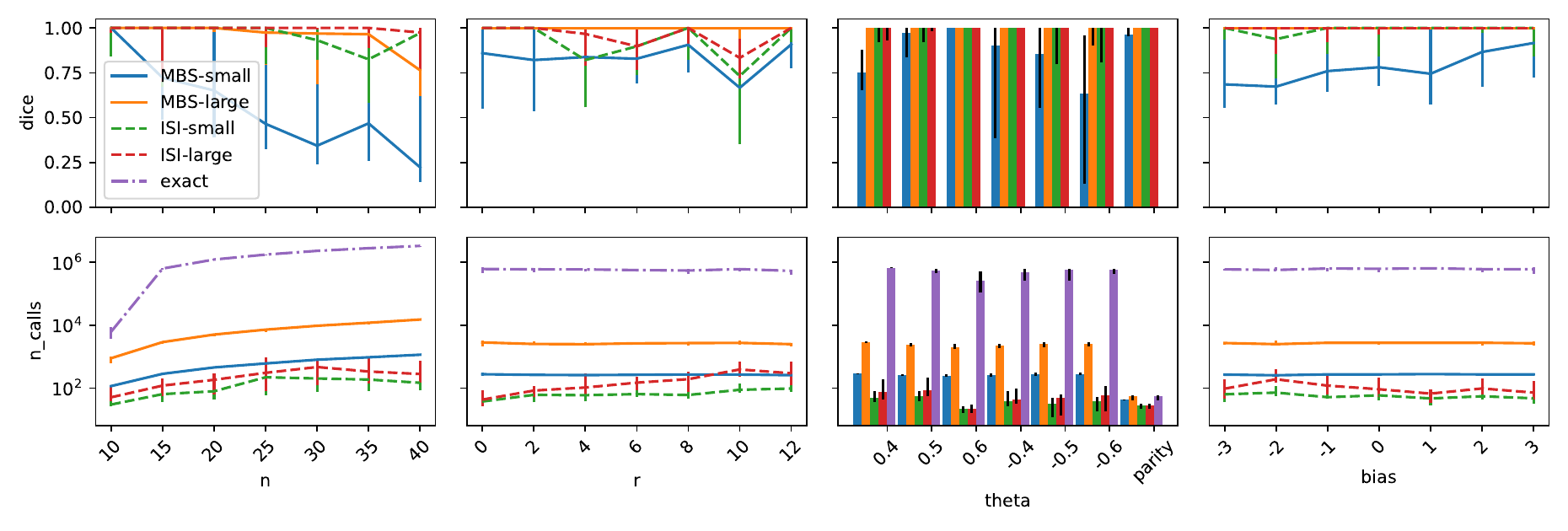}
    \caption{DICE score and number of calls for full cause identification when sweeping through four properties of the Tree SCM (column 1: number of nodes; column 2: number of diamonds; column 3: type of structural assignments; column 4: bias). The line color and style correspond to the type of identification. The lines report the median values for 20 contexts, and the error bars report the 25-75 quantiles.}
    \label{fig:tree-sweep}
\end{figure}

Figure~\ref{fig:tree-sweep} reports the tree-with-shortcuts generator, sweeping the number of nodes $n$, the number of diamonds $r$, the type of structural assignment, and the attachment bias that controls tree depth, around a base configuration ($n=15$, $r=3$, positive thresholds, no bias). As before, DICE remains near-exact for all variants except MBS at $b=2$ (even though ISI's DICE slightly drops for larger values of $r$ and $n$ when the beam size is small).

The number of calls is once again more sensitive. MBS is more sensitive to the number of nodes than ISI. Conversely, MBS is almost not affected at all by the number of reconvergences or by the bias, which both influence ISI. While ISI is less efficient with smaller or larger bias, it remains notably better than MBS with respect to this parameter modification. Conversely, the number of reconvergences is a new example of a property that can make MBS more efficient than ISI. The impact of the structural assignment is similar to what it was in the “layered graph”.

Overall, ISI is usually better than MBS when varying graph structural properties, but larger, longer, or denser “layered graphs” or “shortcut trees” with many reconvergences can produce opposite results.

%% file: tables/table1.tex
\begin{table}[t]
\centering
\begin{tabular}{llrrrrr}
\toprule
 &  & Min DICE & Max DICE & Min calls & Max calls & G / F / D \\
Graph & Algo &  &  &  &  &  \\
\midrule
\multirow[t]{2}{*}{layer-base} & MBS & 0.33 & 1.00 & 154 & 16108 & 4 / 4 / 0 \\
 & ISI & 0.80 & 1.00 & 226 & 488 & 1 / 7 / 0 \\
\cline{1-7}
\multirow[t]{2}{*}{layer-long} & MBS & 0.37 & 1.00 & 202 & 24980 & 5 / 3 / 0 \\
 & ISI & 0.83 & 1.00 & 657 & 981 & 1 / 7 / 0 \\
\cline{1-7}
\multirow[t]{2}{*}{layer-large} & MBS & 0.36 & 1.00 & 154 & 17535 & 6 / 2 / 0 \\
 & ISI & 0.43 & 1.00 & 785 & 6108 & 3 / 5 / 0 \\
\cline{1-7}
\multirow[t]{2}{*}{layer-dense} & MBS & 0.39 & 1.00 & 151 & 16006 & 4 / 4 / 0 \\
 & ISI & 0.74 & 1.00 & 740 & 1594 & 1 / 7 / 0 \\
\cline{1-7}
\multirow[t]{2}{*}{shortcut-base} & MBS & 0.73 & 1.00 & 144 & 15636 & 2 / 6 / 0 \\
 & ISI & 0.72 & 1.00 & 40 & 143 & 1 / 7 / 0 \\
\cline{1-7}
\multirow[t]{2}{*}{shortcut-dense} & MBS & 0.50 & 1.00 & 147 & 15578 & 3 / 5 / 0 \\
 & ISI & 0.79 & 1.00 & 70 & 502 & 1 / 7 / 0 \\
\cline{1-7}
\multirow[t]{2}{*}{shortcut-large} & MBS & 0.21 & 1.00 & 251 & 34035 & 6 / 2 / 0 \\
 & ISI & 0.66 & 0.97 & 83 & 349 & 2 / 6 / 0 \\
\cline{1-7}
\multirow[t]{2}{*}{in-degree-base} & MBS & 1.00 & 1.00 & 224 & 26930 & 0 / 8 / 0 \\
 & ISI & 1.00 & 1.00 & 42 & 49 & 0 / 8 / 0 \\
\cline{1-7}
\multirow[t]{2}{*}{in-degree-dense} & MBS & 0.65 & 1.00 & 212 & 26700 & 5 / 3 / 0 \\
 & ISI & 0.57 & 0.91 & 186 & 4383 & 4 / 4 / 0 \\
\cline{1-7}
\multirow[t]{2}{*}{in-degree-large} & MBS & 0.35 & 0.96 & 424 & 64080 & 8 / 0 / 0 \\
 & ISI & 0.32 & 0.55 & 526 & 22004 & 3 / 5 / 0 \\
\cline{1-7}
\multirow[t]{2}{*}{er-base} & MBS & 1.00 & 1.00 & 154 & 15985 & 0 / 8 / 0 \\
 & ISI & 1.00 & 1.00 & 13 & 23 & 0 / 8 / 0 \\
\cline{1-7}
\multirow[t]{2}{*}{er-dense} & MBS & 0.59 & 1.00 & 150 & 15538 & 4 / 4 / 0 \\
 & ISI & 0.59 & 0.94 & 78 & 1885 & 4 / 4 / 0 \\
\cline{1-7}
\multirow[t]{2}{*}{er-large} & MBS & 0.72 & 1.00 & 252 & 33120 & 3 / 5 / 0 \\
 & ISI & 0.84 & 1.00 & 30 & 102 & 2 / 6 / 0 \\
\cline{1-7}
\bottomrule
\end{tabular}
\caption{DICE score and number of oracle calls across graph types and algorithms, for beam sizes $b \in \{2, 4, 8, 16, 32, 64, 128, 256\}$. G\,/\,F\,/\,D counts the number of beam-size steps (out of 8) where the DICE-per-log-call ratio is growing, flat, or decreasing (threshold $= 0.01$).}
\label{tab:dag_types}
\end{table}

%% file: annexes/complexity.tex
\section{Complexity Analysis} \label{an:complexity-analysis}

This appendix provides a preliminary analysis of the complexity of our algorithms. We conduct a theoretical analysis followed by an empirical one. They both rely on assumptions based on the model used. These results are illustrative and provide early intuition into the complexity of our algorithms; they are not technical contributions.

\subsection{Theoretical Analysis}

The MBS algorithm performs $K\le|\vars|$ iteration steps. At each step $k$, it evaluates the current intervention set $E_k$ whose size is bounded by the beam size $b$ times the number of size-one interventions $|\vars|\times|\dom_m|$, where $\dom_m$ is the largest domain. Then, MBS performs a minimality merge with the already identified causes $\mathbf{C}$, which requires iterations bounded by the size of the largest cause $C_m$.
The number of operations is bounded by:

\[O(K\times b\times|\vars|\times|\dom_m|\times|\mathbf{C}|\times|C_m|)\]

We can further simplify this expression depending on the system or on the use of MBS. Notably, $K$ is either bounded by $|\vars|$ when no maximum number of steps is given to the algorithm or bounded by the provided parameter. $|C_m|$ highly depends on the system nature, but is notably bounded either by $|\vars|$ or by the maximum number of steps, when provided. Concerning $|\mathbf{C}|$, the only formal bound is $2^{|\vars|}$. However, in practice, it can be bounded by the system nature or by the MBS approximations. Notably, in the SMK scenario, both $|\mathbf{C}|$ and $|C_m|$ are bounded regardless of the system size (as only one attacker can cause $DK$ or $SD$). Hence, if we suppose $|\mathbf{C}|$ is bounded, we can expect a complexity linear in $|\dom_m|$, linear in $b$, and either  linear, quadratic, or cubic in $|\vars|$. We can expect $O(|\vars|)$ if we impose a maximum number of steps, and otherwise $O(|\vars|^2)$ if the size of the causes is naturally bounded by the system's nature or $O(|\vars|^3)$ if not.

For the smallest cause identification, we stop when we find a cause, so we never iterate over causes or inside causes. Hence, we can expect linearity with respect to $b$ and $|\dom_m|$, and either linear or quadratic dependence in $|\vars|$ depending on whether we use \texttt{max\_steps}. 

Approximate ISI enqueues all subsets of all “canceling” interventions and runs the MBS algorithm on instances bounded by the largest number of parents possible for a “canceling” intervention $\Pa_m$. We must combine the complexity of ISI (bounded by $2^{|C_m|}\times|\mathbf{C}|$) and the complexity of MBS called on a subinstance. The number of operations is therefore bounded by:

\[2^{|C_m|}\times|\mathbf{C}|^2\times b\times|\Pa_m|\times|\dom_m|\]

The exponential complexity is concerning, but disappears when the system's nature bounds the size of the largest cause. 

Unfortunately, ISI does not offer a way to do an early stop. The complexity of the algorithm is then the same whether we search for the smallest cause or for the full set of causes. We still report the experimentally observed complexity of the highly approximated version used for smallest cause identification.

\begin{figure}
    \centering
    \includegraphics[width=.7\linewidth]{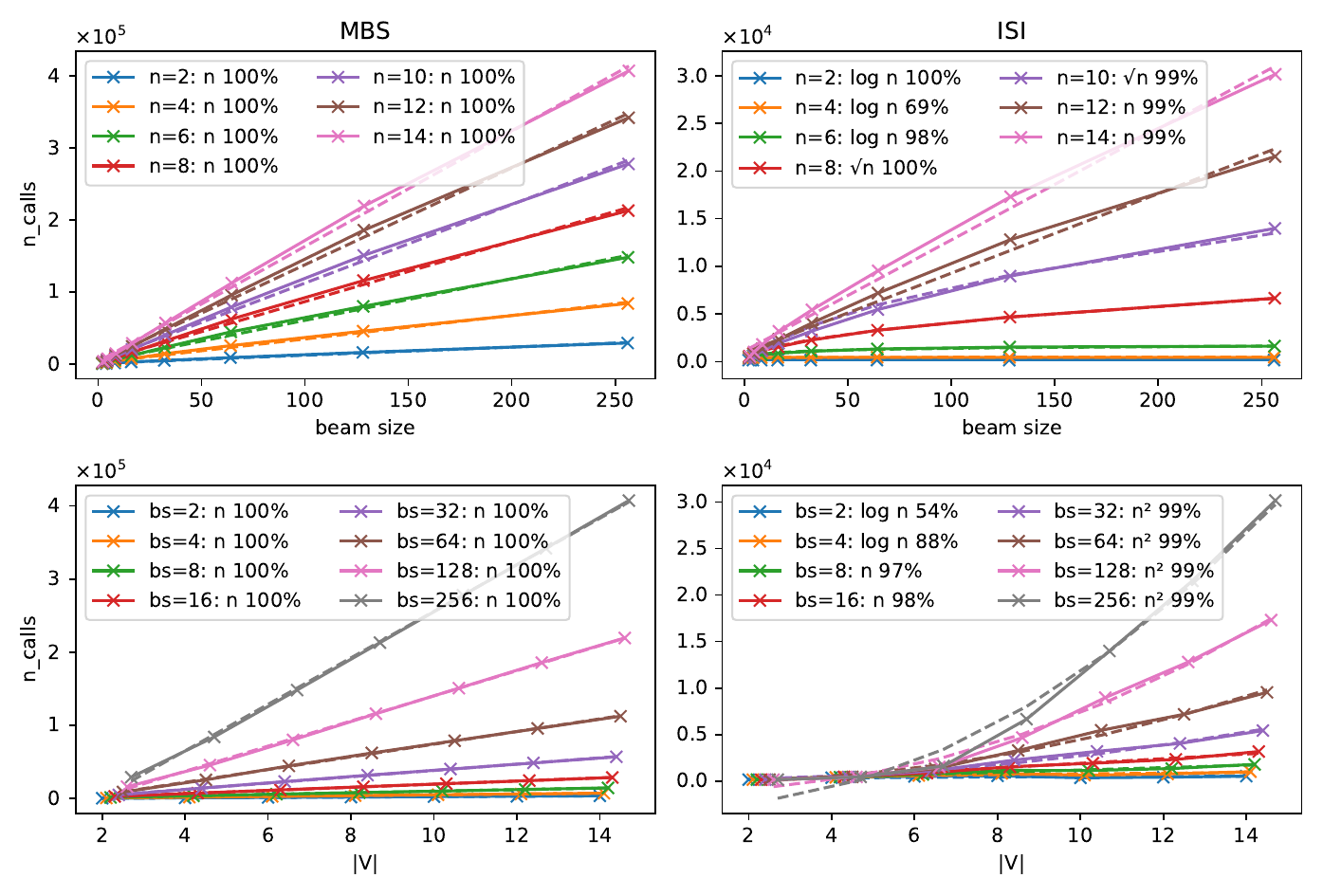}
    \caption{Time complexity against the number of attackers and beam size for the full identification.}
    \label{fig:reg-full}
\end{figure}

\subsection{Empirical Analysis}

We approximate the empirical complexity of our algorithm in terms of the number of variables and beam size. We run regression for several complexities, i.e., we perform a linear regression between a function of the evaluated parameter ($b$ or $|\vars|$) and the number of model calls. We test $O(\text{log}(\cdot))$, $O(\sqrt{\cdot})$, $O(\cdot)$, $O(\cdot^2)$, and $O(\cdot^3)$. We then measure the $R^2$ coefficient for each linear regression and report the best one in \% rounded to the nearest integer.

Figure \ref{fig:reg-full} shows the regressions for the full identification task. MBS scales linearly with both the beam size and the model size. This confirms our analysis, as we used a fixed $\texttt{max\_steps}$ parameter. ISI also scales close to linearly with the beam size. However, the beam size has a noticeable impact on the ISI's scaling factor to the model size. For smaller beam sizes, the scaling is closer to logarithmic, being better than MBS; for larger beam sizes, it is closer to quadratic, being worse than MBS. This highlights that ISI's scaling is more complex than MBS'.

\begin{figure}
    \centering
    \includegraphics[width=.7\linewidth]{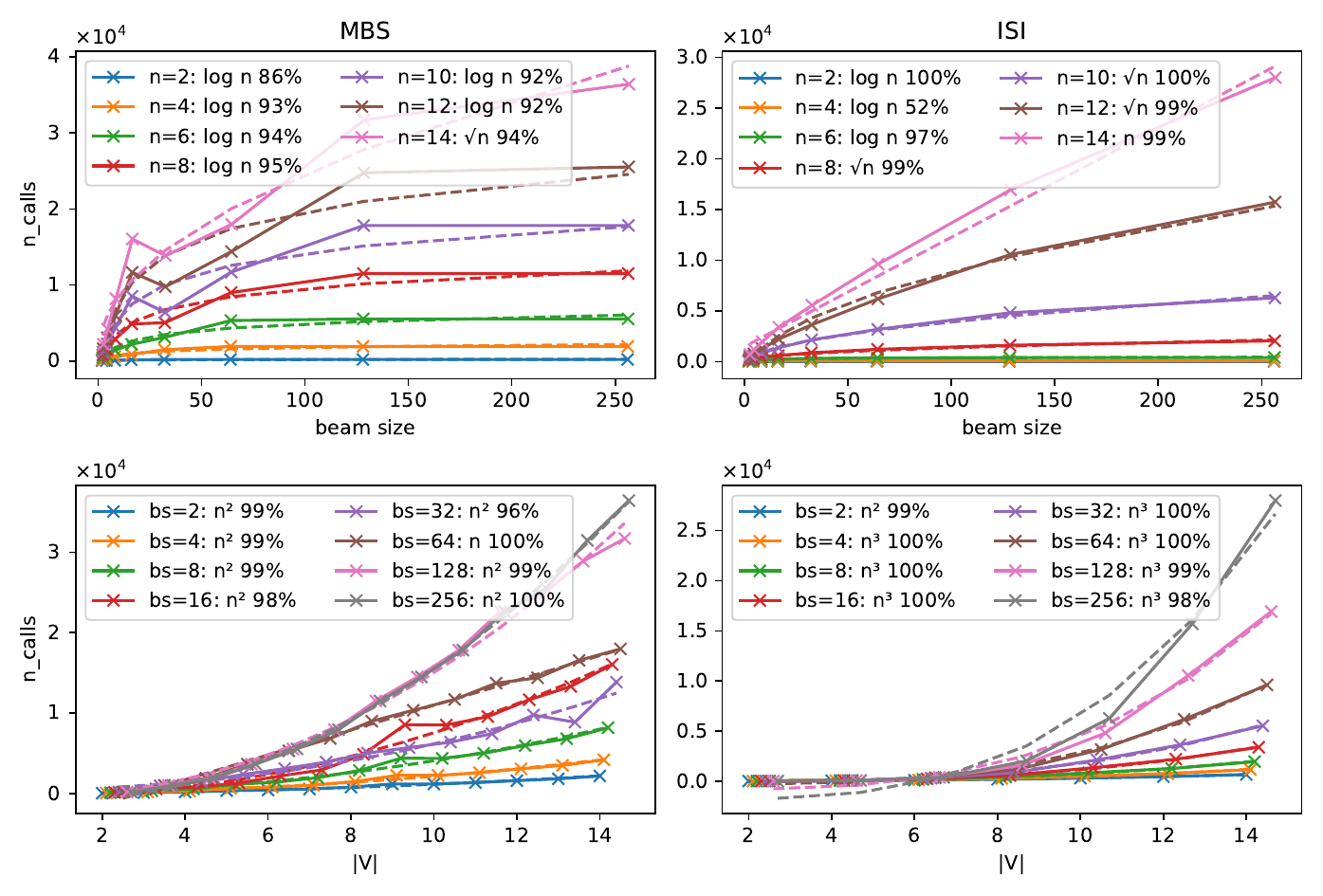}
    \caption{Time complexity against the number of attackers and beam size for the smallest identification.}
    \label{fig:reg-smallest}
\end{figure}

Figure \ref{fig:reg-smallest} shows the regressions for the smallest identification task. MBS scales quadratically with $|V|$ as expected, as no maximum number of steps is imposed. In terms of beam size, the closest complexity is almost always $O(\text{log}(b))$, which is lower than the expected linear complexity. ISI scales mostly sub-linearly (either $O(\text{log}(b))$ or $O(\sqrt{b})$) with the beam size. However, it almost always scales as $O(n^3)$ with the model size. This reveals that using an unlimited beam size worsens the complexity more than the additional approximations.

%% file: annexes/heuristics.tex
\section{Comparison of Heuristics}\label{an:heuristics}

\begin{figure}
    \centering
    \includegraphics[width=.7\linewidth]{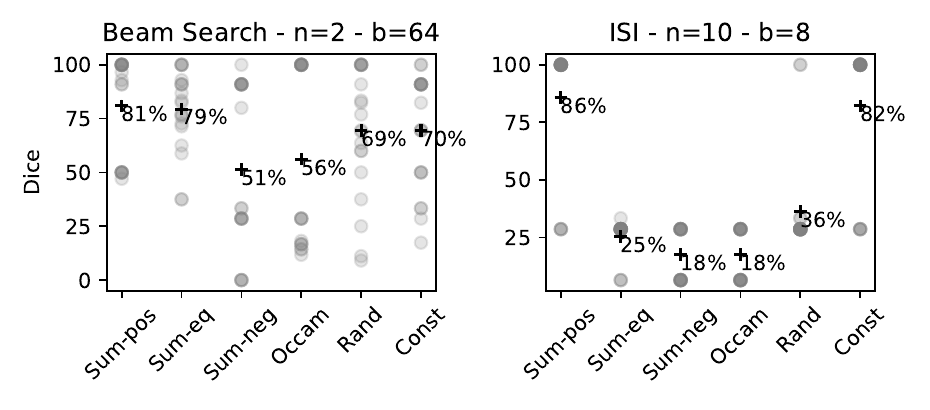}
    \caption{Comparing DICE scores for various heuristics. Dots are the scores for each run, crosses are the averaged values.}
    \label{fig:heuristics}
\end{figure}

This appendix provides illustrative heuristic functions for our SMK model and shows that they influence the algorithm's performance, mostly as expected based on their design. 
Section~\ref{sec:discussion-algo-limitations} discusses the practical challenge of designing heuristics for real-world systems and gives a concrete example of how domain knowledge can guide this choice.

We present six heuristics based on the model counterfactual states that are to be minimized. First, \textit{“sum-pos”}, which we used for the main paper content experiments, computes the sum of positive values. We expect it to perform well as our model uses “and” and “or” based logical formulas with very few “not”. Second, we use \textit{“sum-eq”}, which computes the number of counterfactual values observed in a counterfactual state. This heuristic favors large interventions that apply to variables “far” from $\target$. Then, we use \textit{“sum-neg”}, the complement of “sum-pos”, which counts the number of variables with value 0. We expect this heuristic to yield poor results. Then, we use the complement of “sum-eq”, which counts the number of variables that remain at their actual values. We call this heuristic \textit{“Occam,”} as it can be seen as a measure of how simple it is to describe the counterfactual state compared to the actual one. This measure can be used to assess the quality of causes, as simple ones will provide better explanations \citep{millerExplanationArtificialIntelligence2019,kahnemanNormTheoryComparing1986}. However, we do not expect this metric to yield good results as it favors interventions with little effect on the model, which likely does not imply that we are close to “canceling” $\target$. Finally, we use two control heuristics: the \textit{“random”} heuristic that returns a random integer between 0 and $|\vars|$ and the \textit{“constant”} heuristic that always returns $|\vars|/2$.

Figure \ref{fig:heuristics} shows the DICE scores obtained with our heuristics in the Boolean model, first with MBS, $n=2$ and $b=64$, second with ISI, $n=10$ and $b=8$. These settings correspond to medium DICE values with our original heuristic. In the first case, “sum-pos” and “sum-eq” both perform well and beat the two control heuristics, contrary to “sum-neg” and “Occam”. In the second case, “sum-pos” is still the best heuristic, but the “constant” heuristic almost reaches the same DICE score. As we could expect, the “random” heuristic yields worse results, and “sum-neg” and “Occam” even worse ones. Quite unexpectedly, “sum-eq” also yields worse results than “random”. 

The neutral heuristics (“random” and “constant”) outperforming the counter-aligned ones (“sum-neg” and “Occam”) illustrates that carrying no signal is preferable to actively pointing away from the cancellation goal. This also explains why “Occam”, despite being a natural measure of explanation quality, performs poorly as a search heuristic: minimizing the footprint of an intervention is desirable for explanations but counter-productive for finding interventions that cancel $T$.

%% file: annexes/proofs.tex
\section{Proofs of Theorems}\label{an:proofs}

This appendix specifies the ISI algorithm and proves that it is \emph{sound} (every pair it returns satisfies AC2), \emph{complete} (every HP-cause is among the pairs it returns) and \emph{exact} (the inclusion-minimal pairs it returns are exactly the HP-causes).

The proofs rely on a single lemma about structural equations, the \emph{region lemma} (Lemma~\ref{lem:region}): if one takes any set $L$ of variables and forces its entire parent boundary to the values it takes under some intervention $\xi$, then every variable of $L$ behaves exactly as it does under $\xi$, whatever happens elsewhere in the model.

ISI searches a tree of \emph{subproblems}.  A subproblem records a region $L$ already settled around the target, a set $I$ of free variables lying on the parent boundary of that region, and sets of forced counterfactual interventions (variables $R_C$ forced to values $r_C$) and forced actual interventions (variables $R_W$ forced to their actual values). A solution of a subproblem splits $I$ exhaustively in two: the variables given non-actual values join the cause, the variables held at their actual values join the witness; the split must keep the target flipped. Expanding a solution absorbs a part $S$ of the cause into the region and replaces it by $\Pa(S)$. The completeness proof exhibits, for every HP-cause, one explicit branch of this search that reaches it: the region grows outward from the target and its parent boundary is at every moment split into deviating and non-deviating parents. The region lemma certifies at each stage that the target is still flipped, and the construction stops exactly when the boundary has become the HP-cause itself.

\subsection{Setting and Notation}

We fix a structural causal model $\Mod = (\vars,\exo,\assign,\Dom)$ with finitely many endogenous variables and finite domains, an acyclic causal graph, and a context $u$, together with a target variable $\target \in \vars$.  Without loss of generality $\vars = \An(T) \cup \{T\}$, where $\An(\cdot)$ denotes ancestors (variables outside $\An(T)$ can affect nothing we test). $\Pa(Z)$ is the set of parents of $Z$, extended to sets by $\Pa(S) = \bigcup_{Z \in S}\Pa(Z)$.

We write $v^{*}$ for the assignment of actual values determined by $u$.  For any $X\subseteq\vars,x\in\Dom(X)$ and the associated intervention $\xi = [X \leftarrow x]$ we write $v^{\xi}$ for the resulting assignment. Having fixed a model $\Mod$ and a context $u$ means that we have a fixed set of actual values $v^*\in\Dom(\vars)$. For any sets of variables $X\subseteq\vars$ with values $x\in\Dom(X)$ and $Y\subseteq X$, we denote $\restr{x}{Y}\in\Dom(Y)$ the restriction of values $x$ to variables in $Y$. We generalize this notation to variables (instead of sets of variables). For any set of variables $X\subseteq\vars$, we denote $x^*=\restr{v^*}{X}$ as the set of actual values of $X$.

The main paper used $\target$ as a Boolean predicate. To facilitate graph notations, this appendix considers $T$ as a variable of the graph. Hence, the cancellation condition AC2 becomes $(T\neq t^*)$ instead of $(\lnot T)$. By convention, $T$ never belongs to a cause or to a witness.

\subsection{Actual Causes}

\begin{definition}[Canceling interventions and witnesses]\label{def:aac2}
For $\emptyset \neq C \subseteq \vars \setminus \{T\}$ and $c \in \Dom(C)$ with $c_X \neq x^{*}$
for every $X \in C$, put
\[
   \Wit(C,c) \;=\; \bigl\{\, W \subseteq \vars \setminus (C\cup\{T\}) \;\bigm|\;
   (\Mod,u) \models [C \leftarrow c,\, W \leftarrow w^{*}](T\neq t^*) \,\bigr\}.
\]
$\AACtwo$ is the set of pairs $(C,c)$ as above with $\Wit(C,c) \neq \emptyset$.  (Note that
$W = \emptyset$ is permitted, so $\AACtwo$ contains in particular the pairs that flip $T$
with no witness at all.)
\end{definition}

\begin{definition}[HP-causes]\label{def:hp}
$\mathcal{C}$ is the set of $C$ such that $(C,c) \in \AACtwo$ for some $c$ (AC1--AC2) and no
strict subset $C'' \subsetneq C$ satisfies $(C'',c'') \in \AACtwo$ for any $c''$ (AC3).
\end{definition}

\subsection{Subproblems and the Counterfactual Closure}

\begin{definition}[Subproblem]\label{def:subproblem}
    A \emph{subproblem} is a tuple $\kappa = (L, I, R_C, r_C, R_W)$ where
    \begin{enumerate}
        \item[(i)] $L, I, R_C, R_W$ are pairwise disjoint sets of variables with $T\in L$ (the \emph{settled variables}, the \emph{free instance}, the \emph{counterfactual forced variables}, the \emph{actual forced variables});
        \item[(ii)] $r_C \in \Dom(R_C)$ with $\restr{{r_{C}}}{X} \neq x^{*}$ for every $X \in R_C$.
    \end{enumerate}
\end{definition}

\begin{definition}[Solutions of a subproblem]\label{def:solutions}
    Let $\kappa = (L,I,R_C,r_C,R_W)$ be a subproblem. A solution of $\kappa$ is given by a partition $I=E\cup F$, $E\cap F=\emptyset$, together with values $e\in\Dom(E)$ such that $\restr{e}{X}\neq x^*$ for every $X\in E$. It determines the pair and the witness
    \[
        C = R_C\cup E,\qquad c = r_C\oplus e,\qquad W = R_W\cup F,
    \]
    where $r_C\oplus e\in\Dom(C)$ denotes the values equal to $r_C$ on $R_C$ and to $e$ on $E$, and it is a solution when
    \[
        C \neq \emptyset\qquad\text{and}\qquad (\Mod,u) \models [C \leftarrow c,\, W \leftarrow w^{*}](T\neq t^*).
    \]
    We write $\Sol(\kappa)$ for the set of solutions of $\kappa$, denoted by the triples $(C,c,W)$ they determine. We write $\AACtwo[\kappa] = \{(C,c) \mid (C,c,W) \in \Sol(\kappa)\}$ for the set of \emph{constrained pairs} that admit a canceling intervention.
\end{definition}

The free instance is split exhaustively: every variable of $I$ is either given a non-actual value and joins the cause, or held at its actual value and joins the witness. The forced variables belong to the pair and to the witness as well, so that $C\subseteq R_C\cup I$ and $W\subseteq R_W\cup I$.

\begin{definition}[Expansion]\label{def:expansion}
Let $\kappa = (L,I,R_C,r_C,R_W)$ be a subproblem, $(C,c,W) \in \Sol(\kappa)$ and $\emptyset
\neq S \subseteq C \cap I$.  The \emph{expansion of $\kappa$ along $(C,c,W,S)$} is
\[
  \cfx_{}(\kappa\;;\;C,c,W,S)\;=\;
  \Bigl(\; \underbrace{L \cup S}_{L'}, \;\;
           \underbrace{\Pa(S) \setminus (C \cup W \cup L')}_{I'}, \;\;
           \underbrace{C \setminus S}_{R_C'}, \;\; \underbrace{\restr{c}{C \setminus S}}_{r_C'}, \;\; \underbrace{W}_{R_W'} \;\Bigr).
\]
We write $\cfx(\kappa)=\{\cfx(\kappa;C,c,W,S)\;|\;\emptyset\neq S\subseteq C\cap I,\;(C,c,W)\in\Sol(\kappa)\}$ for the set of all subproblems so obtained.
\end{definition}

\begin{lemma}[Well-formedness]\label{lem:wf}
For any subproblem $\kappa=(L,I,R_C,r_C,R_W)$, solution $(C,c,W)\in\Sol(\kappa)$, and non-empty subset $\emptyset\neq S\subseteq C\cap I$, the expansion $\kappa'=\cfx(\kappa;C,c,W,S)$ is a subproblem, and $L' \supsetneq L$.
\end{lemma}

\begin{proof}
(i) For $I'$ with $L'$, $R_C'$, $R_W'$: $I'$ is disjoint from $C\setminus S\subseteq C$, from $W$ and from $L'$ by construction.

For $R_C'$ and $R_W'$: by Definition~\ref{def:solutions} we have $C \cap W = \emptyset$. Hence, $(C \setminus S) \cap W = \emptyset$.

For $L'$ and $R_C'$: by Definition~\ref{def:solutions} we have $C\subseteq R_C\cup I$. By Definition~\ref{def:subproblem} we have $R_C \cap L = \emptyset$ and $I \cap L = \emptyset$. Hence $C \cap L = \emptyset$. By construction $S\cap R_C'=\emptyset$.

For $L'$ and $R_W'$: by Definition~\ref{def:solutions} we have $W\subseteq (R_W\cup I)$ and $C\cap W=\emptyset$. By assumption we have $S\subseteq C\cap I$. By Definition~\ref{def:subproblem} we have $(R_W\cup I)\cap L=\emptyset$. Hence $W\cap L=\emptyset$ and $W\cap S=\emptyset$.

Finally $T\in L\subseteq L'$.

(ii) The values $\restr{c}{C\setminus S}$ are non-actual by
Definition~\ref{def:solutions}.

For $L'\supsetneq L$: we have $S\neq\emptyset$ and $S\subseteq I$, which is disjoint from $L$ by Definition~\ref{def:subproblem}; hence $L'=L\cup S\supsetneq L$.
\end{proof}

\begin{definition}[Counterfactual closure]\label{def:closure}
The \emph{root subproblem} is $\kappa_0 = (\{T\},\, \Pa(T),\, \emptyset,\, \emptyset,\,
\emptyset)$, which is well formed since $T \notin \Pa(T)$ by acyclicity.  $\CFX(\kappa_0)$ is
the least set of subproblems containing $\kappa_0$ and closed under $\kappa \mapsto
\cfx(\kappa)$.
\end{definition}

\begin{proposition}[Termination]\label{prop:termination}
Every $\kappa \in \CFX(\kappa_0)$ is reachable from $\kappa_0$ by at most $|\vars|-1$ expansions,
and $\CFX(\kappa_0)$ is finite.
\end{proposition}

\begin{proof}
By Lemma~\ref{lem:wf}, each expansion strictly enlarges the settled region, which is a subset
of $\vars$ and equals $\{T\}$ at the root; hence the depth of the expansion tree is at most
$|\vars|-1$.  Each subproblem has finitely many solutions and finitely many choices of $S$, so the
tree is finitely branching, and therefore finite.
\end{proof}

\paragraph{Specification of \textsc{Search}.}
$\textsc{Search}(\Mod,u,T,\kappa)$ takes a subproblem $\kappa = (L,I,R_C,r_C,R_W)$ and returns
$\Sol(\kappa)$, i.e.\ all partitions of $I$ into variables $E$ with values $e$ and variables $F$ held at their actual values, such that:
\[
  \restr{v^{\zeta}}{T} \;\neq\; t^{*}
  \qquad\text{where}\qquad
  \zeta \;=\; \bigl[\,R_C \leftarrow r_C,\;\; E \leftarrow e,\;\;
        F \leftarrow f^*,\;\; R_W \leftarrow \restr{v^{*}}{R_W}\,\bigr].
\]
This is the intervention $[C \leftarrow c,\, W \leftarrow w^{*}]$ of Definition~\ref{def:solutions}, written out along the four components of $\kappa$. The condition $C\neq\emptyset$ of Definition~\ref{def:solutions} need not be tested separately: if $R_C=\emptyset$ and $E=\emptyset$, then $\zeta$ forces every variable it touches to its actual value, so $v^{\zeta}=v^{*}$ and the target is not flipped.

ISI computes $\CFX(\kappa_0)$ by iterated expansion, calling \textsc{Search} on each subproblem. It halts by Proposition~\ref{prop:termination}.  Its output is
$\min_{\subseteq}\mathcal{F}$ where
\[
   \mathcal{F} \;=\; \bigl\{\, C \;\bigm|\; \exists c,\; \exists \kappa \in \CFX(\kappa_0):\;
   (C,c) \in \AACtwo[\kappa] \,\bigr\},
\]
and the three results below state that $\mathcal{F}$ consists of AC2 pairs, contains every
HP-cause, and has exactly the HP-causes as its minimal elements.

\subsection{The Region Lemma}

\begin{lemma}[Region lemma]\label{lem:region}
Let $X\subseteq\vars$, $x\in\Dom(X)$. Let $\xi=[X\leftarrow x]$ be an intervention and $v^\xi$ its outcome.  Let $L \subseteq \vars$ be a set
of variables none of which is forced by $\xi$, i.e., $L\cap X=\emptyset$. Let $\zeta=[Y\leftarrow y]$, $Y\subseteq\vars$, $y\in\Dom(Y)$ be an intervention such that
\begin{enumerate}
\item[(i)] no variable of $L$ is forced by $\zeta$: $Y\cap L =\emptyset$;
\item[(ii)] variables of $\Pa(L) \setminus L$ are forced by $\zeta$ to their value under $\xi$: $\Pa(L)\setminus L\subseteq Y$ and $\restr{y}{\Pa(L)\setminus L}=\restr{v^\xi}{\Pa(L)\setminus L}$.
\end{enumerate}
Then $v^\zeta$ and $v^\xi$ agree on $L$: $\restr{v^\zeta}{L}=\restr{v^\xi}{L}$. Notably, if $T\in L$ and $\xi$ “cancels” $T$, then (i)-(ii) imply that $\zeta$ also cancels $T$.
\end{lemma}

\begin{proof}
By induction along a topological order of $L$.

Initialization: Let $Z\in L$ be the first variable of $L$ in the topological order, so that $\Pa(Z)\cap L=\emptyset$ and hence $\Pa(Z)\subseteq Y$: all parents of $Z$ are forced by $\zeta$ to their values under $\xi$. Since $Z$ is forced neither by $\zeta$ ($Y\cap L=\emptyset$) nor by $\xi$ ($X\cap L=\emptyset$), its value is set by its parents under both interventions. Hence $\restr{v^\zeta}{Z}=\restr{v^\xi}{Z}$.

Induction: Let $Z \in L$ and assume the claim for the variables of $L$ that precede $Z$.  Since $Z$ is not forced by $\zeta$, its value is set by its parents. Every $P \in \Pa(Z)$ lies either in $L$, in which case $\restr{v^\zeta}{P}=\restr{v^\xi}{P}$ by the induction hypothesis, or in $\Pa(L) \setminus L$. In the second case, $\zeta$ forces $P$ to its value under $\xi$.  Hence all parents of $Z$ take their value under $\xi$. Since $Z$ is not forced by $\xi$ either, $\restr{v^\zeta}{Z}=\restr{v^\xi}{Z}$.
\end{proof}

\subsection{Soundness}

\begin{theorem}[Soundness]\label{thm:sound}
For every $\kappa \in \CFX(\kappa_0)$ and every $(C,c) \in \AACtwo[\kappa]$ we have $(C,c) \in
\AACtwo$.  Consequently $\mathcal{F} \subseteq \{C \mid \exists c,\ (C,c) \in \AACtwo\}$.
\end{theorem}

\begin{proof}
Let $(C,c,W)$ be the corresponding solution of $\kappa$. By Definition~\ref{def:solutions}, $C \neq \emptyset$; $c$ is pointwise non-actual, on $R_C$ by Definition~\ref{def:subproblem}(ii) and on $E$ by Definition~\ref{def:solutions}. By Definition~\ref{def:subproblem}(i), the four components of $\kappa$ are pairwise disjoint and $T\in L$, so $C\subseteq \vars\setminus\{T\}$ and $W \subseteq \vars \setminus (C\cup\{T\})$.
\end{proof}

\subsection{Completeness}

Fix for this section an HP-cause $C' \in \mathcal{C}$, with values $c'$ and witness $W' \in \Wit(C',c')$.  Write
\[
  \xi' = [\,C' \leftarrow c',\ W' \leftarrow w'^{*}\,], \qquad v' = v^{\xi'}, \qquad
  D = \{\, Z \in \vars \mid \restr{v'}{Z} \neq z^{*} \,\}
\]
for the canceling intervention, its outcome, and its set of \emph{deviating} variables.  By
construction $T \in D$ (this is the cancellation $\restr{v'}{T} \neq t^{*}$), $C' \subseteq D$ (the
members of $C'$ are forced to non-actual values) and $W' \cap D = \emptyset$ (the members of
$W'$ are forced to their actual values, and $T \notin W'$ by convention).

\begin{definition}[Backtracking chain of $C'$]\label{def:chain}
Set
\[
  L_1 = \{T\}, \qquad K_1 = \Pa(T) \cap D, \qquad B_1 = \Pa(T) \setminus D, \qquad
  \kappa_1 = \kappa_0 ,
\]
so that $I_1 = \Pa(T)$ and there are no forced variables in $\kappa_1$.
For $j \geq 1$: if $K_j \subseteq C'$ the chain stops and we set $m = j$; otherwise let
\begin{gather*}
  S_j = K_j \setminus C', \qquad
  L_{j+1} = L_j \cup S_j, \qquad
  I_{j+1} = \Pa(S_j) \setminus (K_j \cup B_j \cup L_{j+1}), \\[2pt]
  \kappa_{j+1} = \bigl(L_{j+1},\; I_{j+1},\; K_j \setminus S_j,\;
                        \restr{v'}{K_j \setminus S_j},\; B_j\bigr), \\[2pt]
  K_{j+1} = (K_j \setminus S_j) \cup (I_{j+1} \cap D), \qquad
  B_{j+1} = B_j \cup (I_{j+1} \setminus D).
\end{gather*}
\end{definition}

$L_j$ is the region already settled around the target; its parent boundary is entirely forced, with the deviating boundary parents in $K_j$ (at their values under $\xi'$) and the non-deviating ones in $B_j$ (at their actual values kept under $\xi'$). Each stage absorbs into the region the boundary parents that are not members of $C'$ and pushes the boundary one layer of parents upward, until only members of $C'$ remain on it.

Note that $K_1 = I_1 \cap D$ and that, for $j \geq 2$, the definition gives $K_j = (K_{j-1} \cap C') \cup (I_j \cap D)$, since $K_{j-1} \setminus S_{j-1} = K_{j-1} \cap C'$. Hence, when the chain continues: $\emptyset \neq S_j = K_j \setminus C' \subseteq I_j \cap D$ and $S_j \subseteq K_j \cap I_j$.

\begin{lemma}[Chain invariants]\label{lem:invariants}
At every stage $j \geq 1$ of the chain:
\begin{enumerate}
\item[(a)] (i) $K_j$ has counterfactual values under $\xi'$; (ii) $B_j$ has actual values under $\xi'$; (iii) $K_j$ and $B_j$ are disjoint; (iv) $L_j$ is disjoint from $K_j$ and $B_j$; (v) $L_j$ has counterfactual values under $\xi'$ and is disjoint from $C'$. Formally:
\[
    (i)\ K_j \subseteq D \; \quad
    (ii)\ B_j \cap D = \emptyset \;\quad
    (iii)\ K_j \cap B_j = \emptyset \; \quad
    (iv)\ L_j \cap (K_j \cup B_j) = \emptyset \;\quad
    (v)\ L_j \subseteq D \setminus C'
\]
\item[(b)] The parent boundary of $L_j$ is contained in $K_j$ and $B_j$: $\Pa(L_j) \setminus L_j \subseteq K_j \cup B_j$;
\item[(c)] (i) $\kappa_j$ is a subproblem, (ii) $\kappa_j \in \CFX(\kappa_0)$, and (iii) $\bigl(K_j, \restr{v'}{K_j}, B_j\bigr) \in \Sol(\kappa_j)$ is a solution of $\kappa_j$.
\end{enumerate}
\end{lemma}

\begin{proof}
By induction on $j$.  Throughout, write
$\zeta_j = [\,K_j \leftarrow \restr{v'}{K_j},\ B_j \leftarrow b_j^{*}\,]$ and
note that, as soon as (a) holds at stage $j$, \emph{every variable of $K_j \cup B_j$ is forced
by $\zeta_j$ to its value under $\xi'$}: this is immediate for $K_j$, and for $B_j$ it holds because
$B_j \cap D = \emptyset$ means $v'$ and $v^{*}$ agree on $B_j$.

\medskip\noindent\emph{Base $j = 1$.}
(a) $K_1 \subseteq D$ and $B_1 \cap D = \emptyset$ hold by definition, as does $K_1 \cap B_1 =
\emptyset$.
By acyclicity, $L_1 \cap (K_1 \cup B_1) = \emptyset$. $L_1 = \{T\} \subseteq D \setminus C'$ because $T \in D$ and $T \notin C'$ by convention.

(b) $\Pa(L_1) \setminus L_1 = \Pa(T) = K_1 \cup B_1$.

(c) (i) $\kappa_1$ is a subproblem: Definition~\ref{def:subproblem}(i) holds by acyclicity, with $T\in L_1$, and Definition~\ref{def:subproblem}(ii) holds vacuously since $R_C=\emptyset$.

(ii) $\kappa_1 = \kappa_0 \in \CFX(\kappa_0)$ by definition.

(iii) We have $I_1=\Pa(T)=K_1\cup B_1$, so $E=K_1$ and $F=B_1$ is a partition of $I_1$, and since $R_C=R_W=\emptyset$ it determines the pair $K_1$ and the witness $B_1$. Values of $K_1$ are pointwise non-actual by construction ($K_1\subseteq D$). We have $K_1\neq\emptyset$: $T$ is not forced by $\xi'$ by convention, so its value under $\xi'$ is set by its parents; were they all non-deviating, $T$ would take its actual value, contradicting $T\in D$.
For the cancellation, we apply Lemma~\ref{lem:region} using $\xi=\xi'$, $\zeta=\zeta_1$ and $L=L_1$. By acyclicity, no variable of $L$ is forced by $\zeta_1$. By convention ($T\notin C'\cup W'$) we have the same for $\xi'$. By the observation above, every variable of $\Pa(L_1)\setminus L_1=K_1\cup B_1$ is forced by $\zeta_1$ to its value under $\xi'$, which is condition (ii) of Lemma~\ref{lem:region}. Hence $\restr{v^{\zeta_1}}{T}=\restr{v'}{T}\neq t^*$.

\medskip\noindent\emph{Step $j \to j+1$.}
(a) (i) $K_{j+1}$ is made of $K_j$ (in $D$ by hypothesis) and of the part of $I_{j+1}$ which intersects $D$.

(ii) Similarly, $B_{j+1}$ is made of $B_j$ (disjoint from $D$ by hypothesis) and of the part of $I_{j+1}$ disjoint from $D$.

(iii) If we expand $K_{j+1}\cap B_{j+1}$, we obtain four terms. The intersection of $K_j\setminus S_j$ and $B_j$, which is empty by assumption; the intersection of $K_j\setminus S_j$ and $I_{j+1}\setminus D$, which is empty because $K_j\subseteq D$; the intersection of $I_{j+1}\cap D$ and $B_j$, empty because $B_j\cap D=\emptyset$; and the intersection of $I_{j+1}\cap D$ and $I_{j+1}\setminus D$, which is also empty. Hence, $K_{j+1}\cap B_{j+1}=\emptyset$.

(iv) If we decompose $L_{j+1}\cap(K_{j+1}\cup B_{j+1})$, we obtain two terms. For $L_{j+1}=L_j\cup S_j$ and $K_{j+1}=(K_j\setminus S_j)\cup(I_{j+1}\cap D)$, $K_j\setminus S_j$ is disjoint from $L_{j+1}$ because $K_j\cap L_j=\emptyset$ by assumption and $(K_j\setminus S_j)\cap S_j=\emptyset$ by construction; and $I_{j+1}\cap D$ is disjoint from $L_{j+1}$ by construction of $I_{j+1}$. For $L_{j+1}$ and $B_{j+1}$, the same decomposition gives us $L_{j+1}$ disjoint from $B_j$ because $L_j$ is by assumption and $S_j\subseteq K_j$ with $K_j\cap B_j=\emptyset$; and $L_{j+1}$ disjoint from $I_{j+1}\setminus D$ by construction of $I_{j+1}$.

(v) By assumption, $L_j\subseteq D\setminus C'$ and $S_j\subseteq K_j\setminus C'\subseteq D\setminus C'$. Hence, $L_{j+1}=L_j\cup S_j\subseteq D\setminus C'$.

(b) Let $P\in \Pa(L_{j+1})\setminus L_{j+1}$. Then either $P\in \Pa(L_j)\setminus L_{j+1}$ or $P\in \Pa(S_j)\setminus L_{j+1}$. On the one hand, if $P\in \Pa(L_j)\setminus L_{j+1}$, then $P\in\Pa(L_j)\setminus L_j$. Hence, $P\in K_j\cup B_j$ by assumption. If $P\in K_j$, then it is in $K_{j+1}$ (because $P\notin S_j$ as $P\notin L_{j+1}$ by assumption). If $P\in B_j$ then $P\in B_{j+1}$. In both cases, if $P\in \Pa(L_j)\setminus L_{j+1}$ then $P\in K_{j+1}\cup B_{j+1}$. On the other hand, we can assume that $P\in \Pa(S_j)\setminus L_{j+1}$ and that $P\notin (K_j\cup B_j)$ (as it has already been treated in the previous case). Hence $P\in\Pa(S_j)\setminus(K_j\cup B_j\cup L_{j+1})=I_{j+1}$, and $I_{j+1}=(I_{j+1}\cap D)\cup(I_{j+1}\setminus D)\subseteq K_{j+1}\cup B_{j+1}$. In any case, $P\in K_{j+1}\cup B_{j+1}$.

(c) (i)--(ii) By the induction hypothesis, $\kappa_j \in \CFX(\kappa_0)$ and $(K_j,\restr{v'}{K_j},B_j)\in\Sol(\kappa_j)$. By the observation after Definition~\ref{def:chain}, $\emptyset \neq S_j\subseteq K_j\cap I_j$, so $\kappa_{j+1}=\cfx(\kappa_j;K_j,\restr{v'}{K_j},B_j,S_j)$, whose components match Definition~\ref{def:chain}. It is a subproblem (Lemma~\ref{lem:wf}), lies in $\cfx(\kappa_j)$ (Definition~\ref{def:expansion}), and hence in $\CFX(\kappa_0)$ (Definition~\ref{def:closure}).

(iii) By construction $K_{j+1}=(K_j\setminus S_j)\cup(I_{j+1}\cap D)$ and $B_{j+1}=B_j\cup(I_{j+1}\setminus D)$, so $E=I_{j+1}\cap D$ and $F=I_{j+1}\setminus D$ is a partition of $I_{j+1}$ determining the pair $K_{j+1}$ and the witness $B_{j+1}$, with values $\restr{v'}{K_{j+1}}$ that are non-actual as $K_{j+1}\subseteq D$. For the cancellation, we apply Lemma~\ref{lem:region} with $L=L_{j+1}$, $\xi=\xi'$ and $\zeta=\zeta_{j+1}$.

\begin{itemize}
    \item no variable of $L_{j+1}$ is forced by $\xi'$: by (a.v), $L_{j+1} \subseteq D \setminus C'$ is disjoint from $C'$, and it is disjoint from $W'$ since $W' \cap D = \emptyset$;
    \item no variable of $L_{j+1}$ is forced by $\zeta_{j+1}$: this is $L_{j+1} \cap (K_{j+1} \cup B_{j+1}) = \emptyset$ from (a.iv);
    \item every variable of $\Pa(L_{j+1}) \setminus L_{j+1}$ is forced by $\zeta_{j+1}$ to its values under $\xi'$: by (b) such a variable lies in $K_{j+1} \cup B_{j+1}$, and all of those are forced to their values under $\xi'$.
\end{itemize}

Additionally, $T\in L_1\subseteq L_{j+1}$, hence $\restr{v^{\zeta_{j+1}}}{T}=\restr{v'}{T}\neq t^*$. Finally $K_{j+1}\neq\emptyset$: let $Z$ be a variable of $L_{j+1}$ minimal in a topological order, so that every parent of $Z$ lies in $\Pa(L_{j+1})\setminus L_{j+1}\subseteq K_{j+1}\cup B_{j+1}$; by (a.v) and $W'\cap D=\emptyset$, $Z$ is not forced by $\xi'$, so its value under $\xi'$ is set by its parents; were $K_{j+1}$ empty, all of them would be non-deviating and $Z$ would take its actual value, contradicting $L_{j+1}\subseteq D$. Hence $(K_{j+1},\restr{v'}{K_{j+1}},B_{j+1})\in\Sol(\kappa_{j+1})$.
\end{proof}

\begin{lemma}[The chain reaches $C'$]\label{lem:reaches}
The chain stops after $m \leq |\vars|$ stages, and $K_m = C'$ with $\restr{v'}{C'} = c'$.
\end{lemma}

\begin{proof}
While the chain runs, $S_j \neq \emptyset$ and $S_j \subseteq K_j$ is disjoint from $L_j$ by Lemma~\ref{lem:invariants}(a), so $L_{j+1} \supsetneq L_j$; as $L_1 = \{T\}$ and $L_j \subseteq \vars$, there are at most $|\vars|$ stages.  At the stop, $\emptyset \neq K_m \subseteq C'$ and $\bigl(K_m, \restr{v'}{K_m}\bigr) \in \AACtwo[\kappa_m]$ by Lemma~\ref{lem:invariants}(c), hence $\bigl(K_m, \restr{v'}{K_m}\bigr) \in \AACtwo$ by Theorem~\ref{thm:sound}.  If $K_m \subsetneq C'$, this exhibits a strict subset of $C'$ in $\AACtwo$, contradicting AC3.  Therefore $K_m = C'$, and $\restr{v'}{C'} = c'$ because $\xi'$ forces $C'$ to $c'$.
\end{proof}

\begin{theorem}[Completeness]\label{thm:complete}
For every HP-cause $C' \in \mathcal{C}$ with values $c'$, there is $\kappa \in
\CFX(\kappa_0)$, reachable from $\kappa_0$ by at most $|\vars|-1$ expansions, such that $(C',c')
\in \AACtwo[\kappa]$.  In particular $\mathcal{C} \subseteq \mathcal{F}$.
\end{theorem}

\begin{proof}
Take $\kappa = \kappa_m$ from Definition~\ref{def:chain} and combine
Lemmas~\ref{lem:invariants}(c) and~\ref{lem:reaches}: the chain visits $m \leq |\vars|$
subproblems, i.e.\ performs $m-1 \leq |\vars|-1$ expansions, and its last pair is $(C',c')$.
\end{proof}

\subsection{Exactness}

\begin{corollary}[Exactness]\label{cor:exact}
$\min_{\subseteq}\mathcal{F} = \mathcal{C}$.  That is, ISI returns exactly the HP-causes.
\end{corollary}

\begin{proof}
$\mathcal{C} \subseteq \mathcal{F}$ by Theorem~\ref{thm:complete}, and every element of
$\mathcal{F}$ satisfies AC1--AC2 by Theorem~\ref{thm:sound}.

Let $C' \in \mathcal{C}$.  Any strict subset of $C'$ lying in $\mathcal{F}$ would satisfy
AC1--AC2, contradicting AC3; hence $C'$ is minimal in $\mathcal{F}$, and $\mathcal{C}
\subseteq \min_{\subseteq}\mathcal{F}$.
Conversely let $M \in \min_{\subseteq}\mathcal{F}$.  By Theorem~\ref{thm:sound}, $(M,c) \in \AACtwo$ for some $c$. Among the subsets of $M$ belonging to $\AACtwo$ for some values (a non-empty finite family), choose an inclusion-minimal element $M_0$.  Every strict subset of $M_0$ is a subset of $M$, hence not in $\AACtwo$ for any values, so $M_0$ satisfies AC3 and $M_0 \in \mathcal{C}$.  By Theorem~\ref{thm:complete}, $M_0 \in \mathcal{F}$, and minimality of $M$ in $\mathcal{F}$ forces $M = M_0 \in \mathcal{C}$.
\end{proof}